\documentclass{article}

\usepackage{iclr2026_conference,times}
\usepackage{bm}

\usepackage{amsmath,amsfonts,bm}

\def\eqref#1{equation~\ref{#1}}

\def\1{\bm{1}}

\DeclareMathAlphabet{\mathsfit}{\encodingdefault}{\sfdefault}{m}{sl}
\SetMathAlphabet{\mathsfit}{bold}{\encodingdefault}{\sfdefault}{bx}{n}

\usepackage[T1]{fontenc}
\usepackage[utf8]{inputenc}
\usepackage{microtype}
\usepackage{inconsolata}

\usepackage{amsthm}
\usepackage{amsmath}
\usepackage{amsfonts}
\usepackage{amssymb}
\usepackage{bbm}
\usepackage{dsfont}

\usepackage{graphicx}
\usepackage{booktabs}
\usepackage{array}
\usepackage{longtable}
\usepackage{threeparttable}
\usepackage{multirow}
\usepackage{colortbl}
\usepackage{wrapfig}
\usepackage{float}
\usepackage{placeins}
\usepackage{subcaption}
\usepackage[font=small]{caption}

\usepackage{xcolor}
\usepackage{color}
\usepackage{pifont}
\usepackage{marvosym}
\usepackage{soul}

\usepackage{tikz}
\usetikzlibrary{decorations.markings, arrows.meta, shapes.geometric, calc, patterns, patterns.meta, backgrounds, spath3, intersections, svg.path, hobby}
\usepackage[most]{tcolorbox}
\tcbuselibrary{listings,breakable}
\usepackage{xurl}
\usepackage{url}
\usepackage{hyperref}
\usepackage{listings}
\usepackage{enumitem}
\usepackage[ruled,vlined,linesnumbered]{algorithm2e}

\usepackage{lipsum}
\usepackage{censor}
\usepackage{pgf}
\usepackage{titletoc}
\usepackage{svg}

\usepackage{adjustbox}
\usepackage{wrapstuff}

\SetAlCapNameFnt{\small}
\SetAlCapFnt{\small}

\definecolor{lightgreen}{HTML}{d5e8d4}
\definecolor{LightRed}{HTML}{f8cecc}
\definecolor{LightYellow}{HTML}{fff2cc}
\definecolor{diffgreen}{HTML}{82B366}
\definecolor{diffred}{HTML}{B85450}
\definecolor{diffgray}{RGB}{90,90,90}

\definecolor{gradBad}{HTML}{f8cecc}   
\definecolor{gradMid}{HTML}{fff2cc}   
\definecolor{gradGood}{HTML}{d5e8d4}  

\makeatletter
\newcommand{\@gradmix}[3]{
  \pgfmathtruncatemacro{\@gpct}{min(100,max(0,(#3-#1)/(#2-#1)*100))}%
  \ifnum\@gpct<50
    \pgfmathtruncatemacro{\@gloc}{\@gpct*2}%
    \edef\gradcellcmd{\noexpand\cellcolor{gradMid!\@gloc!gradBad}}%
  \else
    \pgfmathtruncatemacro{\@gloc}{(\@gpct-50)*2}%
    \edef\gradcellcmd{\noexpand\cellcolor{gradGood!\@gloc!gradMid}}%
  \fi
}
\newcommand{\gradcell}[4]{\@gradmix{#1}{#2}{#3}\gradcellcmd #4}

\newcommand{\cAcc}[2]{\gradcell{0}{93.2}{#1}{#2}}
\newcommand{\cImpr}[2]{\gradcell{-60}{83.5}{#1}{#2}}
\newcommand{\cTime}[2]{\gradcell{155}{40.14}{#1}{#2}}

\expandafter\def\expandafter\normalsize\expandafter{%
    \normalsize%
    \setlength\abovedisplayskip{3pt}%
    \setlength\belowdisplayskip{3pt}%
    \setlength\abovedisplayshortskip{0pt}%
    \setlength\belowdisplayshortskip{0pt}%
}

\lstdefinelanguage{diff}{
  morecomment=[f][\color{diffgreen}]{+},
  morecomment=[f][\color{diffred}]{-},
  morecomment=[f][\color{diffgray}]{@@},
}

\lstdefinestyle{diffstyle}{
  language=diff,
  basicstyle=\ttfamily\small,
  columns=fullflexible,
  keepspaces=true,
  breaklines=true,
  frame=single,
  rulecolor=\color{black!20},
  backgroundcolor=\color{black!3},
  xleftmargin=1em,
  xrightmargin=1em,
  aboveskip=0.8em,
  belowskip=0.8em,
}

\newtcblisting{promptfigurebox}{
  colback=gray!5,
  colframe=gray!30,
  boxrule=0.4pt,
  arc=1pt,
  left=5pt,
  right=5pt,
  top=5pt,
  bottom=5pt,
  listing only,
  listing options={
    basicstyle=\ttfamily\footnotesize,
    breaklines=true,
    columns=fullflexible,
    keepspaces=true
  }
}

\SetKwInput{KwInput}{Input}
\SetKwInput{KwOutput}{Output}
\SetKwComment{tcp}{// }{}
\DontPrintSemicolon

\usepackage[most]{tcolorbox}
\usepackage{xcolor}
\usepackage{fontawesome5}

\usepackage{utfsym}

\definecolor{abstractbg}{HTML}{F2F2F2}

\newcommand{\paperresource}[3]{%
  \noindent
  #1\enspace\textbf{#2:}\enspace
  \href{#3}{\nolinkurl{#3}}\par
}

\title{Mendel G\"odel Machine:\\
Recursive Self-Improving Coding Agents\\via Comparative Evolution}

\iclrfinalcopy

\author{
\textbf{\textnormal{Changzhi Liu}}\textsuperscript{$\ast,\S,\usym{1F582}$} \quad
\textbf{\textnormal{Yilun Liu}}\textsuperscript{$\dagger,\ddagger,\S,\usym{1F582}$} \quad
\textbf{\textnormal{Sikuan Yan}}\textsuperscript{$\dagger,\ddagger$} \quad
\textbf{\textnormal{Volker Tresp}}\textsuperscript{$\dagger,\ddagger$} \quad
\textbf{\textnormal{Yunpu Ma}}\textsuperscript{$\dagger,\ddagger,\usym{1F582}$} \\
\textsuperscript{$\ast$}\text{University of Electronic Science and Technology of China}\\
\textsuperscript{$\dagger$}\text{Ludwig Maximilian University of Munich} \quad
\textsuperscript{$\ddagger$}\text{Munich Center for Machine Learning} \\
\normalsize \textsuperscript{\usym{1F582}}\texttt{changzhiliu1@gmail.com} \quad \texttt{yilun.liu@tum.de} \quad \texttt{cognitive.yunpu@gmail.com}\\
\normalsize \textsuperscript{\S}\textit{Equal contribution.}
}

\begin{document}
\maketitle




        \vspace{-2.5ex}
\begin{abstract}
        \vspace{-1ex}
        
Self-improving coding agents that iteratively rewrite their own source code have demonstrated impressive performance on coding tasks.
However, existing solutions generally derive self-modification from a single failure trajectory at a time, overlooking rich comparative signals available in the agent's expanding archive of past attempts.
According to Mendelian principles of controlled inheritance, we introduce Mendel G\"odel Machine (MGM).
In addition to the general single-trajectory \emph{clonal mutation},
MGM includes two new types of self-modification that better utilizes evidences accumulated: the \emph{reaction-norm mutation} edits an agent based on its trajectories on multiple tasks simultaneously, and the \emph{cross-lineage hybridization} edits an agent using the trajectory of a reference agent from another lineage on the same task.
Under an additive fitness landscape model, we prove theoretically and demonstrate via controlled surrogate simulation that the new strategies facilitate a faster and better convergence over single-trajectory baselines.
Experiments on SWE-bench and Polyglot confirm MGM's consistent improvement in performance, efficiency, and generalizability. 

\vspace{0.6em}

\paperresource{\faGlobe}{Project Page}
  {https://reallcz.github.io/MGM/}

\paperresource{\faGithub}{Code}
  {https://github.com/RealLcz/MGM}

\end{abstract}

        \vspace{-1.5ex}
\section{Introduction}
        \vspace{-0.5ex}

The vision of artificial intelligence recursively rewriting itself to become better traces back several decades~\citep{schmidhuber1987evolutionary}.
G\"odel Machine~\citep{schmidhuber2003goedel, schmidhuber2007godel} conceives a mathematically rigorous, self-referential modification process when a provable (measurable) benefit can be derived.
Recent advances in large language models (LLMs) and coding agents have started to empirically realize such an idea~\citep{yang2024sweagent, ADAS, gao2026survey}. 
\citet{robeyns2025sica} show that an agent equipped with basic file-editing tools can autonomously refactor its own codebase and lift its performance.
\citet{zhang2026darwin} reframe this loop as open-ended evolution, 
maintaining an expanding archive of agent variants, from which at each iteration one is sampled to seed the next self-modification.
\citet{HGM} improve the sampling policy by using the aggregated performance of agents' descendants as signals guiding evaluation and expansion operations.

However, progress so far has focused on improving the archive that stores generated agents and evaluated traces, and on the process that samples which agent to evaluate or edit next.
The \emph{self-modification} process, during which the agent edits its own source code, has remained essentially underexplored: each self-modification step is conditioned only on one agent's single trajectory (typically a recent failure) on one task.
The expanding archive, which records all agent variants ever generated together with their behavior on all evaluated tasks, is used only as a leaderboard for sampling, overlooking rich comparative evidence that can facilitate better-informed edits.

\begin{figure*}[t]
    \centering
        \includegraphics[width=\linewidth]{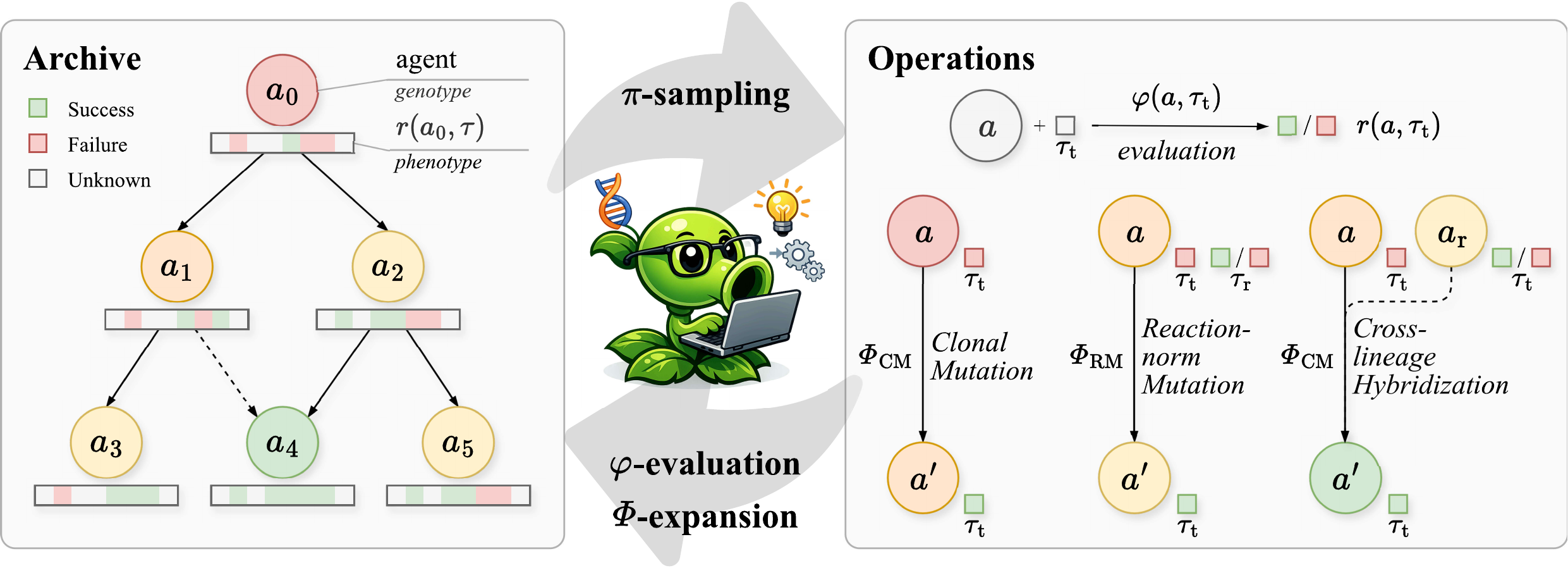}
        \vspace{-3.5ex}
        \caption{\textbf{Mendel G\"odel Machine}. MGM organizes self-modification via controlled inheritance based on evidence across tasks and lineages. 
        The archive maintains a lineage tree of agent variants; each stores its source code as genotype and evaluation outcomes as phenotype.
        Each iteration applies $\pi$-sampling that selects an operation to perform with the necessary resources from the archive. 
        $\varphi$-evaluation executes the agent on untested tasks and records its trajectory and results.
        Clonal mutation $\varPhi_\mathrm{CM}$ edits the agent based on a single failure trajectory on target task $\tau_\mathrm{t}$.
        Reaction-norm mutation $\varPhi_\mathrm{RM}$ uses the agent's trajectories across reference tasks $\tau_\mathrm{r}$.
        Cross-lineage hybridization $\varPhi_\mathrm{CH}$ uses a reference agent $a_\mathrm{r}$ from a different lineage that attempted the same task.        
        }
        \label{fig:fig1}
        \vspace{0.5ex}
\end{figure*}

\begin{wrapstuff}[r, top=4, type=figure, width=0.52\linewidth, abovesep=0ex, belowsep=0ex]
        \centering
        \vspace{-0.1ex}
        \includegraphics[width=.99\linewidth]{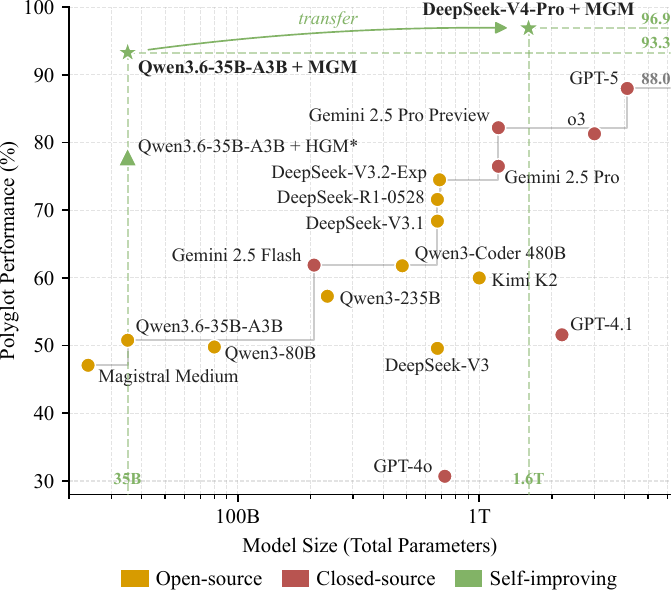}
        \vspace{-3.5ex}
        \caption[Polyglot performance]{\textbf{Polyglot performance.} Result marked with asterisk is from Polyglot-60; all other scores are from the complete Polyglot‑225\footnotemark. 
        Closed-source model sizes follows estimates reported by \citet{li2026Incompressible}.
        }
        \label{fig:polyglot}
        \vspace{-0.5ex}
\end{wrapstuff}

We identify two types of such comparative signals.
First, when an agent is evaluated across multiple tasks, the pattern of its successes and failures forms a reaction norm~\citep{woltereck1909reaktionsnorm, pigliucci2001phenotypic}: a stable, genotype-specific profile of how performance varies across environments.
Recurring failure modes can more plausibly distinct genotype-level defects from task-specific accidents.
Second, when multiple agents have attempted the same task, their trajectories reveal transferable behavioral traits.
Conditioning edits on these contrastive evidence enables targeted ability transfer across lineages and reduce redundant exploration.
Building on these observations,  we introduce \textbf{Mendel G\"odel Machine (MGM)}, which isolates heritable effects through controlled comparisons as in Mendelian genetics. 
As shown in Figure~\ref{fig:fig1}, we structure self-modification into three operators:
\emph{clonal mutation} denotes standard single-agent, single-trajectory self-modification; 
\emph{reaction-norm mutation} edits an agent conditioned on its trajectories across multiple tasks; and
\emph{cross-lineage hybridization} edits an agent using a reference agent's trajectory on the same task.
All strategies directly operate on trajectories accumulated during routine archive evaluation and therefore incur no extra task evaluations.

To isolate the contribution of each dimension rigorously, we develop
a formal additive fitness landscape model in which each agent is represented as a binary vector in genotype space, and self-improvement progress is measured as the reduction in Hamming distance to an oracle genotype.
We demonstrate that MGM yields a strictly faster expected convergence than single-trajectory baselines, and we validate this in controlled Monte Carlo surrogate simulations.

Experiments on Polyglot~\citep{polyglot} and SWE-bench~\citep{SWEbench} confirm MGM's consistent gains over baselines in both performance and efficiency. 
Purely through evolving the agent scaffold, MGM advances Qwen3.6-35B-A3B on Polyglot from 50.8\% to 93.3\%, surpassing closed-source GPT-5~\citep{GPT5} with $\sim$117$\times$ fewer parameters, as Figure~\ref{fig:polyglot} shows.
MGM also exhibits stronger generalizability across both unseen benchmarks and backbone LLMs. 
Notably, transferring the Qwen-evolved scaffold to DeepSeek-V4-Pro yields 96.9\% on Polyglot.
These findings demonstrate that MGM's richer comparative conditioning discovers reusable, workflow-level improvements that support strong coding-agent performance, indicating its potential as a scalable approach for future self-improving agent development.

\footnotetext{Sourced from Aider-Polyglot Leaderboard \url{https://llm-stats.com/benchmarks/aider-polyglot}.}
\section{Preliminaries}
\label{sec:background}

We denote an executable coding agent $a\in\mathcal{A}$, together with its auxiliary scaffolding, as the genotype that self-improvement seeks to evolve. 
Running $a$ on a task $\tau\sim\mathcal{D}$ yields an evaluation trajectory $\varphi(a,\tau)$ and a binary outcome $r(a,\tau)\in\{0,1\}$, which is viewed as its phenotype.
The expected utility of $a$ is
\begin{align}
U(a)=\mathbb{E}_{\tau\sim\mathcal{D}}[r(a,\tau)].
\end{align}
A self-modification process $\varPhi$ lets the parent agent edit its own code using diagnostic evidence $E$, a finite set of stored trajectory--outcome pairs:
\begin{align}
a' \leftarrow \varPhi(a,E),\qquad
E\subseteq\{(\varphi(a,\tau),r(a,\tau))\}.
\end{align}

DGM~\citep{zhang2026darwin} introduces an archive-based self-improvement framework by maintaining an expanding tree of generated agents and their evaluation trajectories. 
HGM~\citep{HGM} further formulates it as a fixed-budget tree-search problem. 
Let $\mathcal{G}_t$ be the archive tree at
step $t$ and $\mathcal{V}_t$ its node set. For node $a_i\in\mathcal{V}_t$,
let $S_i$ be its evaluated tasks, $F_i\subseteq S_i$ its failed tasks, and
\begin{align}
n_\mathrm{s}(a_i)=|S_i|-|F_i|,\qquad n_\mathrm{f}(a_i)=|F_i|.
\end{align}
At each step $t$, HGM chooses to allocate either an evaluation or an expansion to its archive. An expansion happens when
\begin{align}
N_t^\alpha \ge |\mathcal{V}_t|,\qquad
N_t=\sum_{a\in\mathcal{V}_t}(n_\mathrm{s}(a)+n_\mathrm{f}(a)),
\end{align}
where $\alpha\in[0,1]$ is a widening parameter, 
and otherwise another task evaluation is performed. 

The evaluation policy samples $a$ using Thompson sampling from the node-level posterior
\begin{align}
\pi_a\sim
\mathrm{Beta}(\kappa(1+n_\mathrm{s}(a)),\kappa(1+n_\mathrm{f}(a))),
\end{align}
where $\kappa>0$ is the concentration parameter controlling the exploration--exploitation trade-off.
The expansion policy samples $a$ to expand using clade-level evidence.
Let $C_t(a)$ denote the subtree rooted at $a$, and define
\begin{align}
n_\mathrm{s}^C(a)=\sum_{a'\in C_t(a)}n_\mathrm{s}(a'), \qquad
n_\mathrm{f}^C(a)=\sum_{a'\in C_t(a)}n_\mathrm{f}(a').
\end{align}
HGM samples expansion candidates from
\begin{align}
\pi_a^C\sim
\mathrm{Beta}(\kappa(1+n_\mathrm{s}^C(a)),\kappa(1+n_\mathrm{f}^C(a))).
\end{align}

\section{Mendel G\"odel Machine}
\label{sec:methodology}

MGM builds on the aforementioned tree-search framework of selection, evaluation, and expansion policies. 
Instead of relying on single-trajectory self-modification, MGM partitions the expansion operator $\varPhi$ into three specialized sub-operators based on the type of diagnostic evidence $E$ available in the archive:
clonal mutation $\varPhi_{\rm CM}$, reaction-norm mutation $\varPhi_{\rm RM}$, and cross-lineage hybridization $\varPhi_{\rm CH}$, as Figure~\ref{fig:fig1} shows.

\subsection{Mendelian Self-Modification Operators}

Analogous to Mendelian genetics, which seeks to isolate heritable effects through controlled comparisons, we design self-modification operators as diagnostic processes that ask the selected agent to make general improvements to its genotype based on different phenotype evidence. 

\paragraph{Clonal Mutation.}    
Clonal mutation is the standard single-agent, single-trajectory
self-improvement operator. It is used when MGM has only one informative
failure or cannot construct a reliable comparison. Given a selected
node \(i\) and a failed task \(\tau\in F_i\), the evidence is
\begin{equation}
    E_{\mathrm{CM}}(i,\tau)=\{(\varphi(a_i,\tau),r(a_i,\tau))\}.
\end{equation}
The editor diagnoses the failure and modifies \(a_i\) to avoid similar
failures in future tasks:
\begin{equation}
    a' \leftarrow \varPhi_{\mathrm{CM}}(a_i,E_{\mathrm{CM}}).
\end{equation}
This operator preserves the behavior of HGM-style self-modification
and ensures that the search can proceed even when the archive is still
small.

\paragraph{Reaction-norm Mutation.}
A reaction norm describes how one genotype expresses different phenotypes under different environments~\citep{woltereck1909reaktionsnorm, pigliucci2001phenotypic}.
MGM incorporates this idea and designs reaction-norm mutation $\varPhi_\mathrm{RM}$ by comparing multiple phenotypes of the same genotype across different tasks.  
If the same agent fails, or behaves inconsistently, across multiple tasks, the resulting pattern can provide richer diagnostic evidence of the agent's genuine weakness, rather than accidental task-specific errors.

Formally, $\varPhi_\mathrm{RM}$ becomes available for \(a_i\) when it has accumulated enough $\varphi(a_i,\tau)$ trajectories
\begin{equation}
\label{eq:RM-req}
    |S_i| \ge m_{\mathrm{RM}},
\end{equation}
and there exist at least two trajectories, with a failed one as the target \(\tau_\mathrm{t}\in S_i\). The reference $\tau_\mathrm{r}$ may be any other trajectory in $S_i$, preferably a failed one if available. 
The evidence
\begin{equation}
\begin{aligned}
E_{\mathrm{RM}}(a_i,\tau_\mathrm{t},\tau_\mathrm{r})
= \{&
    \big(\varphi(a_i,\tau_\mathrm{t}), r(a_i,\tau_\mathrm{t})\big),\big(\varphi(a_i,\tau_\mathrm{r}), r(a_i,\tau_\mathrm{r})\big)
   \}.
\end{aligned}
\end{equation}
is given to the self-modification process, where the agent is asked to identify a recurring or contrastive behavioral pattern shared by the provided trajectories and to implement a general improvement
\begin{equation}
    a' \leftarrow \varPhi_{\mathrm{RM}}(a_i,E_{\mathrm{RM}}).
\end{equation}

\paragraph{Cross-lineage Hybridization.}
Cross-lineage hybridization compares different genotypes under the same task environment. 
It is available when two nodes have attempted at least one common target task, and the task is not already solved by both:
\begin{equation}
\label{eq:CH-req}
\begin{aligned}
&\exists j\ne i,\ \exists \tau_\mathrm{t} \in S_i\cap S_j
&\quad \text{s.t.}\quad 
\neg\big(r(a_i,\tau_\mathrm{t} )=1 \land r(a_j,\tau_\mathrm{t} )=1\big).
\end{aligned}
\end{equation}
MGM designates one failed agent as the target $a_\mathrm{t}$ to improve.
If the reference agent $a_\mathrm{r}$ fails likewise, MGM uses the comparison to identify complementary failure modes; if $a_\mathrm{r}$ solved $\tau_\mathrm{t}$, differences in their genotypes are used as corrective signals to guide $a_\mathrm{t}$'s self-modification.
The evidence is
\begin{equation}
\begin{aligned}
E_{\mathrm{CH}}(a_\mathrm{t}, a_\mathrm{r},\tau_\mathrm{t} )
= \{&
(\varphi(a_\mathrm{t},\tau_\mathrm{t} ), r(a_\mathrm{t},\tau_\mathrm{t} )),
(\varphi(a_\mathrm{r},\tau_\mathrm{t} ), r(a_\mathrm{r},\tau_\mathrm{t} ))
\}.
\end{aligned}
\end{equation}
where \(a_\mathrm{t}\) is the target agent and \(a_\mathrm{r}\) is the reference agent.
The child is attached to the primary lineage:
\begin{equation}
    a' \leftarrow \varPhi_{\mathrm{CH}}(a_\mathrm{t},E_{\mathrm{CH}}).
\end{equation}

This hybridization operation is a diagnostic process. MGM does not splice source files from one agent into another. 
Instead, it asks the target agent itself to extract a transferable behavioral trait from the reference trajectory and adapt that trait to the target agent's own codebase.
This encourages the failing lineage to derive and inherit a genuine improvement rather than task-specific behaviors.

\subsection{Sampling Tasks and Operators}

MGM additionally maintains a global pool
\begin{equation}
    \mathcal{P}_t=\bigcup_{i\in \mathcal{V}_t} F_i ,
\end{equation}
which stores tasks that have exposed failures in any previously evaluated agent. 
This pool is not a separate evaluation benchmark and
does not introduce extra $\varphi$-evaluations. 
It is maintained to control how future evaluation tasks are sampled. 
When selecting a new task for an agent, MGM samples from tasks not yet attempted by that agent, but
assigns a predetermined weight to tasks in $\mathcal{P}_t$:
\begin{equation}
    w_i(\tau)=
    \begin{cases}
    \beta_{\mathrm{fail}}, & \tau\in \mathcal{P}_t,\\
    1, & \tau\notin \mathcal{P}_t,
    \end{cases}
    \qquad
    \tau\notin S_i ,
\end{equation}
where $\beta_{\mathrm{fail}}$ is the failed-pool boost.
This task sampling design concentrates
evaluation on tasks that are known to reveal weaknesses in at least one lineage, increasing the diagnostic value of each $\varphi$-evaluation. 
It also deliberately creates overlap across lineages. Since $\varPhi_\textrm{CH}$ requires agents to have attempted a shared task, the pool increases the availability of controlled cross-lineage comparisons without spending additional evaluation budget, making transferable behavioral traits easier to discover.

For strategy selection, MGM inherits HGM's Thompson-sampling policy $\pi$, which at each step chooses between initiating a $\varphi$-evaluation for an existing node or a $\varPhi$-expansion of the evolution tree, using the clade-level and node-level Beta posteriors defined in Section~\ref{sec:background}.
The difference lies in how an expansion is executed. 
MGM partitions the single $\varPhi$ operator that HGM and DGM use ($\varPhi_{\mathrm{CM}}$) into
three sub-operators $\{\varPhi_{\mathrm{CM}}, \varPhi_{\mathrm{RM}}, \varPhi_{\mathrm{CH}}\}$.
When $\pi$ selects a $\varPhi$-expansion for parent $a_i$, MGM first constructs the set of eligible operators
$\Omega_i\subseteq\{\varPhi_{\mathrm{CM}},\varPhi_{\mathrm{RM}},\varPhi_{\mathrm{CH}}\}$
from the archive.
\(\varPhi_{\mathrm{CM}}\) is eligible whenever the selected agent has at least one failed task,
i.e., $F_i\neq\varnothing$.
\(\varPhi_{\mathrm{RM}}\) becomes eligible if the agent has been evaluated on at least $m_{\mathrm{RM}}$ distinct tasks (Equation~\ref{eq:RM-req}), and there exist trajectories for two distinct tasks \(\tau_\mathrm{t},\tau_\mathrm{r}\in S_i\) with at least one failure. 
\(\varPhi_{\mathrm{CH}}\) becomes eligible if agents from two lineages have attempted a shared task \(\tau_\mathrm{t}\) that is not solved by both (Equation~\ref{eq:CH-req}).

MGM then samples among eligible operators with configurable
weights $\lambda_{\mathrm{CM}},\lambda_{\mathrm{RM}},\lambda_{\mathrm{CH}}$:
\begin{equation}
    \Pr(\sigma\mid i)=
    \frac{\lambda_\sigma}
    {\sum_{\sigma'\in\Omega_i}\lambda_{\sigma'}},
    \qquad \sigma\in\Omega_i.
\end{equation}
The selected operator determines the evidence $E_\sigma$, and the
child is produced by
\begin{equation}
    a' \leftarrow \varPhi_\sigma(a_\mathrm{t},E_\sigma).
\end{equation}
For $\varPhi_{\mathrm{CM}}$ and $\varPhi_{\mathrm{RM}}$, the target agent is the selected parent $a_\mathrm{t}=a_i$.
For $\varPhi_{\mathrm{CH}}$, if exactly one agent solves the shared task, the failing agent
is edited using the successful agent as reference; if both fail, the higher-utility lineage serves as the primary target.
If $\Omega_i=\varnothing$, MGM skips the expansion and $\pi$ allocates another $\varphi$-evaluation instead.

The complete pseudocode of MGM is provided in Appendix~\ref{app:pseudocode}.

\section{Simulations}
\label{sec:setup}
\label{sec:simulation}
\vspace{-0.5ex}

\begin{wrapfigure}{r}{0.45\textwidth}
\vspace{-6ex}
        \centering
        \includegraphics[width=0.78\linewidth]{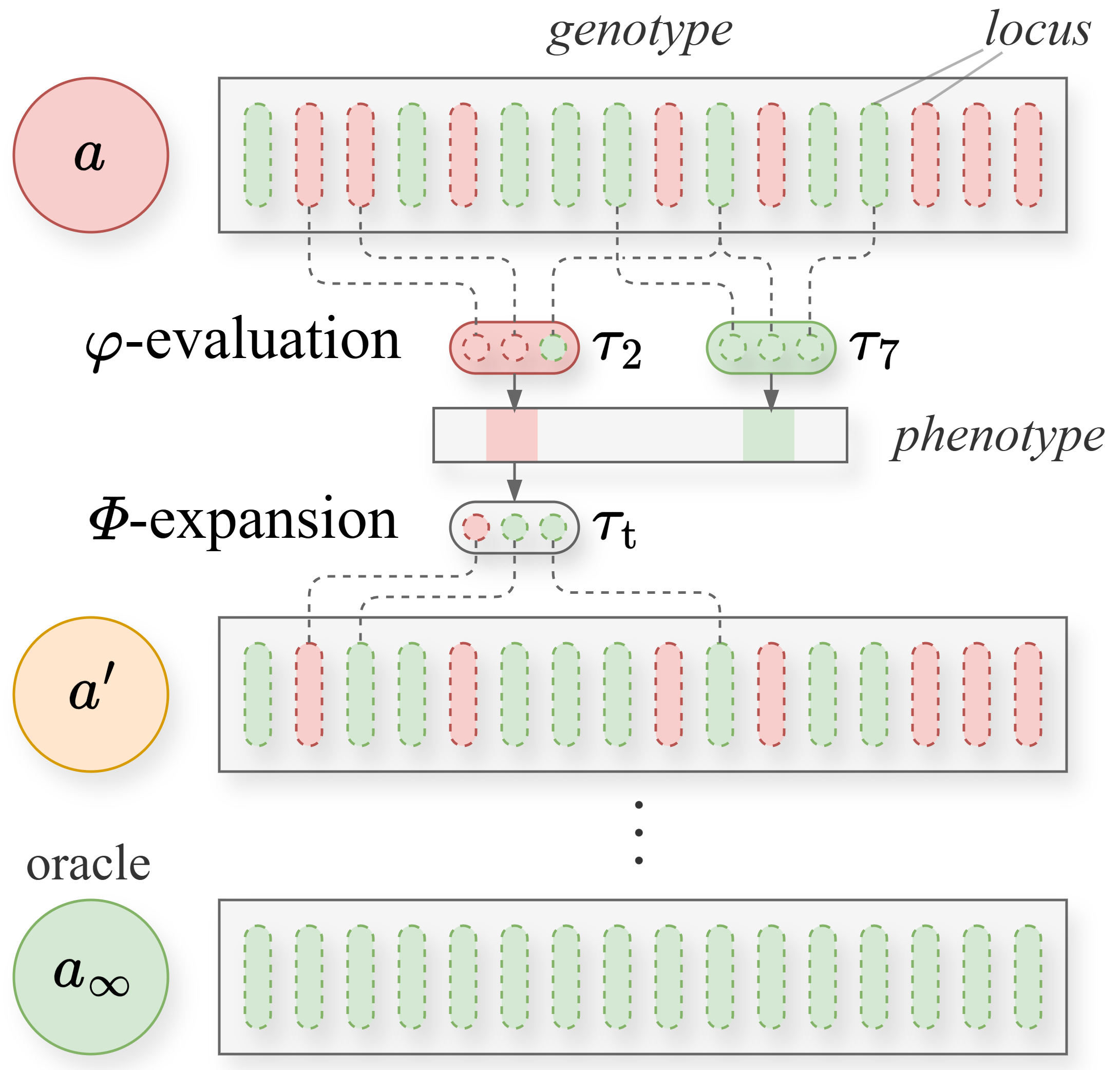}
        \caption{\textbf{Additive fitness landscape model.}
        Each agent $a$ carries a binary genotype with loci that are either correct or mismatched relative to an oracle $a_\infty$.
        The genotype is not directly observable and can only be examined through $\varphi$-evaluation, where each task $\tau_i$ requires a fixed subset of loci to be all correct for a successful phenotype.
        $\varPhi$-expansion reads phenotype records and modifies the genotype by flipping the examined loci, which may correct mismatched ones or corrupt correct ones under certain probabilities.
        }
        \label{fig:genotype-phenotype}
\vspace{-6ex}
\end{wrapfigure}

To validate MGM's design decisions in isolation of implementation details and benchmark choices, we instantiate controlled surrogate models for methods mentioned in Sections~\ref{sec:background} and~\ref{sec:methodology}. 
We demonstrate that comparative evidence can improve the effective fix probability of self-modification by reducing diagnostic uncertainty, and study how such diagnostic advantage affects performance and efficiency.

\subsection{Additive Fitness Landscape}
\vspace{-0.5ex}

Each agent's genotype is modeled as a binary vector $\mathbf{g}\in\{0,1\}^{L}$ with $L$ loci, each corresponding to a minimum scaffold-level capability or implementation choice.
A fixed oracle genotype $\mathbf{g}^{*}\in\{0,1\}^{L}$ represents the optimal program; without loss of generality, we set $\mathbf{g}^{*}=\mathbf{1}$.
The genotype is not directly exposed; instead, its Hamming distance to the oracle is
\begin{equation}
  d(\mathbf{g}) 
  \;=\; 
  \sum_{\ell=1}^{L} 
  \mathbf{1}\!\left[g_{\ell}\neq g^{*}_{\ell}\right],
\end{equation}
where $d(\mathbf{g})=0$ if and only if $\mathbf{g}=\mathbf{g}^{*}$, serving as a zero-error lower bound.
Each run starts from an initial genotype with $d_{0}$ mismatched loci, which self-improvement must correct to reach the oracle.

The task pool contains $N$ tasks.
Each task $\tau$ examines a subset of loci $R_{\tau}\subseteq[L]$ with $|R_{\tau}|=k$.
The agent solves the task only when all required loci are correct:
\begin{align}
  r(a,\tau)=1
  \quad\Longleftrightarrow\quad
  R_{\tau}\cap M(a)=\varnothing,\\  
  M(a)=\{\ell\in[L]: g_{\ell}(a)\neq g^{*}_{\ell}\}
\end{align}
where $M(a)$
denotes the set of incorrect loci of agent $a$.
Under independent locus sampling, the resulting per-task success probability for an agent at edit distance $d$ is
\begin{equation}
  P(r{=}1\mid d)
  =
  \Bigl(\tfrac{L-d}{L}\Bigr)^{k}.
\end{equation}
At the start of each run, this probability equals $((L-d_{0})/L)^{k}$.
This models coding tasks as requiring multiple scaffold-level capabilities to be simultaneously correct, such as localization, reasoning, editing, and validation.
All parameter values are listed in Table~\ref{tab:simulation_params}.

\subsection{Comparative Evidence as Diagnostic Compression}
\vspace{-0.5ex}

We demonstrate how the comparative operators in MGM can induce a higher effective fix probability than single-trajectory mutation.
A self-modification operator does not observe the hidden incorrect loci directly.
Instead, given diagnostic evidence $E$, it transiently constructs an implicit candidate set $C_{\sigma}(E)\subseteq[L]$ of loci that may explain the observed failure, where $\sigma\in\{\mathrm{CM},\mathrm{RM},\mathrm{CH}\}$.
Suppose that once the editor targets an actually incorrect locus, it repairs it with probability $s\in(0,1]$.
Then the effective fix probability of operator $\sigma$ is
\begin{equation}
  p_{f}^{\sigma}
  =
  s\cdot
  \Pr_{\ell\sim C_{\sigma}(E)}
  [\ell\in M(a)].
\end{equation}
Therefore, the fix probability increases when the evidence yields a candidate set with a higher density of truly incorrect loci.

\(\varPhi_\textrm{CM}\) observes a single failed task $\tau_{t}$.
Since this evidence only implies $R_{\tau_{t}}\cap M(a)\neq\varnothing$, the natural candidate set is $C_{\mathrm{CM}}=R_{\tau_{t}}$.
By contrast, \(\varPhi_\textrm{RM}\) compares multiple trajectories of the same genotype.
When two failures share a recurring scaffold-level defect, the common explanatory region is compressed to $C_{\mathrm{RM}}=R_{\tau_{t}}\cap R_{\tau_{r}}$, which is smaller than $R_{\tau_{t}}$ in expectation.
\(\varPhi_\textrm{CH}\) compares different genotypes on the same task.
When a reference agent succeeds on the task while the target agent fails, the reference trajectory acts as a contrastive control and filters out non-causal task-relevant loci from $R_{\tau}$. 
This gives the following proposition.

\vspace{-0.5ex}
\paragraph{Proposition 1.}
\label{prop1}
Under the aforementioned model and sound comparative evidence, \(\varPhi_\textrm{RM}\) and  \(\varPhi_\textrm{CH}\) have strictly higher effective fix probability than \(\varPhi_\textrm{CM}\):
\begin{equation}
  p_{f}^{\mathrm{RM}}>p_{f}^{\mathrm{CM}},
  \qquad
  p_{f}^{\mathrm{CH}}>p_{f}^{\mathrm{CM}}.
\end{equation}
The full derivation is provided in Appendix~\ref{app:comparative-fix-probability}.
Intuitively, comparative evidence improves self-modification not by making the editor intrinsically stronger, but by reducing diagnostic uncertainty: the editor searches over a smaller and cleaner set of candidate defects.

\vspace{-0.5ex}
\subsection{Monte Carlo Simulation}
\vspace{-0.5ex}

Guided by Proposition~1, we instantiate the diagnostic advantage of comparative evidence through a controllable fix-probability ratio
\begin{equation}
  \rho
  =
  \frac{p_{f}^{\mathrm{RM}}}{p_{f}^{\mathrm{CM}}}
  =
  \frac{p_{f}^{\mathrm{CH}}}{p_{f}^{\mathrm{CM}}}.
\end{equation}
The null setting $\rho=1$ corresponds to the case where comparative evidence provides no additional diagnostic benefit, while $\rho>1$ models increasing levels of diagnostic compression.

The simulation isolates axes on which the three methods differ.
For budget allocation, DGM assigns a fixed $n_{\mathrm{eval}}$ evaluations to every population member per generation, distributing budget uniformly regardless of node quality.
HGM uses adaptive allocation, concentrating evaluations on promising nodes and triggering edits according to its tree-search policy.
DGM and HGM use only clonal mutation $\varPhi_{\mathrm{CM}}$.
MGM follows the same evaluation-expansion framework but additionally employs $\varPhi_{\mathrm{RM}}$ and $\varPhi_{\mathrm{CH}}$ when their availability conditions hold.

To ensure that the comparison isolates the value of diagnostic evidence rather than unequal compute, all edit operators incur the same cost:
\begin{equation}
  c_{\mathrm{CM}}
  =
  c_{\mathrm{RM}}
  =
  c_{\mathrm{CH}}
  =
  c_{\varphi}.
\end{equation}
Each method runs for a fixed total budget of $B$ resource units with $n_{\mathrm{seeds}}$ independent Monte Carlo seeds.
At each seed, a fresh task pool and initial genotype are regenerated.
The minimum edit distance across all active nodes is recorded at evenly spaced budget checkpoints; final performance is reported as mean $\pm$ 95\,\% CI across seeds.
A separate parameter sweep varies the initial edit distance $d_{0}$ and the diagnostic-advantage ratio $\rho$ to test robustness of the ordering.
Detailed configurations for our simulation can be found in Table~\ref{tab:simulation_params}.

\begin{figure*}[t]
    \centering
    \begin{minipage}[t]{0.48\linewidth}
        \centering
        \scriptsize
        \setlength{\tabcolsep}{0pt}
        \begin{tabular}{m{1.5em}  m{0.235\linewidth} m{0.235\linewidth} m{0.235\linewidth} m{0.235\linewidth}}
            \rotatebox{90}{ $d_0 = 80$} &
            \includegraphics[width=\linewidth]{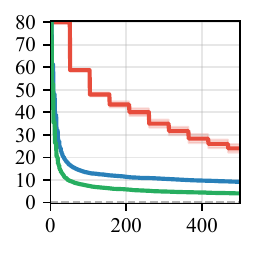} &
            \includegraphics[width=\linewidth]{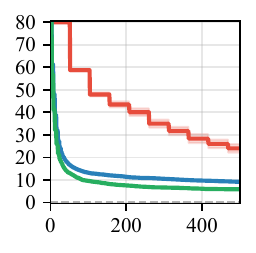} &
            \includegraphics[width=\linewidth]{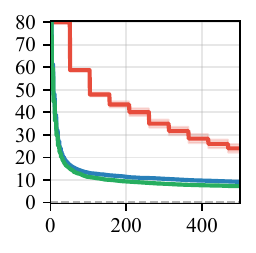} &
            \includegraphics[width=\linewidth]{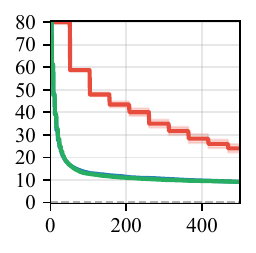} \\
            \rotatebox{90}{$d_0 = 40$} &
            \includegraphics[width=\linewidth]{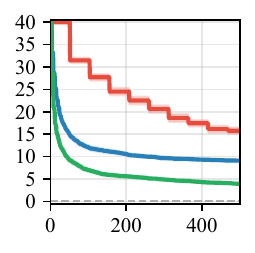} &
            \includegraphics[width=\linewidth]{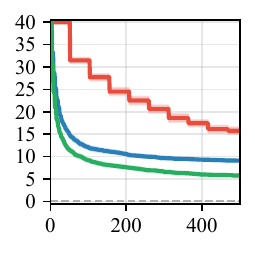} &
            \includegraphics[width=\linewidth]{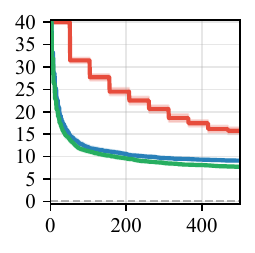} &
            \includegraphics[width=\linewidth]{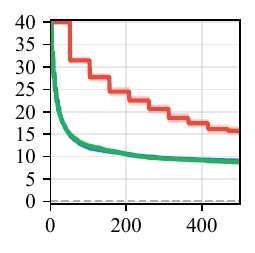} \\
            \rotatebox{90}{$d_0 = 20$} &
            \includegraphics[width=\linewidth]{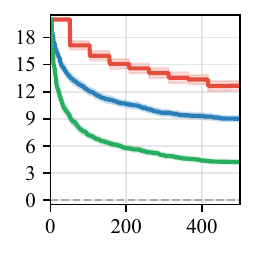} &
            \includegraphics[width=\linewidth]{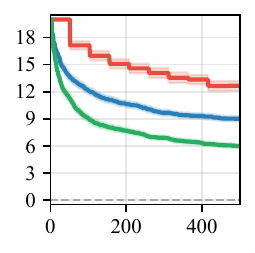} &
            \includegraphics[width=\linewidth]{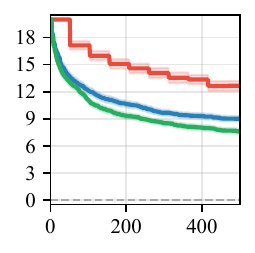} &
            \includegraphics[width=\linewidth]{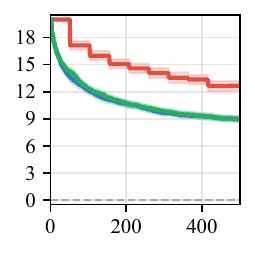} \\
            \rotatebox{90}{$d_0 = 10$} &
            \includegraphics[width=\linewidth]{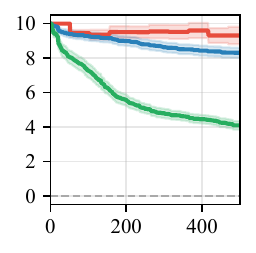} &
            \includegraphics[width=\linewidth]{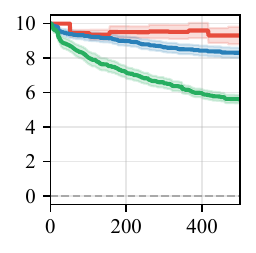} &
            \includegraphics[width=\linewidth]{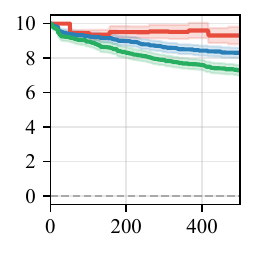} &
            \includegraphics[width=\linewidth]{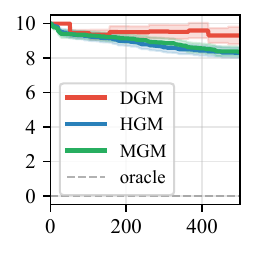} \\
            & \makebox[\linewidth][c]{$\rho= 2.0$} &
              \makebox[\linewidth][c]{$\rho= 1.5$} &
              \makebox[\linewidth][c]{$\rho= 1.2$} &
              \makebox[\linewidth][c]{$\rho= 1.0$} \\
        \end{tabular}
        \vspace{-1.5ex}
        \caption{\textbf{Simulated performance evolution over cumulative budget spent.} Results compared across task difficulty and edit effectiveness.
        Rows vary the initial edit distance $d_0$; columns vary the fix-probability advantage ratio $\rho$.
        Each panel shows the edit distance results (lower is better) averaged across random seeds with 95\,\% CIs. 
        Dashed lines mark oracle optima.
        }
        \label{fig:sim-sweep-evolve}
    \end{minipage}
    \hfill
    \begin{minipage}[t]{0.48\linewidth}
        \centering
        \scriptsize
        \setlength{\tabcolsep}{0pt}
        \begin{tabular}{m{1.5em}  m{0.235\linewidth} m{0.235\linewidth} m{0.235\linewidth} m{0.235\linewidth}}
            \rotatebox{90}{ $d_0 = 80$} &
            \includegraphics[width=\linewidth]{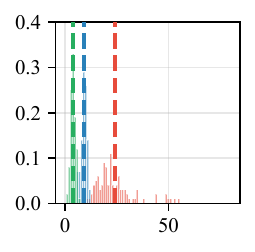} &
            \includegraphics[width=\linewidth]{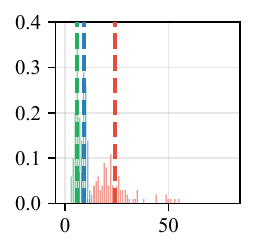} &
            \includegraphics[width=\linewidth]{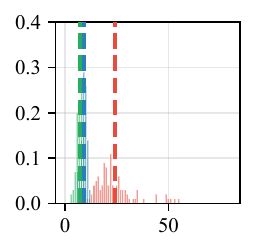} &
            \includegraphics[width=\linewidth]{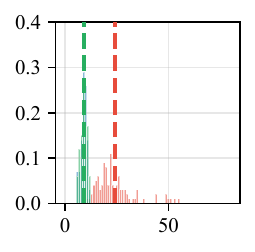} \\
            \rotatebox{90}{$d_0 = 40$} &
            \includegraphics[width=\linewidth]{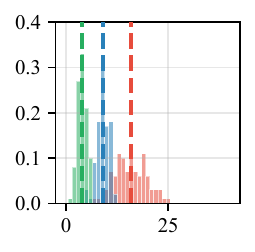} &
            \includegraphics[width=\linewidth]{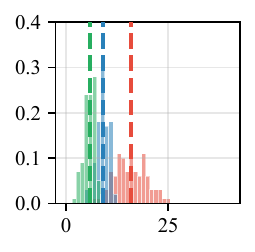} &
            \includegraphics[width=\linewidth]{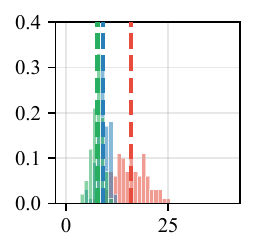} &
            \includegraphics[width=\linewidth]{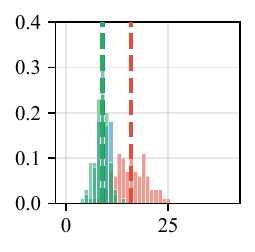} \\
            \rotatebox{90}{$d_0 = 20$} &
            \includegraphics[width=\linewidth]{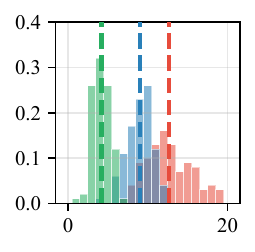} &
            \includegraphics[width=\linewidth]{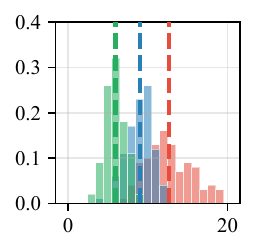} &
            \includegraphics[width=\linewidth]{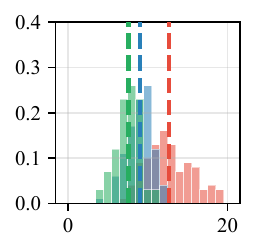} &
            \includegraphics[width=\linewidth]{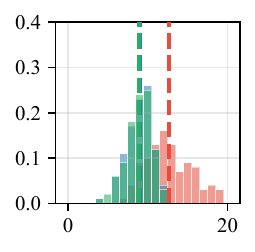} \\
            \rotatebox{90}{$d_0 = 10$} &
            \includegraphics[width=\linewidth]{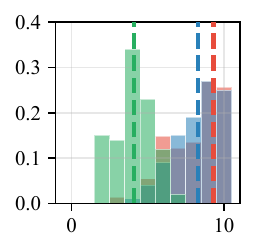} &
            \includegraphics[width=\linewidth]{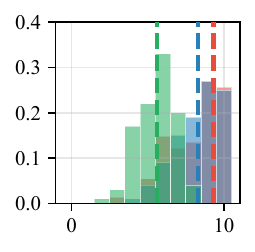} &
            \includegraphics[width=\linewidth]{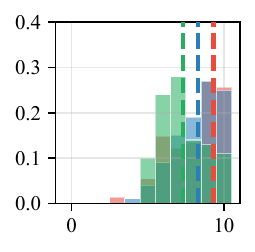} &
            \includegraphics[width=\linewidth]{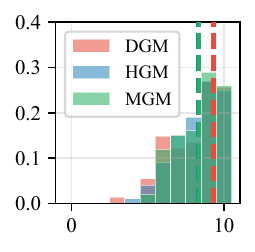} \\
            & \makebox[\linewidth][c]{$\rho= 2.0$} &
              \makebox[\linewidth][c]{$\rho= 1.5$} &
              \makebox[\linewidth][c]{$\rho= 1.2$} &
              \makebox[\linewidth][c]{$\rho= 1.0$} \\
        \end{tabular}
        \vspace{-1.5ex}
        \caption{\textbf{Simulated final performance distribution.} Results compared across task difficulty and edit effectiveness.
        Rows vary the initial edit distance $d_0$; columns vary the fix-probability advantage ratio $\rho$.
        Each panel shows the distribution of final edit distance results (lower is better) across all random seeds, with dashed lines marking per-method means.
        }
        \label{fig:sim-sweep-final}
    \end{minipage}
\end{figure*}

The parameter sensitivity results in Figure~\ref{fig:sim-sweep-evolve} and Figure~\ref{fig:sim-sweep-final} test the robustness of MGM across a broad range of initial difficulties $d_{0}$ and operator-quality settings $\rho$.
MGM consistently outperforms other baselines in both final performance and convergence speed across all $d_{0}$ and $\rho>1.0$ settings, demonstrating reliable gains across both easier and harder initial conditions.
The null case $\rho=1$ suggests that when comparative operators provide no fix-quality advantage, MGM collapses back to HGM-like behavior.
This confirms that the simulated gain is caused directly by the diagnostic-quality advantage formalized in Proposition~1 in Section~\ref{prop1} and Appendix~\ref{app:comparative-fix-probability}.
As $\rho$ increases, MGM's advantage grows steadily, and this trend is especially pronounced at smaller $d_{0}$, where fewer but more precise self-improvements are required and higher-quality diagnostic signals translate into clearer performance gains.
Figure~\ref{fig:sim-sweep-final} additionally shows the distribution of final performance across seeds, where MGM achieves the lowest mean and tightest spread, while the broader distributions of other baselines reflect less stable budget allocation.

\section{Experiments}
We evaluate MGM on challenging software-engineering benchmarks to answer three research questions: 
Does MGM exhibit better performance and efficiency? 
Can agents evolved by MGM generalize to other challenging tasks or models? 
What is the contribution of each key component in MGM? 

Our experiments span several challenging and representative benchmarks, including SWE-bench Verified~\citep{SWEbench}, SWE-bench Pro~\citep{SWEbench-Pro}, SWE-bench Multilingual~\citep{SWEbench-Multilingual,yang2025swesmith}, and Polyglot~\citep{polyglot}. 
In all experiments, agents are not given access to private test cases or test results during evolution. 
To ensure fair comparisons and control computational cost, unless otherwise specified, all results reported in this section are evaluated on the same two 60-task subsets of SWE-bench Verified and Polyglot as \citet{HGM} and \citet{zhang2026darwin}.




\subsection{Performance}\label{highergreater}

To assess whether MGM exhibits greater self-improvement potential than HGM, following the setting of \citet{zhang2026darwin, HGM}, we evaluate both methods on SWE-bench Verified and Polyglot under an identical computational budget of 200 evaluations. We adopt Qwen3.6-35B-A3B~\citep{qwen36_35b_a3b} as the backbone LLM. To ensure a fair comparison, both methods start from the same ancestor agent. We report the accuracy of the best-belief agent evolved by each method.

\begin{table*}[t]
\centering
\small
\renewcommand{\arraystretch}{1.15}
\addtolength{\tabcolsep}{-.2pt} 
    \begin{tabular}{rccccccccc}
    \toprule
    & \multicolumn{3}{c}{SWE-bench Verified-60}
    & \multicolumn{3}{c}{Polyglot-60} 
    & \multicolumn{3}{c}{Avg.} \\
    \cmidrule(lr){2-4} \cmidrule(lr){5-7} \cmidrule(lr){8-10}
    Agent & Initial & HGM & MGM & Initial & HGM & MGM & Initial & HGM & MGM \\
    \midrule

    
    Accuracy
    & \cAcc{68.3}{68.3}
    & \cAcc{73.3}{73.3\textsuperscript{+5.0}}
    & \cAcc{78.3}{\textbf{78.3\textsuperscript{+10.0}}}
    & \cAcc{50.8}{50.8}
    & \cAcc{77.9}{77.9\textsuperscript{+27.1}}
    & \cAcc{93.2}{\textbf{93.2\textsuperscript{+42.4}}}
    & \cAcc{59.6}{59.6}
    & \cAcc{75.6}{75.6\textsuperscript{+16.0}}
    & \cAcc{85.8}{\textbf{85.8\textsuperscript{+26.2}}}
    \\
    
    \% Impr.
    & --
    & \cImpr{7.3}{$\uparrow\,$7.3\%}
    & \cImpr{14.6}{\textbf{$\uparrow\,$14.6\%}}
    & --
    & \cImpr{53.3}{$\uparrow\,$53.3\%}
    & \cImpr{83.5}{\textbf{$\uparrow\,$83.5\%}}
    & --
    & \cImpr{26.8}{$\uparrow\,$26.8\%}
    & \cImpr{44.0}{\textbf{$\uparrow\,$44.0\%}} \\

    Time
    & --
    & \cTime{93.02}{93.02 h}
    & \cTime{96.11}{96.11 h}
    & --
    & \cTime{44.20}{44.20 h}
    & \cTime{40.14}{\textbf{40.14 h}}
    & --
    & \cTime{68.61}{68.61 h}
    & \cTime{68.12}{\textbf{68.12 h}}\\
    \bottomrule
    \end{tabular}
\caption{
\textbf{Performance of coding agents evolved on SWE-bench Verified and Polyglot.} Results evolved using Qwen3.6-35B-A3B after 200 $\varphi$-evaluations and 24 $\varPhi$-expansions. 
For each benchmark, HGM and MGM start from the same initial scaffold.
Superscripts in accuracy denote absolute percentage-point improvements over corresponding initial agents.
Time reported as CPU wall-clock time with 8$\times$NVIDIA H100 GPUs.
}
\vspace{1ex}
\label{tab:performance}
\end{table*}

\begin{wrapstuff}[l, top=4, type=figure, width=0.45\linewidth, abovesep=0ex, belowsep=0ex]
    \centering
        \vspace{-0.25ex}
    \includegraphics[width=0.85\linewidth]{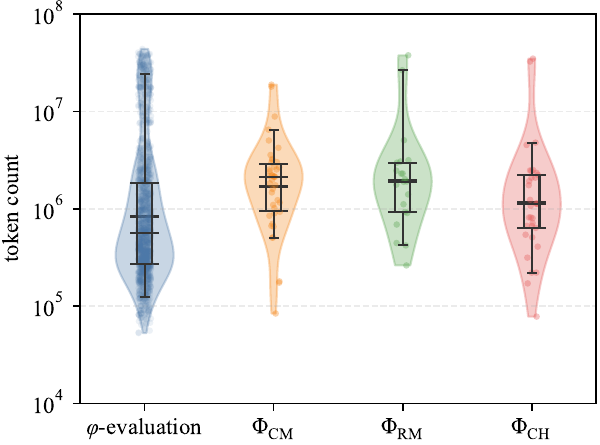}
        \vspace{-1ex}
    \caption[Token costs of each operators for HGM and MGM evolved on Polyglot]{\textbf{Token costs of each operators for HGM and MGM evolved on Polyglot.} The distribution of token costs for all evaluation and expansion operations lies in similar order of magnitude.}
    \label{fig:token_comparison}
\end{wrapstuff}

As shown in Table~\ref{tab:performance}, MGM consistently outperforms HGM. On SWE-bench Verified, both methods start from the same initial accuracy of 68.3\%, which HGM improves to 73.3\%, and MGM improves to 78.3\%. 
On Polyglot, starting from the same 50.8\% agent, HGM improves to 77.9\%, while MGM reaches 93.2\%. 
These result shows that MGM's comparative self-modification operators are effective on both standalone coding tasks as in Polyglot, and also repository-level software-engineering tasks, where failures often involve incompetency in localization, environment understanding, and repair workflows.
Since both methods use exactly the same number of evaluation and expansion operations, this performance gap cannot be attributed to a larger search budget. Instead, it supports our central hypothesis that reusing archived trajectories through reaction-norm mutation and cross-lineage hybridization provides more informative diagnostic evidence than conditioning each self-modification step on a single failure trajectory.
Evaluation on the full 225-task Polyglot benchmark (Figure~\ref{fig:polyglot} and Appendix~\ref{app:full_polyglot}) further confirms the same results.

We further analyze the token consumption of each step in the experiments. Figure~\ref{fig:token_comparison} shows that all evolutionary strategies used by HGM and MGM have comparable average token costs, indicating that the observed performance gains are not achieved simply by increased token expenditure.

\subsection{Generalization}\label{Generalization}

A key question for self-improving coding agents is whether the evolved improvements remain useful beyond the exact setting in which evolution is performed~\citep{zhang2026darwin,HGM}.
Since MGM modifies the agent scaffold rather than the parameters of the underlying LLM, a successful evolution method should ideally discover reusable workflow-level improvements that transfer across both benchmarks and backbone models~\citep{ADAS}.
We therefore evaluate generalization from two complementary perspectives: cross-benchmark generalization and cross-model scaffold transfer.

\paragraph{Cross-benchmark generalization.}
We first evaluate whether scaffolds evolved on Polyglot-60 can transfer to different software-engineering settings without further self-improvement. After evolution, we freeze the evolved scaffolds and evaluate them on subsets of SWE-bench Pro and SWE-bench Multilingual~\citep{SWEbench-Multilingual,yang2025swesmith}. We report the tasks used in Appendix~\ref{app:benchmark_details}. We choose these two benchmarks because they test complementary forms of out-of-distribution generalization. SWE-bench Pro provides a more challenging repository-level software-engineering setting, where solving a task often requires understanding larger codebases, localizing the relevant files, and producing robust patches.
SWE-bench Multilingual further stresses the language generality of the evolved scaffold by covering repositories implemented in diverse programming languages.

\begin{table*}[t]
\centering
\small
\renewcommand{\arraystretch}{1.15}
\addtolength{\tabcolsep}{0pt} 
\begin{tabular}{rccccccccc}
\toprule
& \multicolumn{3}{c}{SWE-bench Pro}
& \multicolumn{3}{c}{SWE-bench Multilingual}
& \multicolumn{3}{c}{Avg.}\\
\cmidrule(lr){2-4} \cmidrule(lr){5-7} \cmidrule(lr){8-10}
Agent & Initial & HGM & MGM  & Initial & HGM & MGM  & Initial & HGM & MGM \\
\midrule
Accuracy
& \cAcc{16.7}{16.7} 
& \cAcc{13.3}{13.3\textsuperscript{-3.4}} 
& \cAcc{26.7}{\textbf{26.7\textsuperscript{+10.0}}}
& \cAcc{41.7}{41.7} 
& \cAcc{43.3}{43.3\textsuperscript{+1.6}} 
& \cAcc{55.0}{\textbf{55.0\textsuperscript{+13.3}}}
& \cAcc{29.2}{29.2} 
& \cAcc{28.3}{28.3\textsuperscript{-0.9}} 
& \cAcc{40.9}{\textbf{40.9\textsuperscript{+11.7}}}\\

\% Impr.
& -- 
& \cImpr{-20.4}{$\downarrow\,$20.4\%} 
& \cImpr{59.9}{\textbf{$\uparrow\,$59.9\%}}
& -- 
& \cImpr{3.8}{$\uparrow\,$3.8\%} 
& \cImpr{31.9}{\textbf{$\uparrow\,$31.9\%}}
& -- 
& \cImpr{-3.1}{$\downarrow\,$3.1\%} 
& \cImpr{40.1}{\textbf{$\uparrow\,$40.1\%}}\\
\bottomrule
\end{tabular}
\vspace{-0.8ex}
\caption{\textbf{Cross-benchmark generalization from Polyglot to SWE-bench Pro and SWE-bench Multilingual.} All agents evolved on Polyglot are evaluated zero-shot on held-out SWE-bench variants, using Qwen3.6-35B-A3B. 
Superscripts in accuracy denote absolute percentage-point improvements over corresponding initial agents.}
\label{tab:cross-benchmark-generalization}
\vspace{-0.2ex}
\end{table*}

As shown in Table~\ref{tab:cross-benchmark-generalization}, the initial scaffold obtains 16.7\% accuracy on SWE-bench Pro and 41.7\% on SWE-bench Multilingual. The HGM-evolved scaffold shows limited transfer: it decreases to 13.3\% on SWE-bench Pro, corresponding to a $-3.4$ percentage-point change, and slightly improves to 43.3\% on SWE-bench Multilingual, corresponding to a $+1.6$ percentage-point gain.
In contrast, MGM achieves positive transfer on both target benchmarks. It reaches 26.7\% on SWE-bench Pro, giving a $+10.0$ percentage-point improvement over the initial scaffold, and 55.0\% on SWE-bench Multilingual, giving a $+13.3$ percentage-point improvement.

These results suggest that MGM's comparative self-modification operators help discover scaffold changes that remain useful beyond the source benchmark. The gains on both SWE-bench Pro and SWE-bench Multilingual indicate that MGM does not overfit to Polyglot-specific failure patterns and is more likely to acquire reusable software-engineering workflows that transfer across benchmark distributions and programming-language settings.

\vspace{-0.5ex}
\paragraph{Cross-model generalization.}
We further evaluate whether the evolved scaffolds remain effective when paired with different backbone LLMs. In this experiment, the scaffolds are evolved on SWE-bench Verified-60 using Qwen3.6-35B-A3B. We then freeze the evolved scaffold and replace only the inference backbone with DeepSeek-V4-Flash and DeepSeek-V4-Pro~\citep{deepseekv4}. This protocol isolates whether the improvement is encoded in the scaffold itself, rather than being specific to the original Qwen3.6 backbone.

\begin{table*}[t]
\vspace{-1ex}
\centering
\small
\renewcommand{\arraystretch}{1.15}
\addtolength{\tabcolsep}{-4.55pt} 
\begin{tabular}{rcccccccccccc}
\toprule
& \multicolumn{3}{c}{\textit{Evolved}}
& \multicolumn{9}{c}{\textit{Transferred}} \\
\cmidrule(lr){2-4} \cmidrule(lr){5-13}
& \multicolumn{3}{c}{Qwen3.6-35B-A3B}
& \multicolumn{3}{c}{DeepSeek-V4-Flash}
& \multicolumn{3}{c}{DeepSeek-V4-Pro}
& \multicolumn{3}{c}{Avg.}\\
\cmidrule(lr){2-4} \cmidrule(lr){5-7} \cmidrule(lr){8-10} \cmidrule(lr){11-13}
Agent & Initial & HGM & MGM & Initial & HGM & MGM & Initial & HGM & MGM & Initial & HGM & MGM \\
\midrule
Acc.
& \cAcc{68.3}{68.3} 
& \cAcc{73.3}{73.3\textsuperscript{+5.0}} 
& \cAcc{78.3}{\textbf{78.3\textsuperscript{+10.0}}}
& \cAcc{50.0}{50.0} 
& \cAcc{60.0}{60.0\textsuperscript{+10.0}} 
& \cAcc{66.7}{\textbf{66.7\textsuperscript{+16.7}}}
& \cAcc{45.0}{45.0} 
& \cAcc{70.0}{70.0\textsuperscript{+25.0}} 
& \cAcc{75.0}{\textbf{75.0\textsuperscript{+30.0}}}
& \cAcc{47.5}{47.5} 
& \cAcc{65.0}{65.0\textsuperscript{+17.5}} 
& \cAcc{70.8}{\textbf{70.8\textsuperscript{+23.3}}}\\

\% Impr.
& -- 
& \cImpr{7.3}{$\uparrow\,$7.3\%} 
& \cImpr{14.6}{\textbf{$\uparrow\,$14.6\%}}
& -- 
& \cImpr{20.0}{$\uparrow\,$20.0\%} 
& \cImpr{33.3}{\textbf{$\uparrow\,$33.3\%}}
& -- 
& \cImpr{55.6}{$\uparrow\,$55.6\%} 
& \cImpr{66.7}{\textbf{$\uparrow\,$66.7\%}}
& -- 
& \cImpr{36.8}{$\uparrow\,$36.8\%} 
& \cImpr{49.1}{\textbf{$\uparrow\,$49.1\%}}\\
\bottomrule
\end{tabular}
\vspace{-0.8ex}
\caption{
\textbf{Cross-model transfer on SWE-bench Verified-60.}
The Qwen3.6-35B-A3B block reports the original evolved agents, while the DeepSeek blocks report transferred performance by using the scaffolds evolved on Qwen3.6-35B-A3B and evaluating with LLM backbone replaced.
Superscripts in accuracy denote absolute percentage-point improvements over corresponding initial agents.
}
\label{tab:cross_model_transfer}
\vspace{-0.5ex}
\end{table*}

Table~\ref{tab:cross_model_transfer} shows that both HGM and MGM produce scaffolds that transfer across models, but MGM transfers more strongly and consistently. Under the original Qwen3.6-35B-A3B backbone, HGM improves the initial scaffold from 68.3\% to 73.3\%, whereas MGM improves it to 78.3\%, giving MGM a +5.0 percentage-point advantage over HGM. After transferring the same scaffolds to DeepSeek-V4-Flash, HGM reaches 60.0\%, while MGM further improves to 66.7\%. On DeepSeek-V4-Pro, HGM reaches 70.0\%, whereas MGM reaches 75.0\%. 
Averaged over the two transferred DeepSeek backbones, MGM achieves 70.8\% accuracy, compared with 65.0\% for HGM and 47.5\% for the initial scaffold. 
The result confirms that MGM does not simply tune the scaffold to the quirks of a single foundation model. Instead, its evolved changes remain beneficial when the same scaffold is executed by substantially different inference backbones.

Beyond the subset-level transfer results above, we further evaluate whether the scaffold evolved on Polyglot with Qwen3.6-35B-A3B remains effective when paired with a stronger inference backbone. Specifically, we freeze the MGM-evolved Polyglot scaffold, replace only the backbone model with DeepSeek-V4-Pro, and evaluate it on the complete 225-task Polyglot benchmark. As Figure~\ref{fig:polyglot} shows, this transferred configuration achieves \textbf{\textit{96.89\%}} accuracy, further advancing the frontier result achieved with Qwen above. 
This result provides additional evidence that MGM truely discovers reusable workflow-level improvements that can be further amplified by stronger foundation models.

Together, the cross-benchmark and cross-model results demonstrate that MGM helps improve not only in-domain performance, but also the generalizability of the evolved agent scaffold across benchmarks and models. 
This points to a potential path for scalable self-improving agent development, where scaffolds evolved on smaller datasets and cheaper backbones can be reused on stronger models.
Appendices~\ref{hybridhelp} and~\ref{general_skills} further provide qualitative analysis demonstrating how these gains actually correspond to reusable workflow-level skills.

\begin{wrapstuff}[r, top=0, type=table, width=0.49\linewidth, abovesep=0ex, belowsep=0ex]
    \centering
    \small
    \renewcommand{\arraystretch}{1}
    \addtolength{\tabcolsep}{-2.2pt} 
    
    \begin{tabular}{rcccc}
    \toprule
        & \multicolumn{4}{c}{Polyglot-60} \\
        \cmidrule(lr){2-5} 
    Agent
    & Initial
    & MGM
    & w/o $\varPhi_\mathrm{RM}$
    & w/o $\varPhi_\mathrm{CH}$ \\
    \midrule
    
        
    Accuracy
    & \cAcc{50.8}{50.8}
    & \cAcc{93.2}{\textbf{93.2\textsuperscript{+42.4}}}
    & \cAcc{79.7}{79.7\textsuperscript{+28.9}}
    & \cAcc{74.6}{74.6\textsuperscript{+23.8}}
    \\
    
    \% Impr.
    & --
    & \cImpr{83.5}{\textbf{$\uparrow\,$83.5\%}} 
    & \cImpr{56.9}{$\uparrow\,$56.9\%}
    & \cImpr{46.9}{$\uparrow\,$46.9\%}
    \\
    
    Time
    & --
    & \cTime{40.14}{40.14 h}
    & \cTime{43.01}{43.01 h}
    & \cTime{39.95}{39.95 h}
    \\
    \bottomrule
    \end{tabular}
    \vspace{-0.5ex}
    \caption{\textbf{Ablation study.}
    Superscripts in accuracy denote absolute percentage-point improvements over corresponding initial agents.
    Time reported as CPU wall-clock time with 8$\times$NVIDIA H100 GPUs.
    }
    \label{tab:mgm_ablation}
\end{wrapstuff}
\subsection{Ablation Study}\label{ablationsec}
\vspace{-0.5ex}

To understand the contribution of each component in MGM, we conduct an ablation study on Polyglot-60. All variants are evaluated under the same computational budget of 200 evaluations, with two parallel workers enabled during evolution. We use Qwen3.6-35B-A3B as the backbone LLM for all settings. Each variant starts from the same initial ancestor agent, which achieves 50.8\% accuracy on Polyglot-60. We compare the full MGM with two ablated variants: MGM without $\varPhi_\mathrm{RM}$ and without $\varPhi_\mathrm{CH}$. Appendix~\ref{app:evolution_trees} visualizes the corresponding evolution trees for the full model, HGM, and ablations.
As shown in Table~\ref{tab:mgm_ablation}, the full MGM achieves the best performance, and removing $\varPhi_\mathrm{RM}$ and $\varPhi_\mathrm{CH}$  leads to a clear performance degradation. 
This suggests that $\varPhi_\mathrm{RM}$  plays an important role in guiding the self-improvement process toward more promising agents, and that $\varPhi_\mathrm{CH}$ is even more critical for effective self-improvement, likely because it helps preserve and reuse useful evolutionary information across iterations. 
The ablated variants have comparable training hours, which also confirms that the gains of MGM come from the combined effect of its key components rather than from differences in computational cost.

\vspace{-0.5ex}
\section{Related Work}
\vspace{-0.5ex}

LLM agent systems and automated agent-design methods study how prompts, tools, and workflows can be optimized as executable scaffolds~\citep{yao2023react, schick2023toolformer, wu2023autogenenablingnextgenllm, yang2024sweagent, zhang2026hyperagents, gao2026survey}.
\citet{ADAS} and \citet{hong2024metagpt} show that meta-agents can iteratively propose new agent designs from an archive of prior discoveries, yielding agents that transfer across tasks and models. 
This line of work establishes agent scaffolding as a searchable program space, with a focus on discovering new agent designs.

Self-improving coding agents instantiate this idea in an inherited self-modification setting~\citep{madaan2023selfrefine, shinn2024reflexion, livesweagent}.
\citet{robeyns2025sica} show that a coding agent can evaluate itself, edit its own codebase, and improve over iterations. 
\citet{zhang2026darwin} extend this process into open-ended evolution by maintaining an archive of self-modified agents, while \citet{HGM} improve archive expansion by estimating the future self-improvement potential of clades. 
MGM follows this archive-based self-improvement setting, but targets a complementary bottleneck: improving the evidence used for each expansion.
By reusing existing trajectories across tasks and lineages, MGM improves each self-modification step  without requiring additional evaluations.

Software-engineering benchmarks such as SWE-bench~\citep{SWEbench}, SWE-bench Pro~\citep{SWEbench-Pro}, SWE-bench Multilingual~\citep{SWEbench-Multilingual,yang2025swesmith}, and Polyglot~\citep{polyglot} evaluate repository-level repair, long-horizon reasoning, and cross-language coding ability, testing whether coding agents can make workflow-level improvements. 

A detailed discussion of additional related work on self-evolving agents, runtime adaptation methods, and datasets is provided in Appendix~\ref{app:related-work-details}.

\vspace{-0.5ex}
\section{Conclusion}
\vspace{-0.5ex}

We introduce the Mendel G\"odel Machine (MGM), an archive-based self-improving coding-agent framework that conditions self-modification on comparative evidence already present in evaluation trajectories.
Beyond clonal mutation, MGM adds reaction-norm mutation and cross-lineage hybridization, which reuse archived phenotypes across tasks and lineages without additional task evaluations.
Under an additive fitness landscape, theory and controlled simulations show that richer diagnostic evidence can raise effective fix probability and accelerate convergence relative to single-trajectory baselines.
Experiments on SWE-bench and Polyglot confirm MGM's consistent gains over HGM in performance and efficiency under a matched budget.
Ablations show that both comparative operators contribute to these gains.
Held-out evaluations further suggest that the evolved scaffolds transfer across benchmarks and backbone LLMs, indicating that richer comparative conditioning can discover reusable workflow-level improvements rather than narrow task-specific patches.
Within the constraints of scaffold‑level evolution under sandboxed, budgeted archive search, MGM potentially points to a compute‑efficient path for scalable self-improving agent development, where scaffolds evolved on smaller and cheaper backbones can later be reused on stronger models.

\section*{Limitations}

Self-improving coding-agent experiments remain expensive.
Even though MGM reuses existing trajectories for its comparative operators, evolution and evaluation on repository-level tasks still consume substantial wall-clock and GPU resources.
This constrains the number of independent evolution seeds and the breadth of hyperparameter sweeps we can report.

MGM is history-dependent.
Its advantage comes from comparative evidence in the archive.
Reaction-norm mutation needs multiple trajectories from the same agent, and cross-lineage hybridization needs overlapping tasks across lineages.
When the archive is small, task overlap is sparse, or informative failures have not yet appeared, MGM has little comparative evidence and may behave like single-trajectory baselines.
The failed-task pool partially mitigates this by increasing diagnostic overlap, but it cannot create informative contrasts before failures accumulate.

MGM improves the evidence given to self-modification, but it does not guarantee that the resulting edit is correct, general, or maintainable.
The Mendelian operators expose useful behavioral contrasts, yet the actual scaffold change is still produced by an LLM-based editor.
High-quality evidence therefore need not yield a high-quality modification, and failed edits can waste budget.
Relatedly, comparative evidence helps only when the backbone can diagnose failure mechanisms from trajectories.
A stronger coding specialist with weaker general reasoning may still produce weaker self-improvement under the same operators.

Our formal analysis and Monte Carlo study use an additive fitness surrogate.
They isolate diagnostic compression under controlled assumptions and do not capture the full complexity of editable agent scaffolds.
Empirically, primary evolution is reported on fixed 60-task subsets under a single matched budget, with broader checks on full Polyglot and held-out SWE-bench variants.
Subset selection and limited seed diversity leave residual uncertainty about variance across random restarts and alternative task samples.
Finally, our claims concern coding-agent scaffolds evaluated on public software-engineering benchmarks.
They do not establish that the same operators would transfer unchanged to non-coding agents or open-ended real-world software maintenance.

\section*{Ethical Considerations}

MGM advances self-improving coding agents that can modify their own source code.
Such systems have a different risk profile from static models because an erroneous or adversarial self-edit can persist in later descendants.
In our experiments, self-modification and evaluation run inside isolated containers with no network access and with read-only mounts of the host file system, following the safety protocol of prior archive-based self-improving agents.
Any deployment beyond this sandboxed coding setting requires a fresh review of the execution boundary.
Giving a self-modifying agent access to production systems, external APIs, or writable storage outside controlled containers could allow unintended changes to spread in ways that are hard to audit or reverse.
Practitioners should retain the isolated execution model and subject any broader action space to explicit safety review before deployment.

The same capability also has dual-use implications.
Scaffold-level self-improvement can amplify both beneficial coding assistance and harmful automation, including generation of malicious software or exploitation of vulnerable systems, if isolation is removed.
Our public release is intended for research on sandboxed self-evolution.
We discourage unconstrained deployment of evolved scaffolds outside controlled environments.

The benchmarks and backbone models used here may also inherit harmful or biased content.
SWE-bench, SWE-bench Pro, and Polyglot draw on public open-source repositories, and the backbone LLMs are trained on large web corpora, all of which may contain insecure code patterns and may overrepresent particular languages, ecosystems, or problem domains.
We did not introduce dedicated bias or safety audits of the evolved agents, so inherited fairness and security issues may persist.
In addition, public benchmark material may overlap with pretraining data, which limits how strongly offline scores should be read as evidence of robust real-world competence.
This work focuses on self-evolution of coding agents only, and our evolved agents are not equipped to act outside software-engineering related/ repositories.
Task-specific safety evaluation remains necessary before any use beyond general-purpose coding assistance.

\section*{Acknowledgments}

The authors gratefully acknowledge the scientific support and HPC resources provided by the hessian.AI Service Center (funded by the Federal Ministry of Research, Technology and Space, BMFTR, grant no.~16IS22091), the hessian.AI Innovation Lab (funded by the Hessian Ministry for Digital Strategy and Innovation, grant no.~S-DIW04/0013/003), and the Karlsruhe Institute of Technology National High Performance Computing Center (NHR@KIT) under the NHR projects 22560, and 24767.
The HoreKa supercomputer at NHR@KIT is funded by the Ministry of Science, Research and the Arts Baden-W\"urttemberg and by the Federal Ministry of Education and Research of Germany.
This work also receives support from the Munich Center for Machine Learning (MCML) and the Program of China Scholarship Council (Grant No.202508080292).
The funding bodies had no role in the design of methodology and experiments, the analysis and interpretation of results, or the writing of manuscript.

\bibliographystyle{iclr2026_conference}
\bibliography{custom}

\appendix
\newpage
\appendix
\startcontents[appendix]

\section*{Appendix Contents}
\printcontents[appendix]{}{1}{\setcounter{tocdepth}{2}}
\newpage

\section{Method Details}
\label{app:method-details}

\subsection{Pseudocode of Mendel G\"odel Machine}\label{app:pseudocode}

Here, we present pseudocode of MGM, which helps understand the full evolution pipeline.

\begin{figure}[h]
\vspace{1ex}
\centering
\begin{minipage}{0.9\textwidth}
\small

\begin{algorithm}[H]
\caption{Mendel G\"odel Machine}
\label{alg:mgm}

\KwInput{
Initial coding agent $a_0$, benchmark suite $\mathcal{B}$,
maximum iterations $T$, strategy weights
$\lambda_A,\lambda_B,\lambda_C$, failed task pool weights $\beta_\mathrm{fail}$
}
\KwOutput{Archive of agents $\mathcal{A}$}

$r(a_0 , \mathcal{B}) \leftarrow \mathrm{evaluate}(a_0,\mathcal{B})$
\tcp*[r]{Evaluate the base agent}

initialize $\mathcal{A} \leftarrow \{(a_0,r(a_0,\mathcal{B})\}$
\tcp*[r]{Start with the base agent}

initialize failed-task pool $\mathcal{P} \leftarrow \emptyset$
\tcp*[r]{Store informative failed tasks}

\For{$t \leftarrow 1$ \KwTo $T$}{

    decide action $u \in \{\mathrm{expand}, \mathrm{evaluate}\}$
    \tcp*[r]{Use Selection Policy}

    \eIf{$u = \mathrm{expand}$}{

        $a' \leftarrow \mathrm{SelectParent}(\mathcal{A})$
        \tcp*[r]{Select a promising lineage}

        $\Omega_{a'} \leftarrow \mathrm{EligibleStrategies}(a',\mathcal{A})$
        \tcp*[r]{Find available mutation operators}

        $\sigma \leftarrow \mathrm{SampleStrategy}(\Omega_{a'};
        \lambda_\mathrm{CM},\lambda_\mathrm{RM},\lambda_\mathrm{CH})$
        \tcp*[r]{Choose CM, RM, or CH}

        $E_\sigma \leftarrow \mathrm{BuildEvidence}(a',\sigma,\mathcal{A})$
        \tcp*[r]{Construct diagnostic evidence}

        $c \leftarrow \mathrm{Edit}_{\sigma}(a',E_\sigma)$
        \tcp*[r]{Self-modification}

        \If{$c.\mathrm{is\_valid}()$}{
            $\mathcal{A} \leftarrow \mathcal{A} \cup \{(c,\emptyset)\}$
            \tcp*[r]{Keep valid child agent}
        }

    }{

        $a \leftarrow \mathrm{SelectAgent}(\mathcal{A})$
        \tcp*[r]{Choose agent to test}

        $\tau \leftarrow \mathrm{SelectTask}(a,\mathcal{B},\mathcal{P},\beta_\mathrm{fail})$
        \tcp*[r]{Consider informative failed tasks}

        $r(a,\tau) \leftarrow \mathrm{evaluate}(a,\tau)$
        \tcp*[r]{Evaluate on one task}

        update $\mathcal{A}$ with $(a,\tau,r(a,\tau))$
        \tcp*[r]{Store trajectory and result}

        \If{$r(a,\tau) = 0$}{
            $\mathcal{P} \leftarrow \mathcal{P} \cup \{\tau\}$
            \tcp*[r]{Update failed-task pool}
        }
    }
}

\Return $\mathcal{A}$

\end{algorithm}

\end{minipage}
\vspace{1ex}
\end{figure}

Algorithm~\ref{alg:mgm} gives the complete MGM loop. The procedure follows the standard archive-based self-improvement structure: each budgeted step either evaluates an existing agent on a task or expands the archive by self-modifying an agent. Evaluations update both the agent archive and the failed-task pool. During expansion, MGM constructs the set of eligible operators from already collected trajectories and samples one of clonal mutation, reaction-norm mutation, or cross-lineage hybridization. The selected operator determines how evidence is assembled for the editor: from one failed trajectory, multiple trajectories of the same agent, or shared-task trajectories across lineages. Therefore, MGM improves the informativeness of each self-modification step without adding extra task evaluations.

\subsection{Primary-Lineage Selection in Cross-Lineage Hybridization}
\label{app:ch_primary_selection}

In the main algorithm, the policy $\pi$ first selects an archive node as an anchor for expansion. For Clonal Mutation and Reaction-norm Mutation, this anchor is also the primary agent whose codebase is edited. Cross-lineage Hybridization differs because the purpose of the operator is to transfer a useful behavior from one lineage to another. Therefore, the anchor node is used to retrieve a comparable peer, but the final primary lineage is determined by the shared-task outcomes.

Concretely, suppose the anchor agent $a_i$ and a peer agent $a_j$ have both attempted a shared task $\tau^\star$, and the task is not solved by both agents. MGM constructs the comparison evidence
\begin{equation}
\begin{aligned}
    E_C(i,j,\tau^\star)
=
\{(\varphi(a_i,\tau^\star), r(a_i,\tau^\star)),\\
  (\varphi(a_j,\tau^\star), r(a_j,\tau^\star))\}.
\end{aligned}
\end{equation}
If exactly one of the two agents solves $\tau^\star$, the failing agent is selected as the primary agent $a_p$, and the successful agent is used as the context donor $a_q$. The resulting child is attached to the failing lineage:
\begin{equation}
    a' \leftarrow \varPhi_C(a_p, E_C), \ a' \in \mathrm{children}(a_p).
\end{equation}
This direction is intentional: the goal is not to further edit the already successful lineage, but to let the failing lineage inherit a transferable behavior observed in the successful trajectory.

If both agents fail on $\tau^\star$, neither trajectory provides a direct success demonstration. In this case, MGM uses the higher-utility lineage as the primary agent and the other lineage as contrastive context. This fallback preserves the archive policy's preference for promising lineages while still allowing the editor to diagnose complementary failure modes from the same-task comparison.

Thus, the node selected by $\pi$ should be interpreted as an anchor for constructing a controlled comparison, not necessarily as the final parent edited by CH. The final child is always attached to the primary lineage determined by the CH comparison. This implementation matches the diagnostic role of hybridization: successful trajectories act as donors when available, and failed--failed comparisons are used to identify general weaknesses rather than task-specific patches.

\section{Theory and Simulation Details}
\label{app:theory-simulation}



\subsection{Theoretical Justification of Comparative Fix Probability}
\label{app:comparative-fix-probability}

This section provides the full derivation for Proposition~1 in Section~\ref{prop1}. The goal is to justify why the comparative operators used by MGM can induce a higher effective fix probability than single-trajectory clonal mutation.

\paragraph{Diagnostic model.}
Let an agent \(a\) be represented by a binary genotype \(g(a)\in\{0,1\}^{L}\), and let the oracle genotype be \(g^\star=\mathbf{1}\). Define the set of incorrect loci as
\begin{equation}
M(a)=\{\ell\in[L]:g_\ell(a)\neq g^\star_\ell\}.
\end{equation}
Each task \(\tau\) requires a subset of loci \(R_\tau\subseteq[L]\), with \(|R_\tau|=k\). The task is solved if and only if all required loci are correct:
\begin{equation}
r(a,\tau)=1
\quad\Longleftrightarrow\quad
R_\tau\cap M(a)=\varnothing.
\end{equation}
Therefore, a failed task only reveals that at least one of its required loci is incorrect:
\begin{equation}
r(a,\tau)=0
\quad\Longrightarrow\quad
R_\tau\cap M(a)\neq\varnothing.
\end{equation}

We view a self-modification operator \(\varPhi_\sigma\) as a diagnostic procedure. Given evidence \(E\), it constructs an implicit candidate set \(C_\sigma(E)\subseteq[L]\) of loci that may explain the observed failure. The editor then attempts to modify one locus in \(C_\sigma(E)\). Suppose that if the selected locus is truly incorrect, the editor repairs it with probability \(s\in(0,1]\). Then the effective fix probability of operator \(\sigma\) is
\begin{equation}
p_f^\sigma
=
s\cdot
\Pr_{\ell\sim C_\sigma(E)}[\ell\in M(a)].
\end{equation}
Thus, an operator has higher fix probability when its evidence produces a candidate set with higher posterior density of truly incorrect loci.

\paragraph{Clonal mutation.}
Clonal mutation uses a single failed trajectory \((\varphi(a,\tau_t),0)\). Since this evidence only implies that at least one locus in \(R_{\tau_t}\) is incorrect, the natural candidate set is
\begin{equation}
C_{\mathrm{CM}}=R_{\tau_t}.
\end{equation}
Let the number of truly incorrect loci among the \(k\) loci required by the failed task be denoted as
\begin{equation}
c=|R_{\tau_t}\cap M(a)|.
\end{equation}
Then we have
\begin{equation}
p_f^{\mathrm{CM}}
=
s\cdot\frac{c}{k}.
\end{equation}
In the sparse-defect case where a failure is caused by one dominant missing capability, \(c=1\), and therefore
\begin{equation}
p_f^{\mathrm{CM}}=\frac{s}{k}.
\end{equation}

\paragraph{Reaction-norm mutation.}
Reaction-norm mutation uses multiple trajectories from the same genotype. Consider two failed tasks \(\tau_t\) and \(\tau_r\) of the same agent. We assume that these two failures share a recurring causal defect \(b\in M(a)\), so that
\begin{equation}
b\in R_{\tau_t}\cap R_{\tau_r}.
\end{equation}
This models the case where the same agent repeatedly fails because of the same scaffold-level weakness rather than unrelated task-specific accidents.

Under this sound-comparison assumption, the common explanatory region is
\begin{equation}
C_{\mathrm{RM}}=R_{\tau_t}\cap R_{\tau_r}.
\end{equation}
Assume the remaining \(k-1\) required loci of each task are sampled independently from \([L]\setminus\{b\}\). Then the expected size of the intersection is
\begin{equation}
\mathbb{E}[|C_{\mathrm{RM}}|]
=
1+\frac{(k-1)^2}{L-1}.
\end{equation}
Since \(L>k\), we have
\begin{equation}
1+\frac{(k-1)^2}{L-1}<k.
\end{equation}
Because the recurring causal defect \(b\) is contained in \(C_{\mathrm{RM}}\), the probability of targeting a truly incorrect locus is larger than in the single-trajectory candidate set. In the sparse-defect case, this gives
\begin{equation}
p_f^{\mathrm{RM}}
\geq
s\cdot
\frac{1}{1+\frac{(k-1)^2}{L-1}}
>
\frac{s}{k}
=
p_f^{\mathrm{CM}}.
\end{equation}
Therefore, reaction-norm mutation improves fix probability by compressing the candidate set from the full task-relevant region \(R_{\tau_t}\) to the intersection of multiple failures that share the same genotype-level defect.

\paragraph{Cross-lineage hybridization.}
Cross-lineage hybridization compares different genotypes on the same task. Consider a target agent \(a_t\) that fails on task \(\tau\), and a reference agent \(a_r\) from another lineage that succeeds:
\begin{equation}
r(a_t,\tau)=0,
\qquad
r(a_r,\tau)=1.
\end{equation}
The target failure implies
\begin{equation}
R_\tau\cap M(a_t)\neq\varnothing,
\end{equation}
whereas the reference success implies
\begin{equation}
R_\tau\cap M(a_r)=\varnothing.
\end{equation}
Thus, the reference agent provides a contrastive control: the same task-relevant loci are sufficient for success in the reference lineage, but at least one of them is defective in the target lineage.

Let the number of causal target defects within the task-relevant region be denoted as 
\begin{equation}
c=|R_\tau\cap M(a_t)|.
\end{equation}
Clonal mutation must search over the whole \(R_\tau\), so
\begin{equation}
p_f^{\mathrm{CM}}
=
s\cdot\frac{c}{k}.
\end{equation}
By comparing the failed target trajectory with the successful reference trajectory, cross-lineage hybridization can filter out loci that are task-relevant but unlikely to explain the target-specific failure. Let the resulting contrastive candidate set be
\begin{equation}
C_{\mathrm{CH}}=(R_\tau\cap M(a_t))\cup N,
\end{equation}
where \(N\) denotes non-causal differences that remain after comparison. Let \(h=|N|\). Then
\begin{equation}
p_f^{\mathrm{CH}}
=
s\cdot\frac{c}{c+h}.
\end{equation}
Whenever the reference comparison removes at least one irrelevant candidate, we have
\begin{equation}
c+h<k.
\end{equation}
Therefore,
\begin{equation}
p_f^{\mathrm{CH}}
=
s\cdot\frac{c}{c+h}
>
s\cdot\frac{c}{k}
=
p_f^{\mathrm{CM}}.
\end{equation}
Thus, cross-lineage hybridization improves fix probability by using a successful or contrastive reference lineage to remove non-causal explanations from the candidate set.

\paragraph{Discussion of assumptions.}
The above derivation is not intended to claim that comparative operators are unconditionally superior. It relies on four assumptions. First, failures are caused by relatively sparse causal defects, so that narrowing the candidate set meaningfully increases the density of true defects. Second, reaction-norm mutation is most beneficial when the compared trajectories share a recurring genotype-level weakness. Third, cross-lineage hybridization requires an informative reference trajectory that serves as a useful contrastive control. Fourth, the editor must be capable of exploiting the comparative evidence. When these assumptions fail, the comparative operators may provide little or no fix-quality advantage; this case is explicitly represented in our simulation by the null setting \(\rho=1\).

\subsection{Simulation Parameters}

\vspace{1ex}

\begin{table*}[h]
\centering
\small
\begin{tabular}{rlcc}
\toprule
Type & Parameter & Symbol & Value \\
\midrule
\multirow{4}{*}{Genotype \& Phenotype}
  & Chain length          & $L$             & $100$  \\
  & Initial edit distance & $d_0$           & $\{10, 20, 40, 80\}$    \\
  & Task pool size        & $N$             & $200$  \\
  & Loci examined per task         & $k$             & $5$    \\
\midrule
\multirow{4}{*}{Costs}
  & Task evaluation       & $c_\tau$        & $1.0$  \\
  & $\varPhi_{\mathrm{CM}}$ edit  & $c_{\mathrm{CM}}$  & $1.0$  \\
  & $\varPhi_{\mathrm{RM}}$ edit  & $c_{\mathrm{RM}}$  & $1.0$  \\
  & $\varPhi_{\mathrm{CH}}$ edit  & $c_{\mathrm{CH}}$  & $1.0$  \\
\midrule
\multirow{5}{*}{Edit quality}
  & Fix prob.\ ($\varPhi_{\mathrm{CM}}$)  & $p_f^{\mathrm{CM}}$  & $0.25$ \\
  & Fix prob.\ ($\varPhi_{\mathrm{RM}}$)  & $p_f^{\mathrm{RM}}$  & $\rho\cdot\varPhi_{\mathrm{CM}}$ \\
  & Fix prob.\ ($\varPhi_{\mathrm{CH}}$)  & $p_f^{\mathrm{CH}}$  & $\rho\cdot\varPhi_{\mathrm{CM}}$ \\
  & Fix-ratio             &  $\rho$   & $\{1.0, 1.2, 1.5, 2.0\}$ \\
  & Break prob.\ (all $\varPhi$ operators) & $p_b$               & $0.05$ \\
\midrule
\multirow{3}{*}{DGM}
  & Population size       & $n_{\mathrm{pop}}$       & $5$    \\
  & Evals per node        & $n_{\mathrm{eval}}$      & $10$   \\
  & Selection fraction    & $s_{\mathrm{frac}}$      & $0.4$  \\
\midrule
\multirow{3}{*}{HGM}
  & Min evals before edit & $n_{\min}$               & $3$    \\
  & Max evals before edit & $n_{\max}$               & $20$   \\
  & Failure threshold     & $\tau_{\mathrm{edit}}$   & $0.5$  \\
\midrule
\multirow{3}{*}{MGM}
  & $\Pr(\varPhi_{\mathrm{RM}} \mid \text{available})$ & $\pi_{\mathrm{RM}}$ & $0.45$ \\
  & $\Pr(\varPhi_{\mathrm{CH}} \mid \text{available})$ & $\pi_{\mathrm{CH}}$ & $0.45$ \\
  & Min tasks for $\varPhi_{\mathrm{RM}}$              & $m_{\mathrm{RM}}$   & $2$   \\
\midrule
\multirow{3}{*}{Monte Carlo}
  & Total budget          & $B$                      & $500$  \\
  & Seeds                 & $n_{\mathrm{seeds}}$     & $100$  \\
  & Budget checkpoints    & $n_{\mathrm{chk}}$       & $500$  \\
\bottomrule
\end{tabular}
\caption{\textbf{Parameter settings for simulation.} All edit costs are equal so that the HGM--MGM comparison isolates diagnostic quality alone. The fix-probability advantage ratio for the main simulation is $\rho = p_f^{\mathrm{RM}} / p_f^{\mathrm{CM}} = 2.0$; the sweep grid varies $\rho$ and $d_0$ to test robustness.}
\label{tab:simulation_params}
\end{table*}

\section{Experiment Details}
\label{app:experiment-details}

\subsection{Compute and Software Environment}
\label{app:compute_env}

All experiments were run on cluster nodes with eight NVIDIA H100 GPUs (80\,GB GPU memory each) and approximately 2\,TB of host memory.
The Qwen models used in our experiments\footnote{\url{https://huggingface.co/Qwen/Qwen3.6-35B-A3B} and \url{https://huggingface.co/Qwen/Qwen3-Coder-Next}} was served with vLLM through an OpenAI-compatible API using tensor parallelism across the visible GPUs.
We set the maximum context length to 262{,}144 tokens and the GPU memory utilization to 92\%.
Benchmark tasks were executed inside Apptainer containers built from a Python~3.10 base image, with the repository bind-mounted into the container.
By default, host networking was disabled, and the container addressed the vLLM endpoint through the host node IP.
For DeepSeek-V4 models, we directly accessed their preview release via the official DeepSeek API\footnote{\url{https://api-docs.deepseek.com/news/news260424/}}. For generation, the Qwen models were run with thinking enabled, a temperature of 1.0, top-
$p$ of 0.95, and top-$k$ of 20, following the sampling configuration loaded by vLLM from the models’ generation configuration. No explicit reasoning-effort level or thinking-token budget was imposed. The DeepSeek-V4 models were evaluated in non-thinking mode, with the API-default temperature and top-$p$ values of 1.0. Parallel tool calls were disabled for both model families.

\subsection{Hyper-parameter Settings}

In this section, to improve transparency and reproducibility, 
we present the hyper-parameter settings used in our code implementations, 
as shown in Table~\ref{tab:hyperparameter}.

\begin{table}[t]
    \centering
    \small
    \begin{tabular}{cc}
        \toprule
        Hyper-parameter & Value \\
        \midrule
        $\beta_{\mathrm{fail}}$ & $1.0$  \\
        $\lambda_{\mathrm{CM}}$ & $0.10$ \\
        $\lambda_{\mathrm{RM}}$ & $0.45$ \\
        $\lambda_{\mathrm{CH}}$ & $0.45$ \\
        $m_{\mathrm{RM}}$       & $2$    \\
        \bottomrule
    \end{tabular}
    \caption{\textbf{Hyper-parameter settings used in all experiments.}}
    \label{tab:hyperparameter}
\end{table}

\section{Benchmark Details}\label{app:benchmark_details}

To control evaluation cost, instead of evaluating MGM on the full SWE-bench Pro and SWE-bench Multilingual, we follow the procedures by~\citet{zhang2026darwin} and utilize ChatGPT to randomly and separately choose 60 representative tasks for each benchmark.

For SWE-bench Pro, the selective 60-task subset is required to be representative and involve all the six Programming Languages, \texttt{JavaScript, Python, Java, C++, TypeScript, Go}. Here we report the 60 tasks we used for our SWE-bench Pro evaluation:

\small
{\footnotesize
\begin{itemize}[leftmargin=*, itemsep=0.25em]
\sloppy

    \item \path{ansible__ansible-0ea40e09d1b35bcb69ff4d9cecf3d0defa4b36e8}
    \item \path{ansible__ansible-189fcb37f973f0b1d52b555728208eeb9a6fce83}
    \item \path{ansible__ansible-3889ddeb4b780ab4bac9ca2e75f8c1991bcabe83}
    \item \path{ansible__ansible-5260527c4a71bfed99d803e687dd19619423b134}
    \item \path{ansible__ansible-a20a52701402a12f91396549df04ac55809f68e9}

    \item \path{internetarchive__openlibrary-308a35d6999427c02b1dbf5211c033ad3b352556}
    \item \path{internetarchive__openlibrary-30bc73a1395fba2300087c7f307e54bb5372b60a}
    \item \path{internetarchive__openlibrary-4b7ea2977be2747496ba792a678940baa985f7ea}
    \item \path{internetarchive__openlibrary-7edd1ef09d91fe0b435707633c5cc9af41dedddf}
    \item \path{internetarchive__openlibrary-9bdfd29fac883e77dcbc4208cab28c06fd963ab2}

    \item \path{qutebrowser__qutebrowser-66cfa15c372fa9e613ea5a82d3b03e4609399fb6}
    \item \path{qutebrowser__qutebrowser-6b320dc18662580e1313d2548fdd6231d2a97e6d}
    \item \path{qutebrowser__qutebrowser-8cd06741bb56cdca49f5cdc0542da97681154315}
    \item \path{qutebrowser__qutebrowser-8f46ba3f6dc7b18375f7aa63c48a1fe461190430}
    \item \path{qutebrowser__qutebrowser-99029144b5109bb1b2a53964a7c129e009980cd9}

    \item \path{flipt-io__flipt-02e21636c58e86c51119b63e0fb5ca7b813b07b1}
    \item \path{flipt-io__flipt-5aef5a14890aa145c22d864a834694bae3a6f112}
    \item \path{flipt-io__flipt-9d25c18b79bc7829a6fb08ec9e8793d5d17e2868}
    \item \path{flipt-io__flipt-b68b8960b8a08540d5198d78c665a7eb0bea4008}
    \item \path{flipt-io__flipt-e2bd19dafa7166c96b082fb2a59eb54b4be0d778}

    \item \path{future-architect__vuls-78b52d6a7f480bd610b692de9bf0c86f57332f23}
    \item \path{future-architect__vuls-86b60e1478e44d28b1aff6b9ac7e95ceb05bc5fc}
    \item \path{future-architect__vuls-e049df50fa1eecdccc5348e27845b5c783ed7c76}

    \item \path{gravitational__teleport-3fa6904377c006497169945428e8197158667910}
    \item \path{gravitational__teleport-3ff75e29fb2153a2637fe7f83e49dc04b1c99c9f}
    \item \path{gravitational__teleport-73cc189b0e9636d418c4470ecce0d9af5dae2f02}
    \item \path{gravitational__teleport-ba6c4a135412c4296dd5551bd94042f0dc024504}

    \item \path{navidrome__navidrome-66b74c81f115c78cb69910b0472eeb376750efc4}
    \item \path{navidrome__navidrome-812dc2090f20ac4f8ac271b6ed95be5889d1a3ca}
    \item \path{navidrome__navidrome-c90468b895f6171e33e937ff20dc915c995274f0}

    \item \path{element-hq__element-web-1077729a19c0ce902e713cf6fab42c91fb7907f1}
    \item \path{element-hq__element-web-41dfec20bfe9b62cddbbbf621bef2e9aa9685157}
    \item \path{element-hq__element-web-53a9b6447bd7e6110ee4a63e2ec0322c250f08d1}
    \item \path{element-hq__element-web-9a31cd0fa849da810b4fac6c6c015145e850b282}
    \item \path{element-hq__element-web-b007ea81b2ccd001b00f332bee65070aa7fc00f9}

    \item \path{NodeBB__NodeBB-397835a05a8e2897324e566b41c5e616e172b4af}
    \item \path{NodeBB__NodeBB-51d8f3b195bddb13a13ddc0de110722774d9bb1b}
    \item \path{NodeBB__NodeBB-97c8569a798075c50e93e585ac741ab55cb7c28b}
    \item \path{NodeBB__NodeBB-be43cd25974681c9743d424238b7536c357dc8d3}
    \item \path{NodeBB__NodeBB-f48ed3658aab7be0f1165d4c1f89af48d7865189}

    \item \path{protonmail__webclients-01b519cd49e6a24d9a05d2eb97f54e420740072e}
    \item \path{protonmail__webclients-08bb09914d0d37b0cd6376d4cab5b77728a43e7b}
    \item \path{protonmail__webclients-51742625834d3bd0d10fe0c7e76b8739a59c6b9f}
    \item \path{protonmail__webclients-6f8916fbadf1d1f4a26640f53b5cf7f55e8bedb7}
    \item \path{protonmail__webclients-8142704f447df6e108d53cab25451c8a94976b92}

    \item \path{tutao__tutanota-09c2776c0fce3db5c6e18da92b5a45dce9f013aa}
    \item \path{tutao__tutanota-12a6cbaa4f8b43c2f85caca0787ab55501539955}
    \item \path{tutao__tutanota-1e516e989b3c0221f4af6b297d9c0e4c43e4adc3}
    \item \path{tutao__tutanota-1ff82aa365763cee2d609c9d19360ad87fdf2ec7}
    \item \path{tutao__tutanota-219bc8f05d7b980e038bc1524cb021bf56397a1b}
    \item \path{tutao__tutanota-40e94dee2bcec2b63f362da283123e9df1874cc1}
    \item \path{tutao__tutanota-4b4e45949096bb288f2b522f657610e480efa3e8}
    \item \path{tutao__tutanota-51818218c6ae33de00cbea3a4d30daac8c34142e}
    \item \path{tutao__tutanota-8513a9e8114a8b42e64f4348335e0f23efa054c4}
    \item \path{tutao__tutanota-b4934a0f3c34d9d7649e944b183137e8fad3e859}
    \item \path{tutao__tutanota-d1aa0ecec288bfc800cfb9133b087c4f81ad8b38}
    \item \path{tutao__tutanota-db90ac26ab78addf72a8efaff3c7acc0fbd6d000}
    \item \path{tutao__tutanota-de49d486feef842101506adf040a0f00ded59519}
    \item \path{tutao__tutanota-fb32e5f9d9fc152a00144d56dd0af01760a2d4dc}
    \item \path{tutao__tutanota-fe240cbf7f0fdd6744ef7bef8cb61676bcdbb621}

\fussy
\end{itemize}
}

\normalsize

SWE-bench Multilingual contains 300 tasks from 42 repositories and spans nine programming languages: \texttt{C}, \texttt{C++}, \texttt{Go}, \texttt{Java}, \texttt{JavaScript}, \texttt{TypeScript}, \texttt{PHP}, \texttt{Ruby}, and \texttt{Rust}. We construct the subset to cover these language groups while keeping the evaluation cost manageable.
Here we report the exact tasks used in our SWE-bench Multilingual evaluation below.

\small
{\footnotesize
\begin{itemize}[leftmargin=*, itemsep=0.25em]
\sloppy

    \item \path{apache__druid-14092}
    \item \path{apache__lucene-13494}
    \item \path{apache__lucene-13704}
    \item \path{astral-sh__ruff-15356}
    \item \path{astral-sh__ruff-15443}

    \item \path{axios__axios-4731}
    \item \path{babel__babel-16130}
    \item \path{briannesbitt__carbon-3005}
    \item \path{briannesbitt__carbon-3041}
    \item \path{briannesbitt__carbon-3103}

    \item \path{caddyserver__caddy-4943}
    \item \path{caddyserver__caddy-5404}
    \item \path{caddyserver__caddy-5870}
    \item \path{caddyserver__caddy-5995}
    \item \path{facebook__docusaurus-9183}

    \item \path{fastlane__fastlane-19207}
    \item \path{fastlane__fastlane-20642}
    \item \path{fastlane__fastlane-20975}
    \item \path{fluent__fluentd-3917}
    \item \path{fmtlib__fmt-2317}

    \item \path{fmtlib__fmt-2457}
    \item \path{gohugoio__hugo-12579}
    \item \path{google__gson-1014}
    \item \path{immutable-js__immutable-js-2006}
    \item \path{jekyll__jekyll-8771}

    \item \path{jqlang__jq-2235}
    \item \path{jqlang__jq-2658}
    \item \path{jqlang__jq-2839}
    \item \path{jqlang__jq-2919}
    \item \path{laravel__framework-51195}

    \item \path{laravel__framework-53914}
    \item \path{laravel__framework-53949}
    \item \path{nushell__nushell-13605}
    \item \path{php-cs-fixer__php-cs-fixer-7635}
    \item \path{phpoffice__phpspreadsheet-3570}

    \item \path{phpoffice__phpspreadsheet-4114}
    \item \path{preactjs__preact-2757}
    \item \path{preactjs__preact-3454}
    \item \path{preactjs__preact-3562}
    \item \path{preactjs__preact-4436}

    \item \path{projectlombok__lombok-3009}
    \item \path{projectlombok__lombok-3350}
    \item \path{projectlombok__lombok-3422}
    \item \path{projectlombok__lombok-3479}
    \item \path{projectlombok__lombok-3594}

    \item \path{prometheus__prometheus-10633}
    \item \path{prometheus__prometheus-12874}
    \item \path{prometheus__prometheus-14861}
    \item \path{redis__redis-11734}
    \item \path{redis__redis-13115}

    \item \path{rubocop__rubocop-13375}
    \item \path{rubocop__rubocop-13479}
    \item \path{rubocop__rubocop-13627}
    \item \path{sharkdp__bat-2650}
    \item \path{tokio-rs__axum-1730}

    \item \path{tokio-rs__tokio-6752}
    \item \path{tokio-rs__tokio-6838}
    \item \path{uutils__coreutils-6575}
    \item \path{uutils__coreutils-6682}
    \item \path{vuejs__core-11739}

\fussy
\end{itemize}
}

\section{Additional Results}
\label{app:additional-results}

\subsection{Full Evaluation on Polyglot}\label{app:full_polyglot}

To verify that the Polyglot improvement reported in the main experiments is not an artifact of the 60-task subset, we additionally evaluate the best MGM-discovered agent on the full Polyglot benchmark. This evaluation is performed only after evolution is complete: the agent is not further updated, and the full benchmark is used only for post-evolution evaluation and is not used to further update the agent. 

\begin{figure}[h]
    \centering
    \includegraphics[width=0.5\linewidth]{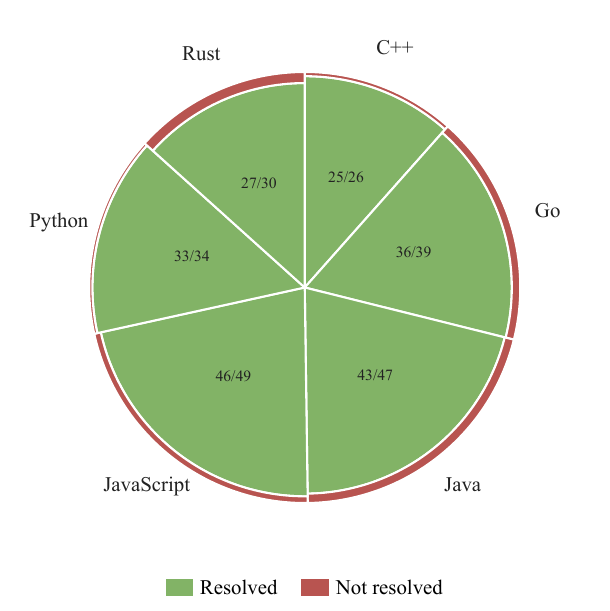}
    \caption{\textbf{Polyglot-225 performance of the MGM-discovered agent by language}. The agent is evaluated after evolution without further self-modification and solves 210 out of 225 tasks overall. The area of pie slices report the proportion of resolved and unresolved tasks within each language, with labels showing the ratio of resolved to total tasks. MGM maintains high accuracy across C++, Go, Java, JavaScript, Python, and Rust, indicating that the evolved improvements transfer across languages rather than specializing to a single language subset.}
    \label{fig:full_polyglot}
\end{figure}

As shown in Figure~\ref{fig:full_polyglot}, the MGM-evolved agent solves 210 out of 225 tasks, achieving an overall accuracy of \textbf{\textit{93.3\%}}. This closely matches the 93.2\% accuracy observed on Polyglot-60 in the main experiment, suggesting that the improvement is stable when moving from the subset evaluation to the full benchmark. The gains are also broadly distributed across languages: the agent solves 25/26 C++ tasks, 36/39 Go tasks, 43/47 Java tasks, 46/49 JavaScript tasks, 33/34 Python tasks, and 27/30 Rust tasks.

These results support the claim that MGM learns reusable, language-agnostic workflow improvements rather than narrow heuristics tied to a particular language or subset. In particular, the consistently high resolution rate across C++, Go, Java, JavaScript, Python, and Rust is consistent with the role of reaction-norm mutation and cross-lineage hybridization: the former identifies recurring agent-level weaknesses across tasks, while the latter transfers useful behavioral traits across lineages. The remaining failures are not concentrated in a single language, indicating that future improvements should likely target harder residual failure modes rather than language-specific specialization.

\subsection{Visualizations of Evolution Trees}
\label{app:evolution_trees}

\begin{figure}[t]
    \centering

    \begin{subfigure}[t]{0.48\linewidth}
        \centering
        \includegraphics[width=\linewidth]{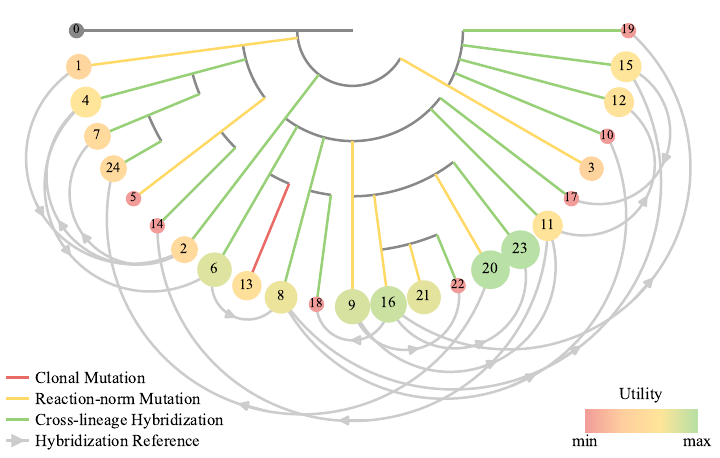}
        \caption{MGM }
        \label{fig:mgm_tree}
    \end{subfigure}
    \hfill
    \begin{subfigure}[t]{0.48\linewidth}
        \centering
        \includegraphics[width=\linewidth]{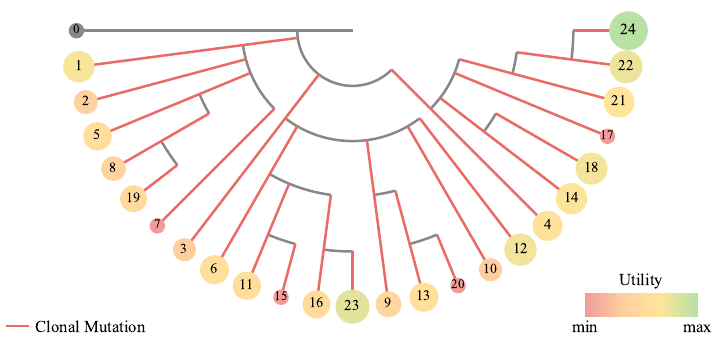}
        \caption{HGM}
        \label{fig:hgm_tree}
    \end{subfigure}
    \vspace{1ex}
    \begin{subfigure}[t]{0.48\linewidth}
        \centering
        \includegraphics[width=\linewidth]{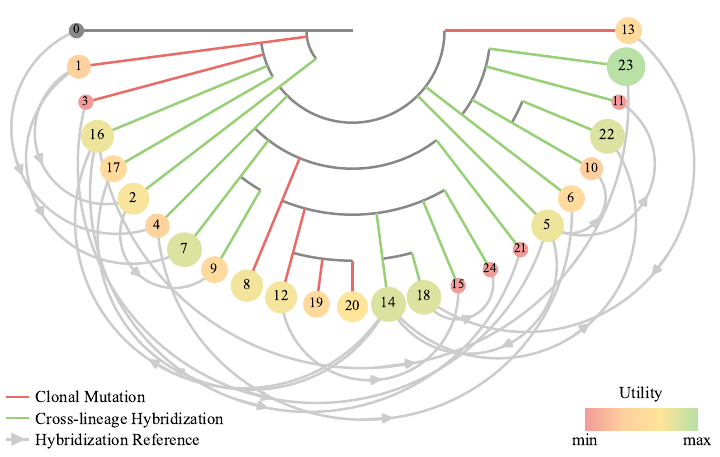}
        \caption{w/o $\varPhi_\textrm{RM}$ ($\varPhi_\textrm{CM}$ and $\varPhi_\textrm{CH}$ only)}
        \label{fig:remove_CH_tree}
    \end{subfigure}
    \hfill
    \begin{subfigure}[t]{0.48\linewidth}
        \centering
        \includegraphics[width=\linewidth]{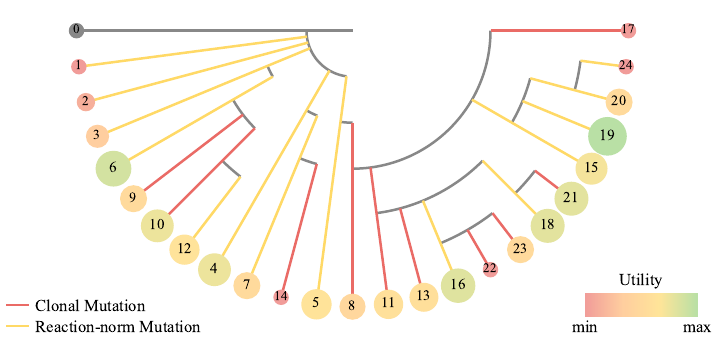}
        \caption{w/o $\varPhi_\textrm{CH}$ ($\varPhi_\textrm{CM}$ and $\varPhi_\textrm{RM}$ only)}
        \label{fig:remove_RM_tree}
    \end{subfigure}

    \caption{\textbf{Evolution trees of MGM, HGM, and MGM's ablated variants.} Results after 200 $\varphi$-evaluations and 24 $\varPhi$-expansions on Polyglot. Nodes are independently colored  by utility estimates aggregated over each tree's $\varphi$-evaluation results.
    }
\end{figure}

All four figures use the same visual encoding: node fill color denotes evaluation accuracy, while edge color denotes the child node’s self-improvement operator ($\varPhi_{\mathrm{CM}}$/$\varPhi_{\mathrm{RM}}$/$\varPhi_{\mathrm{CH}}$). Starred nodes mark the representative agent chosen for each experimental setting.

Figure~\ref{fig:mgm_tree} shows the evolution tree produced by the full MGM system under a budget of 200 task evaluations, yielding 24 nodes. All three operators are active: Clonal Mutation, Reaction-norm Mutation, and Cross-lineage Hybridization. Node fill color encodes evaluation accuracy. 
The node \#20 and \#23 has the highest utility 1.00 but only with 15 and 4 evaluations, respectively.
The node \#16 has utility 0.91 after 35 evaluations and is selected as the final result.

Figure~\ref{fig:hgm_tree} shows the evolution tree of the HGM baseline, which uses Clonal Mutation as its only self-modification operator. Under the same 200-evaluation budget, HGM also produces 24 nodes, but all child edges correspond to single-trajectory clonal edits. 
Compared with full MGM, HGM still explores multiple branches, but it lacks the additional comparative evidence channels represented by edges with different colors. 
Although node \#24 has the highest utility 1.00, it has only been evaluated 5 times, so its color reflects a high-variance estimate rather than a reliable final selection. 
The node \#18 is selected as result, with utility 0.71 after 14 evaluations.

Figure~\ref{fig:remove_CH_tree} shows the ablation that removes Reaction-norm Mutation. Clonal Mutation and Cross-lineage Hybridization remain enabled. The node \#23 with highest utility 1.00 got only 2 evaluations, and the node \#14 with utility 0.84 after 50 evaluations is selected as result.

Figure~\ref{fig:remove_RM_tree} shows the ablation that removes Cross-lineage Hybridization. Clonal Mutation and Reaction-norm Mutation remain active. All improvement is confined to within-lineage evolution. The node \#19 has utility 1.00 after only 2 evaluations, and the selected result is node \#6 with utility 0.88, after 16 evaluations.

\begin{figure}[t]
    \centering
    \vspace{-2ex}
        \includegraphics[width=0.65\linewidth]{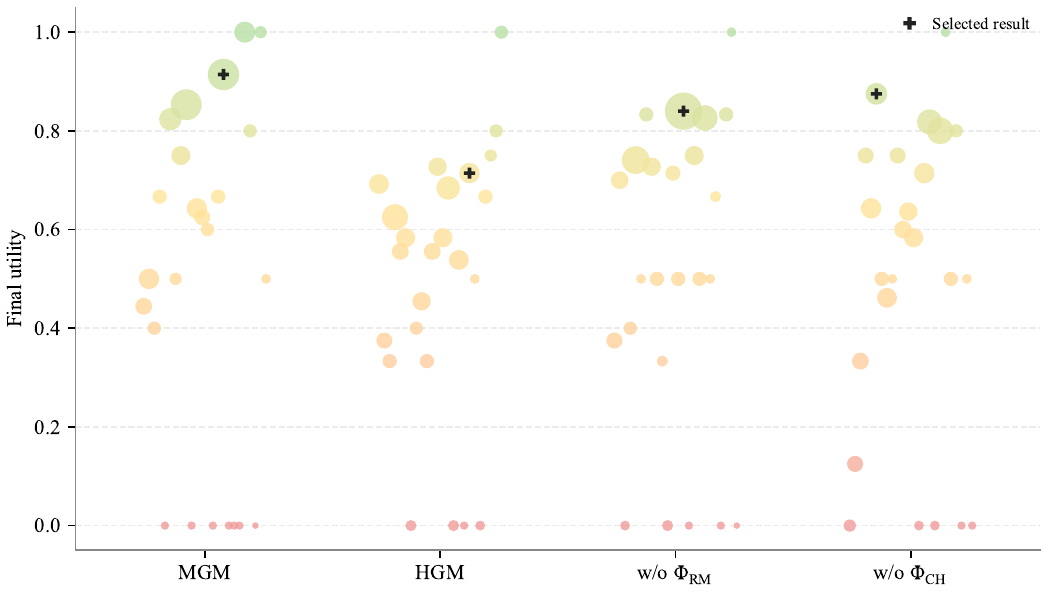}
        \caption{\textbf{Final per-node utilities for MGM, HGM, and the ablations.}
        Each point is one of the 24 evolved nodes.
        Vertical position and fill color encode utility.
        Marker area is proportional to the number of evaluations.
        The marked node in each column is the selected final result.}
        \label{fig:utility_scatter_comparison}
\end{figure}

In Figure~\ref{fig:utility_scatter_comparison} we additionally report the final utilities of all 24 nodes under each configuration.
Point size reflects evaluation count, making high-utility nodes with few measurements visually distinct from better-supported estimates.
The raw maximum is often attained by a sparsely evaluated node, whereas the selected result favors a high utility backed by more evaluations.

\section{Additional Discussion}
\label{app:discussion}

\subsection{When Does Hybridization Help? A Case Study}\label{hybridhelp}

\begin{figure*}[h]
    \vspace{2.5ex}
    \centering
    \includegraphics[width=0.5\linewidth]{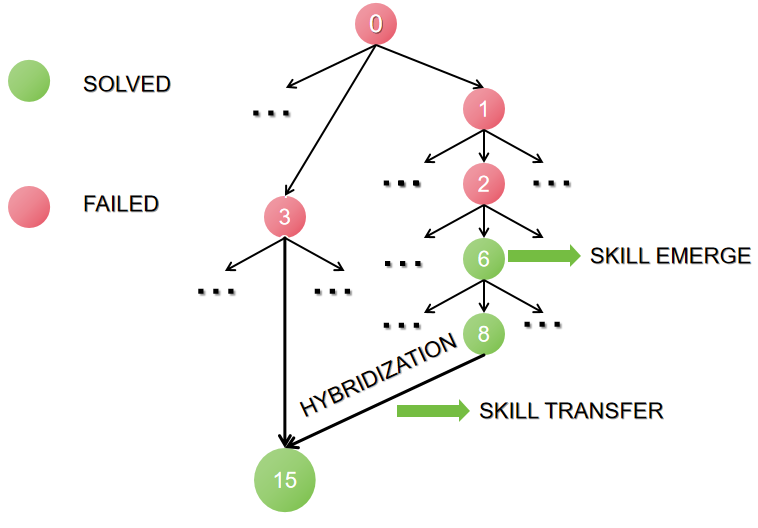}
    \vspace{-1.5ex}
    \caption{\textbf{Illustration of Cross-lineage Hybridization}. Nodes are colored by their outcome on the shared diagnostic task \texttt{javascript\_\_queen-attack}: red nodes fail and green nodes solve it.}
    \label{fig:HybridizationHelp}
    \vspace{1.5ex}
\end{figure*}

Just as Figure~\ref{fig:HybridizationHelp} illustrates, \textbf{Cross-lineage Hybridization} enables the evolutionary process to transfer skills discovered in one archive to another, thereby improving evolutionary efficiency. In this example, both node \#1 and node \#2 initially fail on the same diagnostic task, \texttt{javascript\_\_queen-attack}. Along the donor lineage, node \#2 later produces node \#6, which is the first descendant in this branch to solve the task. The evolved behavior can be summarized as a test-contract skill. Before implementing, the agent learns to read tests as strict behavioral contracts, preserving exact public APIs, expected values, and error messages. This capability is then preserved in node \#8, which also solves \texttt{javascript\_\_queen-attack}. Crucially, node \#8 does not merely improve its own lineage. Through Cross-lineage Hybridization, it serves as a successful context donor for a separate failing lineage rooted at node \#3. The resulting hybrid child, node \#15, also solves the previously failed task. This case shows that hybridization is not simply another mutation operator, instead it acts as a cross-archive mechanism for capability transfer, allowing a reusable skill evolved in one branch to accelerate progress in another branch that had not discovered it independently.

\subsection{Bigger Model Does Not Always Lead to Better Results}

Since recursive self-improvement ultimately edits the agent's own codebase, it is tempting to view self-evolution as another coding task. Under this view, a larger or more coding-specialized foundation model should naturally lead to stronger self-improvement: if a model is better at coding benchmarks, it should also be better at modifying the agent scaffold. However, our experiments here show that this intuition is incomplete. 
Self-improvement is indeed implemented through code editing, but the quality of the edit depends critically on preceding diagnosis steps, where the model must infer why the current scaffold failed and what general modification should be made.

\begin{table}[h]
\vspace{2.5ex}
\centering\small
\setlength{\tabcolsep}{10pt}

\begin{tabular}{rcccccc}
\toprule
&
\multicolumn{3}{c}{Qwen3-Coder-Next-80B-A3B}
&
\multicolumn{3}{c}{Qwen3.6-35B-A3B}
\\
\cmidrule(lr){2-4} 
\cmidrule(lr){5-7}
Agent
& Initial
& HGM
& MGM
& Initial
& HGM
& MGM
\\
\midrule

Accuracy
& \cAcc{33.3}{33.3}
& \cAcc{40.0}{40.0\textsuperscript{+6.7\%}}
& \cAcc{41.7}{\textbf{41.7\textsuperscript{+8.4\%}}}
& \cAcc{68.3}{68.3}
& \cAcc{73.3}{73.3\textsuperscript{+5.0\%}}
& \cAcc{78.3}{\textbf{78.3\textsuperscript{+10.0\%}}}
\\

\% Impr.
& --
& \cImpr{20.1}{$\uparrow$ 20.1\%}
& \cImpr{25.2}{$\uparrow$ \textbf{25.2\%}}
& --
& \cImpr{7.3}{$\uparrow$ 7.3\%}
& \cImpr{14.6}{$\uparrow$ \textbf{14.6\%}}
\\

Time
& --
& \cTime{150}{150 h}
& \cTime{148}{\textbf{148 h}}
& --
& \cTime{93.02}{93.02 h}
& \cTime{96.11}{96.11 h}
\\

\bottomrule
\end{tabular}
\vspace{0.5ex}
\caption{\textbf{
Performance of coding agents evolved on SWE-bench Verified with different models.} Results evolved using 200 $\varphi$-evaluations and 24 $\varPhi$-expansions. 
For each benchmark, HGM and MGM start from the same initial scaffold.
Superscripts in accuracy denote absolute percentage-point improvements over corresponding initial agents.
Time reported as CPU wall-clock time with 8$\times$NVIDIA H100 GPUs.
}
\label{tab:improving_swe_coder}
\vspace{1.5ex}
\end{table}

To examine this, we conduct an additional experiment on SWE-bench Verified using Qwen3-Coder-Next-80B-A3B as the backbone model. Qwen3-Coder-Next is larger and explicitly optimized for coding with comparable coding ability but poorer reasoning ability, so one might expect it to produce stronger self-improving agents. Surprisingly, this is not what we observe. As shown in Table~\ref{tab:improving_swe_coder}, the initial scaffold obtains 33.3\% accuracy. Under the same budget of 200 task evaluations and 24 self-modification expansions, HGM improves the scaffold to 40.0\%, while MGM improves it to 41.7\%. MGM still outperforms HGM under this backbone, but the final evolved performance is substantially lower than the corresponding Qwen3.6-35B-A3B setting, where MGM reaches 61.7\% on SWE-bench Verified-60.

This suggests that self-improvement performance cannot be predicted solely from model size or coding specialization. The model must inspect the trajectory, diagnose the underlying failure, and transform that diagnosis into a robust scaffold-level edit. This diagnosis step is not equivalent to ordinary code completion or local bug fixing. It requires the model to reason about the interaction between prompts, tools, control flow, repository exploration, execution feedback, and prior agent behavior. If the diagnosis is shallow or incorrect, the resulting code edit may still be syntactically valid, but it can target the wrong mechanism, overfit to superficial symptoms, or introduce brittle workflow changes.

\begin{table*}[t]
\centering
\small
\begin{tabular}{rlcc}
\toprule
Benchmark & Main Capability 
& Qwen3-Coder-Next 
& Qwen3.6-35B-A3B  \\
\midrule
MMLU-Redux 
& General knowledge / robust reasoning 
& 91.18 & \textbf{93.3\textsuperscript{+2.12}} \\
MMLU-Pro 
& Harder general knowledge reasoning 
& 80.52 & \textbf{85.2\textsuperscript{+4.68}} \\
GPQA 
& Graduate-level science reasoning 
& 74.49 & \textbf{86.0\textsuperscript{+11.51}} \\
SuperGPQA 
& Broader and harder QA 
& 57.45 & \textbf{64.7\textsuperscript{+7.25}} \\
HMMT Feb 2025 
& Competition-level mathematics 
& 70.21 & \textbf{90.7\textsuperscript{+20.49}} \\
HMMT Nov 2025 
& Competition-level mathematics 
& 75.57 & \textbf{89.1\textsuperscript{+13.53}} \\
\bottomrule
\end{tabular}
\vspace{-0.5ex}
\caption{\textbf{Performance comparison between models.}
Results are compared across general knowledge, science, and mathematics reasoning benchmarks.}
\label{tab:reasoning_comparison}
\end{table*}

To better understand this result, as shown in Table~\ref{tab:reasoning_comparison}, we compare the broader reasoning abilities of Qwen3-Coder-Next-80B-A3B and Qwen3.6-35B-A3B. Although Qwen3-Coder-Next is larger and coding-oriented, Qwen3.6-35B-A3B performs substantially better on a wide range of general and reasoning-intensive benchmarks. Qwen3.6 scores 93.3 on MMLU-Redux, compared with 91.18 for Qwen3-Coder-Next, and 85.2 on MMLU-Pro, compared with 80.52. The gaps become larger on more difficult reasoning tasks: Qwen3.6 outperforms Qwen3-Coder-Next by 11.51 points on GPQA, 7.25 points on SuperGPQA, 20.49 points on HMMT Feb 2025, 13.53 points on HMMT Nov 2025, and 21.47 points on LiveCodeBench v6. These results indicate that Qwen3.6 has a stronger general reasoning profile, despite having fewer total parameters.

This reasoning advantage helps explain why Qwen3.6 leads to stronger self-improvement. In a self-evolving coding agent, the model is not only asked to write code; it is asked to reason about why the current agent failed and how the scaffold should be changed to prevent similar failures in future tasks. The edit itself is a coding operation, but deciding what to edit is a diagnosis and abstraction problem. A coding-specialized model may be strong at implementing local changes, yet still produce weaker self-improvement if it cannot reliably infer the correct failure mechanism from trajectories. Conversely, a model with stronger reasoning ability can generate more accurate diagnoses and therefore produce more reusable scaffold-level modifications.

This observation is particularly important for MGM. MGM introduces reaction-norm mutation and cross-lineage hybridization, both of which increase the amount of comparative evidence available to the self-modification step. However, richer evidence is only useful if the backbone model can reason over it. Reaction-norm mutation requires the model to compare multiple trajectories of the same agent and identify recurring behavioral patterns. Cross-lineage hybridization requires the model to compare different agents on the same task and extract transferable scaffold-level traits. These operations make the diagnosis stage more informative, but also more reasoning-intensive.

Overall, our results do not suggest that larger or more coding-specialized models are ineffective for self-improvement. Qwen3-Coder-Next still enables both HGM and MGM to improve over the initial scaffold, showing that strong  coding backbones can support meaningful self-evolution. However, the improvement is not necessarily monotonic with model size or coding specialization. Although self-evolution ultimately edits code, the quality of the edit depends on whether the model can diagnose the failure mechanism before modifying the scaffold. A larger coding model may therefore produce valid self-edits, but not necessarily better self-edits, if its trajectory-level diagnosis and abstraction ability are weaker. This suggests that backbone selection for self-improving agents should consider a broader capability profile, especially general reasoning and failure-diagnosis ability, rather than relying only on parameter count or coding benchmark performance.

\subsection{Are skills evolved from MGM really more general?}\label{general_skills}

\begin{figure*}[t]
    \centering
    \includegraphics[width=0.8\linewidth]{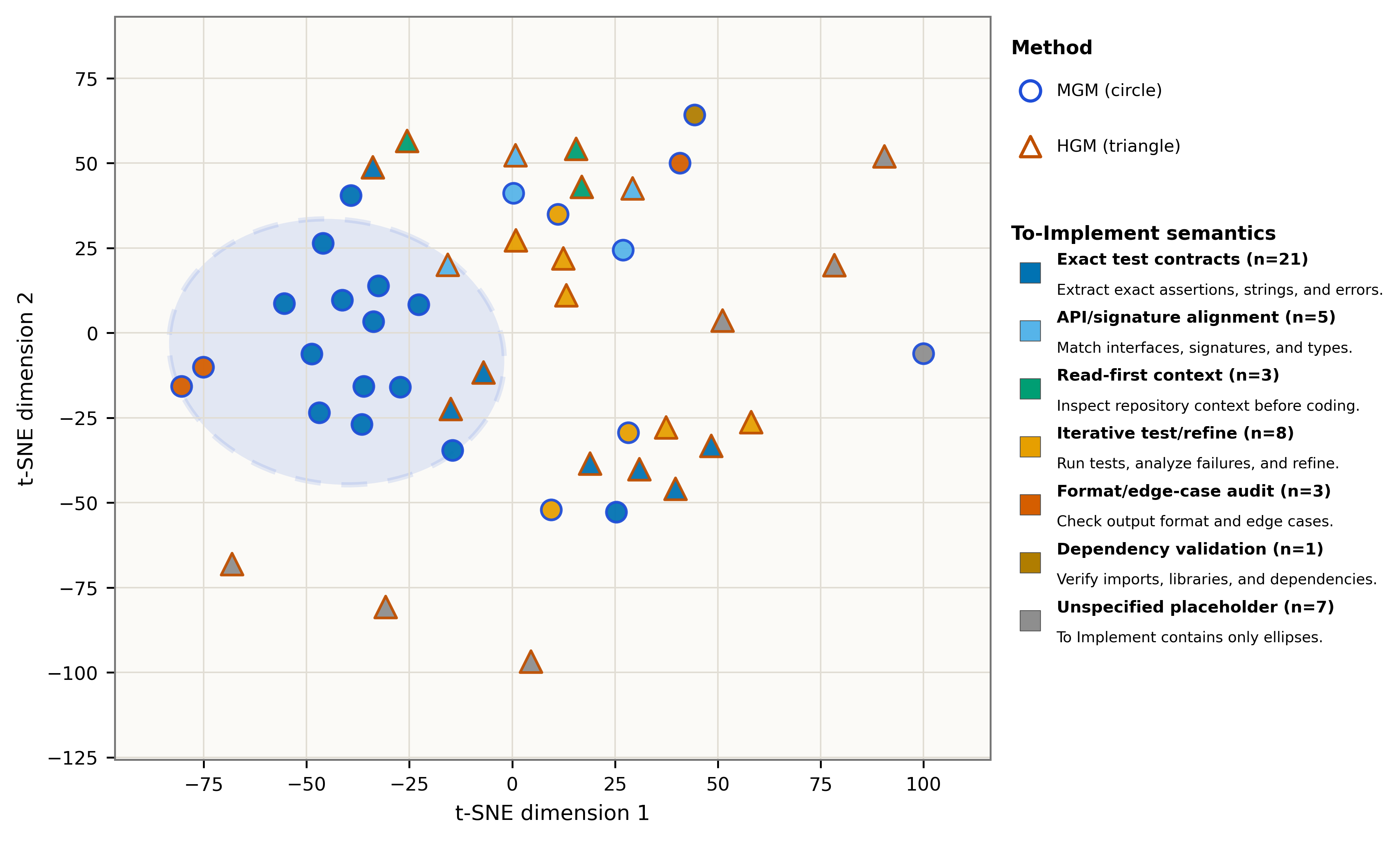}
    \caption{\textbf{Semantic visualization of evolved skills.} A content-level view of the evolved “To Implement” descriptions from MGM and HGM. Each point corresponds to one evolved skill or workflow change, embedded by the semantics of its To Implement text. The marker shape indicates the evolution method, while color indicates the main semantic theme of the proposed change.}
    \label{fig:General_Proof}
\end{figure*}

The visualization~\ref{fig:General_Proof} suggests that MGM produces changes that concentrate around a reusable workflow-level capability. 
Most MGM points lie in a compact semantic region centered on exact test-contract extraction, API-contract adherence, and pre-implementation verification. 
These are not tied to a particular programming language, benchmark instance, or repository-specific bug. 
Instead, they jointly describe a general procedure incorporating reading tests and existing interfaces, extracting the expected contract, implementing against that contract, and verifying exact agreement before finalizing the patch. 
This kind of skills can transfer across many programming tasks because almost all repair problems involve some form of latent contract between tests, existing code, and the expected implementation.

In contrast, HGM points are more widely dispersed in the semantic map. HGM also discovers useful ideas, such as iterative testing, API alignment, context discovery, and auxiliary tool utilities, but these ideas appear less consolidated into a single reusable default workflow. The broader spread of HGM points suggests a more heterogeneous search over possible interventions. Some HGM changes add or wire in tools, while others propose isolated prompt phases or iterative loops. These can be beneficial, but they are less consistently organized around one general mechanism that would apply by default across unseen tasks.

This distinction is important because generalizability should not be measured only by whether a change uses broad language such as general capability. 
A more operational criterion is whether the evolved skill abstracts away from the original training instance and becomes a reusable procedure for a broad class of future tasks. 
Under this criterion, MGM appears more general at the workflow level. It repeatedly evolves skills that convert task-specific feedback into a language-agnostic contract-extraction and verification protocol.
HGM, by comparison, evolves a wider variety of interventions, but they are more fragmented and less clearly integrated into a stable general workflow.

\section{Additional Related Work}
\label{app:related-work-details}

As foundation models continue to advance, including ~\citet{deepseekv32, deepseekv4, kimik2, kimik25, GPT5, minimaxM1}, the attainable performance ceiling of coding agents is also steadily increasing. The development of agentic systems has progressed from human-designed scaffolds~\citep{toolmakers, creator}, to automated agent design\citep{ADAS}, and more recently to self-improving agents\citep{zhang2026darwin, HGM, alita, gao2026survey}.

\subsection{From Agent Design to Inherited Self-Modification}

Modern LLM agents build on a broad line of work on tool use, reasoning-action interleaving, and multi-agent orchestration, including modular tool-augmented systems~\citep{mrkl}, ReAct-style reasoning-and-acting~\citep{yao2023react}, self-supervised tool use~\citep{schick2023toolformer}, and multi-agent frameworks such as AutoGen and MetaGPT~\citep{wu2023autogenenablingnextgenllm, hong2024metagpt}. These general agentic paradigms become especially important in software engineering, where an agent must inspect repositories, edit files, execute commands, and validate patches. Systems such as SWE-agent emphasize the importance of the agent-computer interface: a model's ability to inspect files, edit code, execute commands, and run tests depends heavily on the tools and interaction protocol exposed to it \citep{yang2024sweagent}. HyperAgent instead explores a multi-agent decomposition of software engineering work, assigning specialized roles such as planning, navigation, editing, and execution to different agents \citep{zhang2026hyperagents}. These systems show that scaffold design is central to coding-agent performance, but the scaffold is still largely engineered by humans.

ADAS shifts scaffold design from manual engineering to automated search. Meta Agent Search represents agents as code and uses a meta-agent to invent new agent programs from an archive of prior discoveries \citep{ADAS}. This view is important because it treats prompts, tools, and control flow as searchable program components rather than fixed infrastructure. However, ADAS-style methods usually preserve a separation between the designer and the designed agent. The meta-agent is responsible for generating new target agents, whereas the target agent does not necessarily improve itself through its own execution history.

Self-improving coding agents reduce this separation. SICA shows that a coding agent can use basic file-editing tools to modify its own codebase and improve on benchmarks \citep{robeyns2025sica}. DGM extends this into open-ended evolution by maintaining an archive of self-modified agents and sampling from it to create new descendants \citep{zhang2026darwin}. HGM further observes that an agent's current benchmark performance is not always aligned with its future self-improvement potential, and therefore estimates descendant-based metaproductivity to guide archive expansion \citep{HGM}. These works establish the XGM setting: an agent population evolves through persistent, heritable edits to executable scaffolds.

MGM focuses on a different bottleneck inside this setting. Prior G\"{o}del Machine-style methods mainly decide which node in the archive should be expanded. MGM instead asks what evidence should be given to the editor once an expansion is triggered. A single failed trajectory can be noisy: it may reflect a task-specific accident, a bad local choice, or a general weakness in the agent's design. MGM reduces this ambiguity by constructing controlled comparisons from trajectories that are already present in the archive. Reaction-norm mutation compares the same genotype across multiple environments, making recurring failures more likely to reveal a stable design weakness. Cross-lineage hybridization compares different genotypes on the same task, making behavioral differences easier to interpret as transferable skills. Therefore, MGM improves the diagnostic quality of self-modification without requiring additional task evaluations.

\subsection{Runtime Adaptation versus Training-Time Self-Evolution}

Live-SWE-agent is the most relevant concurrent work to distinguish from MGM. It starts from a minimal agent scaffold and lets the agent expand or revise its own capabilities during the process of solving a real-world software issue \citep{livesweagent}. This is a powerful runtime adaptation mechanism: the agent can create helper tools, refine its own execution procedure, and specialize its workflow to the current repository. Its central advantage is immediacy. It does not require an offline evolution loop before deployment, and the agent can adapt to the concrete structure of the issue it is currently solving.

MGM addresses a different question. Rather than asking how an agent should adapt inside one episode, MGM asks how a population of agents should improve across many episodes so that future agents inherit better scaffolds. The modifications produced by MGM are persistent changes to the agent program. They are evaluated over a task distribution and stored as descendants in an archive. This makes MGM a training-time self-evolution method, even though the training signal is not gradient-based. The objective is to discover general capabilities---better planning, validation, context management, debugging, or tool usage---that remain useful beyond the tasks that exposed them.

This distinction is similar to the difference between chain-of-thought prompting and policy optimization. Chain-of-thought prompting allocates more inference-time computation to a single response, improving the current trajectory without changing the model parameters \citep{wei2022chain}. GRPO, by contrast, is a reinforcement learning algorithm that updates the policy so that future trajectories improve \citep{shao2024deepseekmath}. Live-SWE-agent plays a role analogous to test-time reasoning or skill construction: it improves the current problem-solving process. MGM plays a role analogous to policy learning: it changes the inherited scaffold that future agents use. These paradigms are complementary. A strong practical system could first use MGM to evolve robust base agents offline and then use live runtime adaptation to specialize the evolved agent to the current issue.

\subsection{Open-Ended Search and Algorithm Discovery}

MGM is also related to open-ended and quality-diversity search. In these paradigms, progress does not come only from optimizing a single incumbent solution, but from maintaining an archive of diverse candidates that can serve as stepping stones for future discovery. Quality-diversity methods such as MAP-Elites maintain structured archives of high-performing yet behaviorally diverse solutions, improving both search coverage and downstream adaptability~\citep{mouret2015illuminatingsearchspacesmapping, Cully_2021}. Open-ended evolution further emphasizes that preserving diverse lineages can expose stepping stones that would be missed by purely exploitative optimization~\citep{clune2020aigasaigeneratingalgorithmsalternate}.

Recent foundation-model-based discovery systems instantiate a related idea in code and scientific domains. AlphaEvolve uses a coding agent to iteratively propose, evaluate, and improve programs for scientific and algorithmic discovery~\citep{novikov2025alphaevolvecodingagentscientific}. Generative modeling has also been used to search for mathematical objects and conjecture-relevant structures, suggesting that learned generative models can act as proposal mechanisms for discovery under external evaluators~\citep{ellenberg2025generativemodelingmathematicaldiscovery}. MGM differs from these systems in that its search space is not a standalone algorithm or mathematical object, but the executable scaffold of a coding agent itself. Rather than only preserving diverse candidates as archive entries, MGM reuses the archive as a source of comparative diagnostic evidence for inherited self-modification.

\subsection{Benchmark Coverage and Evaluation Motivation}

SWE-bench is the canonical benchmark for repository-level issue resolution. Unlike function-level benchmarks such as HumanEval~\citep{chen2021evaluating} or MBPP~\citep{austin2021program}, SWE-bench gives the model a real repository and a natural-language GitHub issue, and evaluates whether the generated patch passes tests derived from the corresponding pull request~\citep{SWEbench}. This setting stresses repository navigation, fault localization, patch construction, and validation. SWE-bench Verified further improves reliability by using a human-filtered subset, which is why it has become a standard benchmark for comparing software-engineering agents.

However, SWE-bench Verified alone is not sufficient for evaluating self-evolving agents. First, it is mostly Python-centric, so an evolved agent may overfit to Python idioms, test frameworks, or repository layouts. Second, many tasks are relatively short compared with professional software engineering work. SWE-bench Pro addresses the second limitation by introducing harder long-horizon tasks from a broader range of actively maintained repositories, often requiring deeper investigation and multi-file changes~\citep{SWEbench-Pro}. Polyglot addresses the first limitation by testing coding across several languages, including C++, Go, Java, JavaScript, Python, and Rust~\citep{polyglot}. Multilingual repository-level variants of SWE-bench provide another related direction by extending issue resolution beyond Python repositories~\citep{SWEbench-Multilingual,yang2025swesmith}.

These benchmark choices are aligned with MGM's objective. Reaction-norm mutation is intended to identify general weaknesses that recur across tasks, so it should be evaluated on settings where task diversity matters. Cross-lineage hybridization is intended to transfer useful behaviors between agents, so it should be tested on benchmarks where different strategies may solve different subsets of tasks. SWE-bench Verified, SWE-bench Pro, and Polyglot therefore provide complementary evidence: standard real-world issue resolution, long-horizon robustness, and cross-language generality.

\section{Best Discovered Agents}

\subsection{MGM on Polyglot}

Compared with the initial agent, whose \texttt{forward()} routine performs a single-turn code generation call without structured repository analysis or test feedback, agent (node~\#16) accumulates three successive modifications along the lineage \textit{initial} $\rightarrow$ \#2 $\rightarrow$ \#9 $\rightarrow$ \#16.
The first patch (\#2, introduced via cross-lineage hybridization) adds a read-first prompt directive that requires the agent to inspect stubs, tests, and class hierarchies with the \texttt{editor} tool and to implement only the interfaces already defined in the repository.
The second patch (\#9, a reaction-norm mutation) replaces the single-turn workflow with a two-phase pipeline: in Phase~1, the agent extracts a structured \texttt{contract\_plan} JSON containing exact class signatures, constructors, method definitions, and verbatim error messages from test and stub files. In Phase~2, it implements the solution under strict adherence to that plan and performs an explicit checklist-based self-audit before submission.
The third patch (\#16, also a reaction-norm mutation) extends this design with a test-driven verification loop: after implementation, the agent repeatedly executes the task's test suite (via a new \texttt{run\_tests()} method and supporting utilities in \texttt{utils/test\_utils.py}), feeds raw test output back to the model together with the remaining contract, and iteratively repairs the solution for up to five attempts until tests pass or the budget is exhausted.
Cumulatively, these changes transform the initial agent from a prompt-only code generator into a test-contract-guided, self-auditing, and empirically self-correcting coding agent capable of reducing API hallucination, enforcing exact test constraints, and recovering from runtime failures through language-agnostic test feedback.

\subsection{MGM on SWE-bench Verified}

Listing below shows an example self-modification discovered by MGM on SWE-bench Verified.
Unlike a task-specific repository patch, this modification changes the agent's general debugging workflow.
The evolved scaffold turns the original single-pass repair process into a more trace-aware and patch-constrained procedure.

Concretely, the agent first extracts test function names from the problem statement and test description, and then builds an explicit code-path trace before editing.
When relevant failing tests are available, the scaffold generates a \texttt{code\_path\_trace} that links the failing test to the likely implementation path and provides a targeted fix direction.
After the initial repair attempt, the agent further inspects the message history to identify remaining failed tests and can perform additional revision rounds conditioned on the traced failure path.
This makes the repair process less dependent on a vague natural-language issue description and more directly grounded in executable regression signals.

The modification also introduces lightweight provenance tracking for repository exploration.
By registering callbacks around the editor tool, the agent records which files were actually viewed during the debugging process.
This information is then used by a new diff-minimality filter, which removes patch blocks that are weakly related to the problem statement or to the files inspected by the agent.
As a result, the evolved scaffold encourages localized fixes and discourages broad, accidental, or speculative edits.

The concrete utility added in this example focuses on tracing docstring and autodoc-style failures, but the underlying scaffold-level skill is more general:
MGM discovers a workflow that first grounds the repair in failing tests, then traces the relevant code path, and finally constrains the submitted diff to files supported by the agent's own investigation.
This provides qualitative evidence that MGM can evolve reusable repository-level debugging habits on SWE-bench Verified, rather than merely memorizing a solution to a single benchmark instance.

\onecolumn

\captionsetup[lstlisting]{labelformat=empty}

\begin{lstlisting}[style=diffstyle, caption={Figure 17: Code modification on Polyglot}, basicstyle=\scriptsize\ttfamily]
diff --git a/coding_agent.py b/coding_agent.py
index 39cb47e..d824c3d 100644
--- a/coding_agent.py
+++ b/coding_agent.py
@@ -172,7 +172,21 @@ class AgenticSystem:

 Your task is to make changes to the files in the {self.git_dir} directory to address the <problem_description>. I have already taken care of the required dependencies.
 """
-        instruction = f"{task}\n\nPlease analyze the problem description carefully. Then make edits to the code files to complete the instruction."
+        inspection_directive = """
+**CRITICAL PRE-IMPLEMENTATION REQUIREMENT:**
+
+Before writing any code, carefully inspect the existing repository structure. Use the `editor` tool to `view` all relevant stub files, test files, and related classes. Extract and strictly adhere to their exact method signatures, class hierarchies, and constructor requirements. Do not assume or invent interfaces; implement exactly what the existing code expects.
+
+**Specific Instructions:**
+1. First, explore the repository structure using `view` commands to understand the file layout
+2. Read all stub files, interface definitions, and test files to understand the expected API
+3. Summarize the required method signatures, constructor parameters, and class hierarchies
+4. Only after thoroughly understanding the existing code structure, proceed with your implementation
+5. Implement exactly what the stubs and tests expect - never invent new interfaces
+
+**Reminder:** The coding agent needs to deal with different languages including C++, Go, Java, JavaScript, Python, and Rust. Adhere to the exact method signatures, class hierarchies, and constructor requirements defined in the existing code, regardless of programming language.
+"""
+        instruction = f"{task}\n\n{inspection_directive}\nPlease analyze the problem description carefully. Then make edits to the code files to complete the instruction."
         chat_history, n_llm_calls_used = chat_with_agent(
             instruction,
             model=self.code_model,

diff --git a/coding_agent.py b/coding_agent.py
index d824c3d..25f5411 100644
--- a/coding_agent.py
+++ b/coding_agent.py
@@ -1,6 +1,7 @@
 # This file is adapted from https://github.com/jennyzzt/dgm.

 import argparse
+import json
 import logging
 import os
 import subprocess
@@ -160,41 +161,242 @@ class AgenticSystem:

         return new_msg_history

+    def extract_json_from_response(self, response_content):
+        """
+        Extract JSON content from LLM response which may contain markdown or other text.
+        """
+        # Try to find JSON in code blocks
+        import re
+
+        json_pattern = r"```(?:json)?\s*(\{.*?\})\s*```"
+        match = re.search(json_pattern, response_content, re.DOTALL)
+        if match:
+            try:
+                return json.loads(match.group(1))
+            except json.JSONDecodeError:
+                pass
+
+        # Try to find a JSON object directly
+        try:
+            # Find the first opening brace and last closing brace
+            start = response_content.find("{")
+            end = response_content.rfind("}") + 1
+            if start >= 0 and end > start:
+                json_str = response_content[start:end]
+                return json.loads(json_str)
+        except (json.JSONDecodeError, ValueError):
+            pass
+
+        # If all else fails, return the raw content
+        return response_content
+
     def forward(self, timeout):
         """
         The forward function for the AgenticSystem.
+        Implements a two-phase workflow:
+        1. Contract Extraction Phase: Agent reads test/stub files and outputs a structured contract_plan JSON
+        2. Implementation & Verification Phase: Agent implements solution adhering to the plan with self-audit
         """
-        task = f"""I have uploaded a code repository in the directory {self.git_dir}. Help solve the following problem.
+        # Phase 1: Contract Extraction
+        contract_plan_phase_instruction = f"""I have uploaded a code repository in the directory {self.git_dir}. Help solve the following problem.

 <problem_description>
 {self.problem_statement}
 </problem_description>

-Your task is to make changes to the files in the {self.git_dir} directory to address the <problem_description>. I have already taken care of the required dependencies.
+# PHASE 1: CONTRACT EXTRACTION
+
+Your FIRST task is to extract the exact API contract from the repository. Do NOT write any implementation code in this phase.
+
+**Instructions for Phase 1:**
+
+1. **Explore the repository structure** using the `editor` tool to understand the file layout.
+
+2. **Read ALL relevant files**:
+   - Test files (files containing tests that define expected behavior)
+   - Stub/Interface files (files containing class definitions, method signatures, type hints)
+   - Any specification or documentation files
+
+3. **Extract the exact contract** which includes:
+   - **Class names and their exact hierarchy** (parent classes, interfaces implemented)
+   - **Constructor signatures** (exact parameter names, types, and order)
+   - **Method signatures** (exact parameter names, types, return types)
+   - **Error messages** (verbatim error strings that must be raised)
+   - **Module-level functions** (exact signatures if applicable)
+   - **File paths** that need modification
+
+4. **Output your findings as a JSON object** with the following structure:
+
+```json
+{{
+  "classes": [
+    {{
+      "name": "ClassName",
+      "parent": "ParentClassName or null",
+      "fields": [
+        {{"name": "field_name", "type": "Type"}}
+      ],
+      "methods": [
+        {{
+          "name": "method_name",
+          "parameters": [
+            {{"name": "param", "type": "Type", "required": true}}
+          ],
+          "return_type": "ReturnType",
+          "is_constructor": false
+        }}
+      ],
+      "file_path": "path/to/file"
+    }}
+  ],
+  "functions": [
+    {{
+      "name": "function_name",
+      "parameters": [
+        {{"name": "param", "type": "Type", "required": true}}
+      ],
+      "return_type": "ReturnType",
+      "file_path": "path/to/file"
+    }}
+  ],
+  "error_messages": [
+    {{"message": "exact error string from tests", "context": "when this error is raised"}}
+  ],
+  "files_to_modify": ["path/to/file1", "path/to/file2"],
+  "test_files": ["path/to/test_file1", "path/to/test_file2"],
+  "key_observations": [
+    "Observation 1: e.g., Constructor takes no arguments",
+    "Observation 2: e.g., Error 'NotImplementedError' must be raised with message 'feature not implemented'"
+  ]
+}}
+```
+
+**CRITICAL REQUIREMENTS FOR PHASE 1:**
+- Extract signatures EXACTLY as defined in the test/stub files
+- Do NOT infer or guess missing information - if something is unclear, note it in key_observations
+- Preserve exact error message strings (case-sensitive, including punctuation)
+- Note the EXACT file paths that contain the contracts
+- This phase is for ANALYSIS ONLY - do not write any implementation code
+
+Output only the JSON object and a brief summary of your findings. No implementation code should be generated in this phase.
 """
-        inspection_directive = """
-**CRITICAL PRE-IMPLEMENTATION REQUIREMENT:**

-Before writing any code, carefully inspect the existing repository structure. Use the `editor` tool to `view` all relevant stub files, test files, and related classes. Extract and strictly adhere to their exact method signatures, class hierarchies, and constructor requirements. Do not assume or invent interfaces; implement exactly what the existing code expects.
+        safe_log("=" * 50)
+        safe_log("PHASE 1: CONTRACT EXTRACTION")
+        safe_log("=" * 50)
+
+        # Phase 1: Extract contract from test/stub files
+        contract_chat_history, contract_calls = chat_with_agent(
+            contract_plan_phase_instruction,
+            model=self.code_model,
+            msg_history=[],
+            logging=safe_log,
+            timeout=timeout,
+        )
+
+        # Extract the JSON contract plan from the response
+        contract_response = ""
+        if contract_chat_history:
+            # Get the last assistant message
+            for msg in reversed(contract_chat_history):
+                if isinstance(msg, dict):
+                    if msg.get("role") == "assistant":
+                        content = msg.get("content", "")
+                        if content:
+                            contract_response = content
+                            break
+                else:
+                    # Handle non-dict response objects
+                    contract_response = str(msg)
+                    break
+
+        contract_plan = self.extract_json_from_response(contract_response)
+
+        if not isinstance(contract_plan, dict):
+            safe_log(f"Warning: Could not extract valid JSON contract plan. Raw response: {contract_response[:500]}")
+            contract_plan = {
+                "classes": [],
+                "functions": [],
+                "error_messages": [],
+                "files_to_modify": [],
+                "test_files": [],
+                "key_observations": [
+                    "Could not parse structured contract. Agent response should be reviewed."
+                ],
+            }
+
+        safe_log(f"Extracted contract plan with {len(contract_plan.get('classes', []))} classes and {len(contract_plan.get('functions', []))} functions")
+        safe_log(f"Files to modify: {contract_plan.get('files_to_modify', [])}")
+        safe_log(f"Key observations: {contract_plan.get('key_observations', [])}")
+
+        # Phase 2: Implementation & Verification
+        phase2_instruction = f"""I have uploaded a code repository in the directory {self.git_dir}. Help solve the following problem.
+
+<problem_description>
+{self.problem_statement}
+</problem_description>
+
+# PHASE 2: IMPLEMENTATION & VERIFICATION

-**Specific Instructions:**
-1. First, explore the repository structure using `view` commands to understand the file layout
-2. Read all stub files, interface definitions, and test files to understand the expected API
-3. Summarize the required method signatures, constructor parameters, and class hierarchies
-4. Only after thoroughly understanding the existing code structure, proceed with your implementation
-5. Implement exactly what the stubs and tests expect - never invent new interfaces
+## CONTRACT PLAN (from Phase 1)

-**Reminder:** The coding agent needs to deal with different languages including C++, Go, Java, JavaScript, Python, and Rust. Adhere to the exact method signatures, class hierarchies, and constructor requirements defined in the existing code, regardless of programming language.
+Based on analysis of test files and stub files, here is the exact API contract you MUST adhere to:
+
+```json
+{json.dumps(contract_plan, indent=2)}
+```
+
+## YOUR TASK
+
+Now you must implement the solution. You are STRICTLY BOUND by the contract above. 
+
+### Step 1: Review the Contract
+- Carefully read through the contract_plan
+- Identify all classes, methods, constructors, and error messages that need to be implemented
+- Note the exact file paths where modifications are needed
+
+### Step 2: Implement the Solution
+- Use the `editor` tool to make changes to the files listed in `files_to_modify`
+- Follow the exact signatures from the contract - do NOT invent new methods or change existing signatures
+- Implement error handling that raises the EXACT error messages specified in `error_messages`
+- Ensure all constructors match the parameters exactly as specified
+
+### Step 3: Self-Audit Checklist
+Before outputting your patch, you MUST perform a self-audit. Review your implementation against this checklist:
+
+1. [ ] **Class Names**: Do all class names match exactly (including inheritance)?
+2. [ ] **Constructor Signatures**: Do all constructor signatures match the contract parameters exactly?
+3. [ ] **Method Signatures**: Do all method signatures (name, parameters, return types) match?
+4. [ ] **Error Messages**: Do all error raises use the EXACT error strings specified?
+5. [ ] **File Paths**: Are all modifications made to the correct files?
+6. [ ] **No Extra Code**: Have you avoided adding methods or features not in the contract?
+
+For each item, explicitly state PASS or FAIL and provide evidence from your implementation.
+
+### Step 4: Output the Final Patch
+After the self-audit (all items must PASS), output your final implementation. Make sure:
+- You use the `editor` tool to edit the files
+- The implementation strictly follows the contract
+- All methods are properly implemented to pass the tests
+
+**IMPORTANT**: Do NOT write any code that deviates from the contract. Every method signature and error message must match exactly.
 """
-        instruction = f"{task}\n\n{inspection_directive}\nPlease analyze the problem description carefully. Then make edits to the code files to complete the instruction."
-        chat_history, n_llm_calls_used = chat_with_agent(
-            instruction,
+
+        safe_log("=" * 50)
+        safe_log("PHASE 2: IMPLEMENTATION & VERIFICATION")
+        safe_log("=" * 50)
+
+        # Phase 2: Implement and verify
+        implementation_chat_history, impl_calls = chat_with_agent(
+            phase2_instruction,
             model=self.code_model,
-            msg_history=[],
+            msg_history=contract_chat_history,  # Continue from phase 1
             logging=safe_log,
             timeout=timeout,
         )
-        chat_history_str = str(chat_history)
+
+        # Return the full chat history for logging
+        return implementation_chat_history


 def main():

diff --git a/coding_agent.py b/coding_agent.py
index 25f5411..e98de16 100644
--- a/coding_agent.py
+++ b/coding_agent.py
@@ -193,9 +193,10 @@ class AgenticSystem:
     def forward(self, timeout):
         """
         The forward function for the AgenticSystem.
-        Implements a two-phase workflow:
+        Implements a multi-phase workflow:
         1. Contract Extraction Phase: Agent reads test/stub files and outputs a structured contract_plan JSON
-        2. Implementation & Verification Phase: Agent implements solution adhering to the plan with self-audit
+        2. Implementation Phase: Agent implements solution adhering to the plan
+        3. Verification Loop: Agent runs tests, analyzes results, and iterates until passing or max attempts
         """
         # Phase 1: Contract Extraction
         contract_plan_phase_instruction = f"""I have uploaded a code repository in the directory {self.git_dir}. Help solve the following problem.
@@ -329,14 +330,14 @@ Output only the JSON object and a brief summary of your findings. No implementat
         safe_log(f"Files to modify: {contract_plan.get('files_to_modify', [])}")
         safe_log(f"Key observations: {contract_plan.get('key_observations', [])}")

-        # Phase 2: Implementation & Verification
+        # Phase 2: Implementation
         phase2_instruction = f"""I have uploaded a code repository in the directory {self.git_dir}. Help solve the following problem.

 <problem_description>
 {self.problem_statement}
 </problem_description>

-# PHASE 2: IMPLEMENTATION & VERIFICATION
+# PHASE 2: IMPLEMENTATION

 ## CONTRACT PLAN (from Phase 1)

@@ -373,20 +374,17 @@ Before outputting your patch, you MUST perform a self-audit. Review your impleme

 For each item, explicitly state PASS or FAIL and provide evidence from your implementation.

-### Step 4: Output the Final Patch
-After the self-audit (all items must PASS), output your final implementation. Make sure:
-- You use the `editor` tool to edit the files
-- The implementation strictly follows the contract
-- All methods are properly implemented to pass the tests
+### Step 4: Output the Final Implementation
+After the self-audit (all items must PASS), make your final implementation using the `editor` tool.

 **IMPORTANT**: Do NOT write any code that deviates from the contract. Every method signature and error message must match exactly.
 """

         safe_log("=" * 50)
-        safe_log("PHASE 2: IMPLEMENTATION & VERIFICATION")
+        safe_log("PHASE 2: IMPLEMENTATION")
         safe_log("=" * 50)

-        # Phase 2: Implement and verify
+        # Phase 2: Implement
         implementation_chat_history, impl_calls = chat_with_agent(
             phase2_instruction,
             model=self.code_model,
@@ -395,8 +393,129 @@ After the self-audit (all items must PASS), output your final implementation. Ma
             timeout=timeout,
         )

+        # Phase 3: Verification Loop
+        current_chat_history = implementation_chat_history
+        max_iterations = 5
+        
+        # Determine if we should run tests based on language
+        should_run_tests = True
+        
+        if should_run_tests:
+            # Prepare the current edits for the agent to see
+            current_edits_msg = self.get_current_edits()
+            
+            for iteration in range(1, max_iterations + 1):
+                safe_log("=" * 50)
+                safe_log(f"PHASE 3: VERIFICATION - Iteration {iteration}/{max_iterations}")
+                safe_log("=" * 50)
+                
+                # Run tests
+                test_result = self.run_tests()
+                test_output = test_result.get("output", "")
+                test_exit_code = test_result.get("exit_code", -1)
+                
+                safe_log(f"Test exit code: {test_exit_code}")
+                
+                if test_exit_code == 0:
+                    # Tests pass!
+                    safe_log("All tests passed!")
+                    break
+                
+                # Tests failed - provide raw output to agent for analysis
+                verification_instruction = f"""## TEST RESULTS (Iteration {iteration})
+
+Tests failed with exit code {test_exit_code}. Here is the raw test output - analyze it to understand what went wrong:
+
+{test_output}
+
+## REMAINING CONTRACT
+
+```json
+{json.dumps(contract_plan, indent=2)}
+```
+
+## YOUR TASK
+
+1. **Analyze the test output above** to understand what's failing
+2. **Identify which specific tests or functionality are failing**
+3. **Use the `editor` tool to fix the implementation**
+4. **Focus on fixing the issues revealed by the test output**
+
+Make targeted fixes to pass the failing tests. Then we will run tests again.
+"""
+                
+                # Add current edits context
+                if current_edits_msg and current_edits_msg != []:
+                    verification_instruction += "\n\n## CURRENT CHANGES\n"
+                    if isinstance(current_edits_msg, list) and len(current_edits_msg) > 0:
+                        edit_msg = current_edits_msg[-1] if isinstance(current_edits_msg[-1], dict) else current_edits_msg
+                        if isinstance(edit_msg, dict) and edit_msg.get("content"):
+                            content = edit_msg["content"]
+                            if isinstance(content, list):
+                                for item in content:
+                                    if isinstance(item, dict) and item.get("type") == "input_text":
+                                        verification_instruction += item.get("text", "")
+                                    elif isinstance(item, str):
+                                        verification_instruction += item
+                            else:
+                                verification_instruction += str(content)
+                
+                # Continue with agent to fix issues
+                current_chat_history, _ = chat_with_agent(
+                    verification_instruction,
+                    model=self.code_model,
+                    msg_history=current_chat_history,
+                    logging=safe_log,
+                    timeout=timeout,
+                )
+                
+                safe_log(f"Agent made changes in iteration {iteration}. Testing again...")
+
         # Return the full chat history for logging
-        return implementation_chat_history
+        return current_chat_history
+
+    def run_tests(self):
+        """
+        Run tests for the current repository.
+        Returns a dict with 'output' (str) and 'exit_code' (int).
+        """
+        try:
+            # Import the test utility
+            from utils.test_utils import run_test_command
+            
+            # Get the test command for the language
+            test_cmd = TEST_COMMANDS.get(self.language, [["pytest", "-rA", "--tb=short"]])
+            
+            # Build the command as a string for bash
+            if isinstance(test_cmd, list):
+                if len(test_cmd) == 1 and isinstance(test_cmd[0], list):
+                    cmd_parts = test_cmd[0]
+                else:
+                    cmd_parts = test_cmd
+            else:
+                cmd_parts = test_cmd
+            
+            # Convert to bash command string
+            if isinstance(cmd_parts[0], list):
+                bash_cmd = " && ".join([' '.join(cmd) for cmd in cmd_parts])
+            else:
+                bash_cmd = " && ".join(cmd_parts)
+            
+            # Run the test command using the utility
+            result = run_test_command(bash_cmd, cwd=self.git_dir)
+            
+            return {
+                "output": result["output"],
+                "exit_code": result["exit_code"]
+            }
+        except Exception as e:
+            safe_log(f"Error running tests: {str(e)}")
+            return {
+                "output": f"Error running tests: {str(e)}",
+                "exit_code": -1
+            }


 def main():
diff --git a/tests/test_test_utils.py b/tests/test_test_utils.py
new file mode 100644
index 0000000..8ac038f
--- /dev/null
+++ b/tests/test_test_utils.py
@@ -0,0 +1,119 @@
+# Tests for test_utils.py
+
+import pytest
+from utils.test_utils import parse_test_output
+
+
+class TestParseTestOutput:
+    """Tests for parse_test_output function."""
+    
+    def test_all_tests_passed(self):
+        """Test parsing output when all tests pass."""
+        output = """
+============================= test session starts ==============================
+platform linux -- Python 3.10.20, pytest-9.0.3, pluggy-1.6.0
+rootdir: /test
+collected 5 items
+
+tests/test_a.py ..                                                        [ 40%]
+tests/test_b.py ...                                                       [100%]
+
+============================== 5 passed in 0.1s ==============================
+"""
+        result = parse_test_output(output, 0)
+        assert result["passed"] == 5
+        assert result["failed"] == 0
+        assert result["error"] == 0
+        assert result["success"] == True
+        assert result["exit_code"] == 0
+    
+    def test_some_tests_failed(self):
+        """Test parsing output when some tests fail."""
+        output = """
+============================= test session starts ==============================
+platform linux -- Python 3.10.20, pytest-9.0.3, pluggy-1.6.0
+rootdir: /test
+collected 10 items
+
+tests/test_a.py ..F...                                                    [ 80%]
+tests/test_b.py ...                                                       [100%]
+
+=================================== FAILURES ===================================
+___________________________ TestClass.test_failed ___________________________
+
+    def test_failed(self):
+>       assert False
+E       assert False
+
+tests/test_a.py:6: AssertionError
+=========================== 1 failed, 9 passed in 0.2s ===========================
+"""
+        result = parse_test_output(output, 1)
+        assert result["passed"] == 9
+        assert result["failed"] == 1
+        assert result["success"] == False
+        assert result["exit_code"] == 1
+    
+    def test_tests_with_errors(self):
+        """Test parsing output when there are errors."""
+        output = """
+============================= test session starts ==============================
+platform linux -- Python 3.10.20, pytest-9.0.3, pluggy-1.6.0
+rootdir: /test
+collected 5 items
+
+tests/test_a.py E...                                                      [ 80%]
+tests/test_b.py ..                                                        [100%]
+
+==================================== ERRORS ====================================
+________________________ ERROR at setup of test_error ________________________
+FileNotFoundError: [Errno 2] No such file or directory
+=========================== 1 error, 4 passed in 0.1s ===========================
+"""
+        result = parse_test_output(output, 1)
+        assert result["passed"] == 4
+        assert result["error"] == 1
+        assert result["success"] == False
+    
+    def test_tests_with_skipped(self):
+        """Test parsing output when there are skipped tests."""
+        output = """
+============================= test session starts ==============================
+platform linux -- Python 3.10.20, pytest-9.0.3, pluggy-1.6.0
+rootdir: /test
+collected 5 items
+
+tests/test_a.py ..s..                                                    [100%]
+
+=========================== 4 passed, 1 skipped in 0.1s ===========================
+"""
+        result = parse_test_output(output, 0)
+        assert result["passed"] == 4
+        assert result["skipped"] == 1
+        assert result["success"] == True
+    
+    def test_empty_output(self):
+        """Test parsing empty output."""
+        result = parse_test_output("", 0)
+        assert result["passed"] == 0
+        assert result["failed"] == 0
+        assert result["success"] == True
+    
+    def test_exit_code_parameter(self):
+        """Test exit code parameter is used."""
+        output = "test completed"
+        result = parse_test_output(output, 5)
+        assert result["exit_code"] == 5
+        assert result["success"] == False
+    
+    def test_success_summary(self):
+        """Test summary for passing tests."""
+        output = "5 passed in 0.1s"
+        result = parse_test_output(output, 0)
+        assert result["summary"] == "Tests passed"
+    
+    def test_failure_summary(self):
+        """Test summary for failing tests."""
+        output = "2 failed, 8 passed in 0.2s"
+        result = parse_test_output(output, 1)
+        assert "2 test(s) failed" in result["summary"]
diff --git a/utils/test_utils.py b/utils/test_utils.py
new file mode 100644
index 0000000..3284dda
--- /dev/null
+++ b/utils/test_utils.py
@@ -0,0 +1,112 @@
+# Utility functions for running tests and analyzing results.
+
+import re
+from typing import Dict, Any
+
+
+def parse_test_output(output: str, exit_code: int = 0) -> Dict[str, Any]:
+    """
+    Parse test output to extract relevant information.
+    Returns a dict with 'passed', 'failed', 'error_count', 'summary'.
+    
+    Args:
+        output: Raw test output string
+        exit_code: Exit code from test execution
+        
+    Returns:
+        Dict with parsed test results
+    """
+    result = {
+        "passed": 0,
+        "failed": 0,
+        "error": 0,
+        "skipped": 0,
+        "exit_code": exit_code,
+        "success": exit_code == 0,
+        "summary": "",
+        "raw_output": output
+    }
+    
+    # Count patterns
+    passed_match = re.search(r"(\d+)\s+passed", output)
+    if passed_match:
+        result["passed"] = int(passed_match.group(1))
+    
+    failed_match = re.search(r"(\d+)\s+failed", output)
+    if failed_match:
+        result["failed"] = int(failed_match.group(1))
+    
+    error_match = re.search(r"(\d+)\s+error", output, re.IGNORECASE)
+    if error_match:
+        result["error"] = int(error_match.group(1))
+    
+    skipped_match = re.search(r"(\d+)\s+skipped", output)
+    if skipped_match:
+        result["skipped"] = int(skipped_match.group(1))
+    
+    # Determine success from exit code and content
+    if exit_code == 0:
+        result["success"] = True
+        result["summary"] = "Tests passed"
+    else:
+        result["success"] = False
+        if result["failed"] > 0:
+            result["summary"] = f"{result['failed']} test(s) failed"
+        elif result["error"] > 0:
+            result["summary"] = f"{result['error']} error(s) occurred"
+        else:
+            result["summary"] = f"Tests failed with exit code {exit_code}"
+    
+    return result
+
+
+def run_test_command(command: str, cwd: str = None) -> Dict[str, Any]:
+    """
+    Run a test command and return the results.
+    
+    Args:
+        command: Test command to run
+        cwd: Working directory
+        
+    Returns:
+        Dict with 'output', 'exit_code', and parsed results
+    """
+    try:
+        from tools.bash import tool_function
+        
+        # Execute the command
+        if cwd:
+            full_command = f"cd {cwd} && {command}"
+        else:
+            full_command = command
+            
+        output = tool_function(full_command)
+        
+        # Try to extract exit code from output
+        exit_code = 0
+        exit_match = re.search(r"exit code[:\s]+(\d+)", output, re.IGNORECASE)
+        if exit_match:
+            exit_code = int(exit_match.group(1))
+        
+        # Parse the output
+        test_result = parse_test_output(output, exit_code)
+        
+        return {
+            "output": output,
+            "exit_code": test_result["exit_code"],
+            "success": test_result["success"],
+            "summary": test_result["summary"],
+            "passed": test_result["passed"],
+            "failed": test_result["failed"],
+            "error": test_result["error"],
+        }
+    except Exception as e:
+        return {
+            "output": f"Error running tests: {str(e)}",
+            "exit_code": -1,
+            "success": False,
+            "summary": f"Test execution failed: {str(e)}",
+            "passed": 0,
+            "failed": 0,
+            "error": 0,
+        }

\end{lstlisting}

\newpage
\begin{lstlisting}[style=diffstyle, caption={Figure 18: Code modification on SWE-bench}, basicstyle=\scriptsize\ttfamily]
diff --git a/coding_agent.py b/coding_agent.py
index d19efd6..98f364a 100644
--- a/coding_agent.py
+++ b/coding_agent.py
@@ -3,16 +3,19 @@
 import argparse
 import logging
 import os
+import re
 import subprocess
 import threading
 from logging.handlers import RotatingFileHandler
 from time import time

 from llm_withtools import (CLAUDE_MODEL, OPENAI_MODEL, chat_with_agent,
-                           convert_msg_history)
-from utils.eval_utils import (get_report_score, msg_history_to_report,
-                              score_tie_breaker)
+                           convert_msg_history, set_explored_files_callback,
+                           clear_explored_files_callback)
+from utils.eval_utils import (enforce_diff_minimality, get_report_score,
+                              msg_history_to_report, score_tie_breaker)
 from utils.git_utils import apply_patch, diff_versus_commit, reset_to_commit
+from utils.trace_utils import extract_test_function_names, generate_trace_instruction

 # Thread-local storage for logger instances
 thread_local = threading.local()
@@ -96,6 +99,9 @@ class AgenticSystem:
         self.instance_id = instance_id if not self_improve else "hgm"
         self.code_model = model

+        # Track all files explored by the agent via the editor tool's view command
+        self.explored_files = []
+
         # Initialize logger and store it in thread-local storage
         self.logger = setup_logger(chat_history_file)

@@ -172,12 +178,88 @@ Your task is to run the regression tests in the {self.git_tempdir} directory to
         )
         return test_report

+    def _extract_test_failures_from_history(self, msg_history):
+        failed_tests = []
+        for msg in msg_history:
+            if not isinstance(msg, dict):
+                continue
+            content = msg.get("content") or msg.get("output") or ""
+            if not isinstance(content, str):
+                content = str(content)
+            for pattern in (
+                r"FAILED\s+\S+?\.py::(test_\w+)",
+                r"\b(test_\w+)\b.*FAILED",
+            ):
+                for match in re.findall(pattern, content, re.IGNORECASE):
+                    name = match[0] if isinstance(match, tuple) else match
+                    if name.startswith("test_") and name not in failed_tests:
+                        failed_tests.append(name)
+        return failed_tests
+
+    def _build_trace_context(self, test_names):
+        if not test_names:
+            return ""
+        blocks = []
+        for test_name in test_names[:2]:
+            blocks.append(
+                generate_trace_instruction(
+                    instance_id=self.instance_id,
+                    test_function_name=test_name,
+                    target_object=test_name.replace("test_", ""),
+                    git_tempdir=self.git_tempdir,
+                )
+            )
+        return "\n".join(blocks)
+
+    def _apply_diff_minimality(self, diff_str):
+        cleaned_diff = enforce_diff_minimality(
+            problem_statement=self.problem_statement,
+            explored_files=self.explored_files,
+            diff_str=diff_str,
+            git_tempdir=self.git_tempdir,
+        )
+        if cleaned_diff != diff_str:
+            reset_to_commit(self.git_tempdir, self.base_commit)
+            result = subprocess.run(
+                ["git", "-C", self.git_tempdir, "apply", "-"],
+                input=cleaned_diff,
+                text=True,
+                capture_output=True,
+            )
+            if result.returncode != 0:
+                print(f"Warning: Failed to apply minimized diff: {result.stderr}")
+            else:
+                print("Applied minimized diff successfully")
+        safe_log(f"Diff minimality complete. Explored files: {self.explored_files}")
+
     def forward(self, timeout=3600):
         timeout -= 60
         start_time = time()
-        """
-        The forward function for the AgenticSystem.
-        """
+        set_explored_files_callback(self.explored_files.append)
+
+        hinted_tests = extract_test_function_names(self.test_description or "")
+        hinted_tests += extract_test_function_names(self.problem_statement or "")
+        deduped = []
+        seen = set()
+        for test_name in hinted_tests:
+            if test_name not in seen:
+                seen.add(test_name)
+                deduped.append(test_name)
+        hinted_tests = deduped
+
+        pre_trace_context = self._build_trace_context(hinted_tests)
+        trace_preamble = ""
+        if pre_trace_context:
+            trace_preamble = (
+                f"\n<code_path_trace>\n{pre_trace_context}\n</code_path_trace>\n"
+                "Use this trace to locate the correct fix path before editing.\n"
+            )
+        else:
+            trace_preamble = (
+                "\nBefore editing, run the most relevant failing test(s) for this issue "
+                f"in {self.git_tempdir} so you can observe actual vs expected output.\n"
+            )
+
         instruction = f"""I have uploaded a Python code repository in the directory {self.git_tempdir}. Help solve the following problem.

 <problem_description>
@@ -187,17 +269,47 @@ Your task is to run the regression tests in the {self.git_tempdir} directory to
 <test_description>
 {self.test_description}
 </test_description>
-
-Your task is to make changes to the files in the {self.git_tempdir} directory to address the <problem_description>. I have already taken care of the required dependencies.
+{trace_preamble}
+Make changes in {self.git_tempdir} to address the problem.
 """
-        chat_history, n_llm_calls_used = chat_with_agent(
+        msg_history, _ = chat_with_agent(
             instruction,
             model=self.code_model,
             msg_history=[],
             logging=safe_log,
             timeout=timeout - (time() - start_time),
         )
-        chat_history_str = str(chat_history)
+
+        generic_history = convert_msg_history(msg_history, self.code_model)
+        failed_tests = self._extract_test_failures_from_history(generic_history)
+        if not failed_tests and hinted_tests:
+            failed_tests = hinted_tests[:1]
+
+        for attempt in range(2):
+            remaining = timeout - (time() - start_time)
+            if not failed_tests or remaining <= 120:
+                break
+            trace_instruction = self._build_trace_context(failed_tests[:1])
+            follow_up = f"""The following test(s) failed: {', '.join(failed_tests)}.
+
+{trace_instruction}
+
+Revise your fix in {self.git_tempdir} so the patch addresses the traced code path and makes the failing test pass.
+"""
+            msg_history, _ = chat_with_agent(
+                follow_up,
+                model=self.code_model,
+                msg_history=msg_history,
+                logging=safe_log,
+                timeout=remaining,
+            )
+            generic_history = convert_msg_history(msg_history, self.code_model)
+            failed_tests = self._extract_test_failures_from_history(generic_history)
+            if not failed_tests:
+                break
+
+        clear_explored_files_callback()
+        self._apply_diff_minimality(self.get_current_edits())


 def main():
diff --git a/llm_withtools.py b/llm_withtools.py
index ba7ea87..b35bae3 100644
--- a/llm_withtools.py
+++ b/llm_withtools.py
@@ -13,6 +13,28 @@ import openai
 from llm import create_client
 from tools import load_all_tools

+# Module-level callback for tracking explored files
+_EXPLORED_FILES_CALLBACK = None
+
+
+def set_explored_files_callback(callback):
+    global _EXPLORED_FILES_CALLBACK
+    _EXPLORED_FILES_CALLBACK = callback
+
+
+def clear_explored_files_callback():
+    global _EXPLORED_FILES_CALLBACK
+    _EXPLORED_FILES_CALLBACK = None
+
+
+def _track_editor_view(tool_name, tool_input):
+    """Helper to track when editor view is called."""
+    if _EXPLORED_FILES_CALLBACK and tool_name == "editor":
+        command = tool_input.get("command", "")
+        path = tool_input.get("path", "")
+        if command == "view" and path:
+            _EXPLORED_FILES_CALLBACK(path)
+
 CLAUDE_MODEL = "anthropic/claude-sonnet-4"
 OPENAI_MODEL = "gpt-5"
 MAX_XML_TOOL_FORMAT_RETRIES = 2
@@ -80,6 +102,9 @@ def _assistant_message_with_tool_call(message, tool_use):


 def process_tool_call(tools_dict, tool_name, tool_input):
+    # Track editor tool usage
+    _track_editor_view(tool_name, tool_input)
+    
     try:
         if tool_name in tools_dict:
             return tools_dict[tool_name]["function"](**tool_input)
diff --git a/utils/eval_utils.py b/utils/eval_utils.py
index 1c6e117..15a9d83 100644
--- a/utils/eval_utils.py
+++ b/utils/eval_utils.py
@@ -125,3 +125,167 @@ Your response will be automatically parsed, so ensure that the string response i
     except Exception as e:
         logging(f"Error in score_tie_breaker: {e}")
     return best_score_index
+
+
+def _extract_file_from_diff_header(line: str) -> str:
+    """
+    Extract the file path from a git diff header line like:
+    'diff --git a/path/to/file.py b/path/to/file.py'
+    Returns the path after 'a/' (which should match the path after 'b/').
+    """
+    # Match the pattern: diff --git a/... b/...
+    import re
+    match = re.search(r'diff --git a/(.+?) b/(.+)$', line)
+    if match:
+        # Return the path from either side (they should be the same)
+        return match.group(1).split('/')[-1] or match.group(2).split('/')[-1]
+    return None
+
+
+def _get_relevant_test_files(explored_files: list, all_files: set) -> set:
+    """
+    Identify test files that are likely related to the explored files.
+    A test file is relevant if:
+    - It's in a 'test_' prefix or '_test' suffix pattern
+    - It's in a 'tests/' or 'test/' directory
+    - The base module name matches (e.g., inspect.py -> test_inspect.py)
+    """
+    relevant = set()
+    for explored in explored_files:
+        # Get the filename without extension
+        import os
+        base = os.path.splitext(os.path.basename(explored))[0]
+        # Check if this is a test-related file (e.g., test_inspect.py for inspect.py)
+        # Look for files that have test_ prefix or _test suffix with matching base name
+        for f in all_files:
+            f_base = os.path.splitext(os.path.basename(f))[0]
+            f_dir = os.path.dirname(f)
+            # Check test directory pattern
+            if os.path.basename(f_dir).lower() in ('test', 'tests'):
+                relevant.add(f)
+                continue
+            # Check naming pattern: test_<module>.py or <module>_test.py
+            if (f_base == f'test_{base}' or 
+                f_base == f'{base}_test' or
+                f == f'test_{base}.py' or
+                f == f'{base}_test.py'):
+                relevant.add(f)
+    return relevant
+
+
+def enforce_diff_minimality(
+    problem_statement: str,
+    explored_files: list,
+    diff_str: str,
+    git_tempdir: str = None,
+) -> str:
+    """
+    Enforces diff minimality by filtering out changes to files that are not 
+    relevant to the problem statement or the files explored by the agent.
+    
+    This function:
+    1. Parses the diff to extract modified file paths
+    2. Computes a relevance score for each file based on:
+       - Whether the file appears in the problem statement (exact match or keyword match)
+       - Whether the file was explored by the agent
+       - Whether it's a test file related to an explored file
+    3. Returns a filtered diff containing only high-relevance files
+    
+    Args:
+        problem_statement: The original problem statement/issue description
+        explored_files: List of file paths that the agent viewed during execution
+        diff_str: The complete diff string to filter
+        git_tempdir: Optional path to the git repository (used to find all files)
+    
+    Returns:
+        A filtered diff string containing only changes to relevant files
+    """
+    import os
+    if not diff_str:
+        return diff_str
+    
+    # Normalize problem statement for matching
+    problem_lower = problem_statement.lower()
+    
+    # Extract file paths from the diff
+    import re
+    diff_lines = diff_str.split('\n')
+    diff_blocks = []
+    current_block = []
+    
+    for line in diff_lines:
+        if line.startswith('diff --git'):
+            if current_block:
+                diff_blocks.append('\n'.join(current_block))
+            current_block = [line]
+        else:
+            current_block.append(line)
+    if current_block:
+        diff_blocks.append('\n'.join(current_block))
+    
+    # Analyze each diff block
+    relevant_blocks = []
+    for block in diff_blocks:
+        if not block.strip():
+            continue
+        
+        # Get the file path from the diff header
+        header_line = block.split('\n')[0]
+        filename = _extract_file_from_diff_header(header_line)
+        if filename is None:
+            # If we can't parse the header, keep the block to be safe
+            relevant_blocks.append(block)
+            continue
+        
+        # Compute relevance score
+        score = 0
+        
+        # Check if the file was explored by the agent
+        for explored in explored_files:
+            explored_name = os.path.basename(explored) if explored else None
+            if explored_name == filename:
+                score += 10  # High relevance - was explored
+                break
+            # Also check if the path contains the filename
+            if filename in explored:
+                score += 5
+        
+        # Check if filename appears in problem statement
+        if filename.lower() in problem_lower:
+            score += 10
+        
+        # Check if any parent directories appear in problem statement
+        for part in filename.split('/'):
+            if part.lower() in problem_lower:
+                score += 5
+                break
+        
+        # Check if it's a test file related to explored modules
+        if git_tempdir:
+            test_files = _get_relevant_test_files(explored_files, set())
+            if any(filename in f for f in test_files):
+                score += 8
+        
+        # Keep blocks with sufficient relevance score
+        # Threshold of 5 ensures we only keep clearly relevant changes
+        if score >= 5:
+            relevant_blocks.append(block)
+        else:
+            # Log for debugging (optional)
+            print(f"[diff_minimality] Filtering out {filename} (score: {score})")
+    
+    # Reconstruct the diff with only relevant blocks
+    result = '\n\n'.join(relevant_blocks)
+    
+    # Ensure trailing newline
+    if result and not result.endswith('\n'):
+        result += '\n'
+    
+    return result
+
+
+def _ensure_patch_trailing_newline(patch_str):
+    """Ensure patch string ends with newline."""
+    if patch_str and not patch_str.endswith('\n'):
+        return patch_str + '\n'
+    return patch_str
diff --git a/utils/trace_utils.py b/utils/trace_utils.py
new file mode 100644
index 0000000..b030135
--- /dev/null
+++ b/utils/trace_utils.py
@@ -0,0 +1,475 @@
+# Utility functions for tracing docstring code paths in autodoc-related issues.
+
+import os
+import re
+import subprocess
+from pathlib import Path
+from typing import Dict, List, Optional, Tuple
+
+
+def run_grep(pattern: str, directory: str, file_pattern: str = "*.py", ignore_case: bool = True, extended: bool = False) -> List[str]:
+    """
+    Run grep in a directory for a pattern in files matching file_pattern.
+    Returns a list of matching lines.
+    
+    Args:
+        pattern: The regex pattern to search for
+        directory: The directory to search in
+        file_pattern: The file pattern to match (e.g., "*.py")
+        ignore_case: Whether to use case-insensitive matching
+        extended: Whether to use extended regex mode (-E flag)
+    """
+    cmd = ["grep", "-r", "--include=" + file_pattern]
+    if ignore_case:
+        cmd.append("-i")
+    if extended:
+        cmd.append("-E")
+    cmd.extend(["-n", "--color=never", pattern, directory])
+    try:
+        result = subprocess.run(
+            cmd,
+            capture_output=True,
+            text=True,
+            timeout=30,
+        )
+        if result.returncode in (0, 1):  # 0 = match found, 1 = no match
+            return result.stdout.strip().split("\n") if result.stdout.strip() else []
+        return []
+    except subprocess.TimeoutExpired:
+        return []
+    except Exception:
+        return []
+
+
+def run_agrep(pattern: str, directory: str, file_pattern: str = "*.py") -> List[str]:
+    """
+    Run ag (the silver searcher) for faster grep-like searching.
+    Falls back to grep if ag is not available.
+    """
+    cmd = ["ag", "--python", "--literal", "--color=never", pattern, directory]
+    try:
+        result = subprocess.run(
+            cmd,
+            capture_output=True,
+            text=True,
+            timeout=30,
+        )
+        if result.returncode in (0, 1):
+            return result.stdout.strip().split("\n") if result.stdout.strip() else []
+        return []
+    except (subprocess.TimeoutExpired, FileNotFoundError):
+        # Fall back to grep
+        return run_grep(pattern, directory, file_pattern)
+
+
+def find_test_file(git_tempdir: str, test_function_name: str) -> Optional[str]:
+    """
+    Find the test file containing the given test function name.
+    Returns the file path relative to git_tempdir, or None if not found.
+    """
+    # Use grep with extended regex to find the function definition
+    # Pattern: def test_function_name(
+    results = run_grep(
+        rf'def\s+{re.escape(test_function_name)}\s*\(',
+        git_tempdir,
+        file_pattern="*.py",
+        extended=True
+    )
+    if results:
+        for line in results:
+            if ":" in line and line.strip():
+                # Extract file path (format is usually "filename:line_number:content")
+                file_path = line.split(":")[0]
+                return file_path
+    return None
+
+
+def extract_do_autodoc_call(test_content: str) -> Optional[Dict[str, str]]:
+    """
+    Extract the do_autodoc(app, '<type>', '<object>') call from test content.
+    Returns a dict with 'type' and 'object' keys, or None if not found.
+    """
+    # Match patterns like: do_autodoc(app, 'class', 'SomeClass')
+    # or: do_autodoc(app, "class", "SomeClass")
+    pattern = r"do_autodoc\s*\(\s*app\s*,\s*['\"]([^'\"]+)['\"]\s*,\s*['\"]([^'\"]+)['\"]\s*\)"
+    match = re.search(pattern, test_content)
+    if match:
+        return {
+            "type": match.group(1),
+            "object": match.group(2),
+        }
+    return None
+
+
+def extract_expected_output(test_content: str) -> Optional[str]:
+    """
+    Extract the expected output from an assertion in the test.
+    Looks for patterns like:
+    - assert 'expected text' in result
+    - assert result == 'expected text'
+    - assert "expected text" in result
+    """
+    # Pattern for assert result == 'expected' or assert result == "expected"
+    pattern = r"assert\s+.*?==\s*['\"](.+?)['\"]"
+    matches = re.findall(pattern, test_content, re.DOTALL)
+    if matches:
+        return matches[-1]  # Return the last match (usually the most relevant)
+    
+    # Pattern for assert 'expected' in result or assert "expected" in result
+    pattern = r"assert\s+['\"](.+?)['\"]\s+in\s+.*?"
+    matches = re.findall(pattern, test_content, re.DOTALL)
+    if matches:
+        return matches[-1]
+    
+    return None
+
+
+def find_documenter_class(git_tempdir: str, obj_type: str) -> Optional[str]:
+    """
+    Find the documenter class that handles the given object type.
+    For example, 'class' -> 'DataDocumenter', 'module' -> 'ModuleDocumenter', etc.
+    """
+    # Map common autodoc types to documenter naming patterns
+    type_to_pattern = {
+        "class": r"(?:^|\s)(DataDocumenter|ClassDocumenter)\b",
+        "function": r"(?:^|\s)(FunctionDocumenter|MethodDocumenter)\b",
+        "method": r"(?:^|\s)(MethodDocumenter)\b",
+        "attribute": r"(?:^|\s)(AttributeDocumenter|DataDocumenter)\b",
+        "module": r"(?:^|\s)(ModuleDocumenter)\b",
+        "exception": r"(?:^|\s)(ExceptionDocumenter|DataDocumenter)\b",
+        "data": r"(?:^|\s)(DataDocumenter)\b",
+    }
+    
+    pattern = type_to_pattern.get(obj_type.lower(), r"(?:^|\s)(\w+Documenter)\b")
+    
+    # Search for documenter classes with extended regex
+    results = run_grep(pattern, git_tempdir, file_pattern="*.py", extended=True)
+    for line in results:
+        match = re.search(pattern, line)
+        if match:
+            return match.group(1)
+    
+    # If no specific match, look for Documenter base class references
+    results = run_grep(r"class\s+\w+Documenter\s*\(", git_tempdir, file_pattern="*.py", extended=True)
+    for line in results:
+        if "Documenter" in line:
+            match = re.search(r"class\s+(\w+Documenter)", line)
+            if match:
+                return match.group(1)
+    
+    return None
+
+
+def trace_get_object_doc(git_tempdir: str, documenter_class: str) -> Optional[Dict]:
+    """
+    Trace the get_object_doc method in the given documenter class.
+    Returns information about how docstrings are read.
+    """
+    # Use extended regex for better pattern matching
+    results = run_grep(
+        r'def\s+get_object_doc\s*\(',
+        git_tempdir,
+        file_pattern="*.py",
+        extended=True
+    )
+    
+    for line in results:
+        if documenter_class.lower() in line.lower() or "Documenter" in line:
+            # Extract file path
+            parts = line.split(":")
+            if len(parts) >= 2:
+                file_path = parts[0]
+                # Read the function to check for __doc__ access
+                try:
+                    content = Path(file_path).read_text()
+                    # Look for self.object.__doc__ or self.object.__doc__
+                    if "self.object.__doc__" in content or "self.object.__doc__" in content:
+                        return {
+                            "file": file_path,
+                            "method": "get_object_doc",
+                            "doc_source": "__doc__",
+                            "uses_self_object_doc": True,
+                        }
+                    # Check for .__doc__ attribute access more generally
+                    doc_access_pattern = r"self\.object\s*\.\s*__doc__"
+                    if re.search(doc_access_pattern, content):
+                        return {
+                            "file": file_path,
+                            "method": "get_object_doc",
+                            "doc_source": "__doc__",
+                            "uses_self_object_doc": True,
+                        }
+                except Exception:
+                    pass
+    return None
+
+
+def check_module_analyzer(git_tempdir: str) -> dict:
+    """
+    Check if ModuleAnalyzer is imported and used in the codebase.
+    Returns a dict with import status and usage information.
+    """
+    # Check for ModuleAnalyzer imports using extended regex
+    # Use \S+ instead of \w+ to handle dotted module paths
+    import_patterns = [
+        r'from\s+\S+\s+import\s+.*ModuleAnalyzer',
+        r'import\s+\S+\.ModuleAnalyzer',
+    ]
+    
+    imported = False
+    import_lines = []
+    
+    for pattern in import_patterns:
+        results = run_grep(pattern, git_tempdir, file_pattern="*.py", extended=True)
+        import_lines.extend(results)
+        if results:
+            imported = True
+    
+    # Check for get_comments usage (ModuleAnalyzer's method for getting #: comments)
+    get_comments_results = run_grep(
+        r'get_comments',
+        git_tempdir,
+        file_pattern="*.py",
+        extended=True
+    )
+    uses_get_comments = len(get_comments_results) > 0
+    
+    # Check for comment-based docstring handling
+    comment_patterns = [
+        r'#:\s*\w',
+        r'comment.*docstring',
+        r'docstring.*comment',
+    ]
+    
+    has_comment_handling = False
+    for pattern in comment_patterns:
+        if run_grep(pattern, git_tempdir, file_pattern="*.py", extended=True):
+            has_comment_handling = True
+            break
+    
+    return {
+        "is_imported": imported,
+        "import_lines": import_lines[:5],  # Limit to first 5 lines
+        "uses_get_comments": uses_get_comments,
+        "has_comment_handling": has_comment_handling,
+    }
+
+
+def find_fix_location(git_tempdir: str, documenter_class: str, doc_info: dict) -> str:
+    """
+    Determine the likely fix location based on the analysis.
+    Returns a human-readable description of where to fix.
+    """
+    if not doc_info:
+        return "Could not determine fix location from code path analysis."
+    
+    file_path = doc_info.get("file", "unknown")
+    doc_source = doc_info.get("doc_source", "unknown")
+    
+    module_analyzer_info = check_module_analyzer(git_tempdir)
+    
+    if doc_source == "__doc__" and not module_analyzer_info.get("is_imported", False):
+        fix_suggestion = (
+            f"Fix: Modify `{documenter_class}.get_object_doc()` to fall back to "
+            f"`ModuleAnalyzer.get_comments()` when `__doc__` is empty. "
+            f"File: `{file_path}`"
+        )
+    elif doc_source == "__doc__" and module_analyzer_info.get("is_imported", False):
+        fix_suggestion = (
+            f"Fix: Modify `{documenter_class}.get_object_doc()` in `{file_path}` to check "
+            f"`ModuleAnalyzer.get_comments()` as a fallback when `self.object.__doc__` is None/empty. "
+            f"The current code only reads from `__doc__` and ignores comment-based docstrings."
+        )
+    else:
+        fix_suggestion = (
+            f"Examine `{documenter_class}.get_object_doc()` in `{file_path}` to understand "
+            f"how docstrings are currently retrieved. Consider adding fallback logic."
+        )
+    
+    return fix_suggestion
+
+
+def trace_docstring_path(
+    instance_id: str,
+    test_function_name: str,
+    target_object: str,
+    git_tempdir: str,
+) -> str:
+    """
+    Trace the docstring code path for autodoc-related issues.
+    
+    This function:
+    1. Parses the test file to extract the do_autodoc(app, '<type>', '<object>') call
+    2. Finds the documenter class that handles the given type
+    3. Traces through get_object_doc() to identify where self.object.__doc__ is read
+    4. Checks if ModuleAnalyzer is imported and used
+    5. Outputs a summary of the code path with fix suggestions
+    
+    Args:
+        instance_id: The instance ID for the issue
+        test_function_name: The name of the failing test function
+        target_object: The target object being documented
+        git_tempdir: Path to the git repository directory
+    
+    Returns:
+        A formatted summary string describing the code path and fix location.
+    """
+    # Step 1: Find and parse the test file
+    test_file = find_test_file(git_tempdir, test_function_name)
+    
+    if not test_file:
+        return (
+            f"[Trace Summary] Could not find test file containing '{test_function_name}' "
+            f"in {git_tempdir}. Manual investigation required."
+        )
+    
+    # Read the test file content
+    try:
+        test_path = os.path.join(git_tempdir, test_file)
+        test_content = Path(test_path).read_text()
+    except Exception as e:
+        return f"[Trace Summary] Could not read test file {test_file}: {e}"
+    
+    # Step 2: Extract do_autodoc call information
+    autodoc_info = extract_do_autodoc_call(test_content)
+    
+    if autodoc_info:
+        obj_type = autodoc_info["type"]
+        obj_name = autodoc_info["object"]
+    else:
+        # Fallback: use target_object if autodoc call not found
+        obj_type = "class"  # Default assumption
+        obj_name = target_object
+    
+    expected_output = extract_expected_output(test_content)
+    
+    # Step 3: Find the documenter class
+    documenter_class = find_documenter_class(git_tempdir, obj_type)
+    
+    if not documenter_class:
+        return (
+            f"[Trace Summary] Could not identify the documenter class handling '{obj_type}' type. "
+            f"Test function: {test_function_name}, Target: {target_object}"
+        )
+    
+    # Step 4: Trace get_object_doc
+    doc_info = trace_get_object_doc(git_tempdir, documenter_class)
+    
+    # Step 5: Check ModuleAnalyzer
+    module_analyzer_info = check_module_analyzer(git_tempdir)
+    
+    # Step 6: Find fix location
+    fix_suggestion = find_fix_location(git_tempdir, documenter_class, doc_info)
+    
+    # Build the summary
+    summary_lines = [
+        "=" * 60,
+        "DOCSTRING CODE PATH TRACE SUMMARY",
+        "=" * 60,
+        "",
+        f"Instance: {instance_id}",
+        f"Test Function: {test_function_name}",
+        f"Target Object: {target_object}",
+        f"Test File: {test_file}",
+        "",
+        "--- Analysis Details ---",
+        "",
+        f"1. autodoc() call detected: do_autodoc(app, '{obj_type}', '{obj_name}')",
+    ]
+    
+    if expected_output:
+        summary_lines.append(f"   Expected output contains: '{expected_output[:100]}...'")
+    
+    summary_lines.extend([
+        "",
+        f"2. Documenter class: {documenter_class} (handles '{obj_type}' objects)",
+    ])
+    
+    if doc_info:
+        summary_lines.append(
+            f"3. Docstring source: {documenter_class}.get_object_doc() "
+            f"in {doc_info.get('file', 'unknown')} reads from {doc_info.get('doc_source', 'unknown')}"
+        )
+    
+    summary_lines.extend([
+        "",
+        "4. ModuleAnalyzer status:",
+        f"   - Imported: {'Yes' if module_analyzer_info['is_imported'] else 'No'}",
+        f"   - Uses get_comments(): {'Yes' if module_analyzer_info['uses_get_comments'] else 'No'}",
+        f"   - Has comment-based handling: {'Yes' if module_analyzer_info['has_comment_handling'] else 'No'}",
+        "",
+        "--- Recommended Fix ---",
+        "",
+        fix_suggestion,
+        "",
+        "=" * 60,
+    ])
+    
+    return "\n".join(summary_lines)
+
+
+def extract_test_function_names(text: str) -> List[str]:
+    """Extract pytest-style test function names mentioned in free text."""
+    if not text:
+        return []
+    names = re.findall(r"\b(test_[A-Za-z0-9_]+)\b", text)
+    seen = set()
+    out = []
+    for name in names:
+        if name not in seen:
+            seen.add(name)
+            out.append(name)
+    return out
+
+
+def generate_trace_instruction(instance_id: str, test_function_name: str, target_object: str, git_tempdir: str) -> str:
+    """
+    Generate a formatted instruction for the agent to trace the docstring path.
+    This is a wrapper that formats the output for consumption by the agent.
+    """
+    trace_result = trace_docstring_path(instance_id, test_function_name, target_object, git_tempdir)
+    
+    instruction = f"""
+<code_path_trace>
+{trace_result}
+</code_path_trace>
+
+Please examine the code path identified above. The trace shows how docstrings are being read and where the fix should be applied. Focus on the recommended fix location and ensure your changes address the root cause.
+"""
+    return instruction
+
+
+def test_trace_docstring_path():
+    """
+    Test function to verify trace_docstring_path works correctly.
+    This is a self-test function that doesn't require a real repository.
+    """
+    # This is a basic sanity check - actual testing requires a real repository
+    try:
+        result = trace_docstring_path(
+            instance_id="test_instance",
+            test_function_name="test_placeholder",
+            target_object="TestClass",
+            git_tempdir="/tmp"
+        )
+        assert "Trace Summary" in result or "Could not" in result
+        return True
+    except Exception as e:
+        print(f"Test error: {e}")
+        return False
+
+
+if __name__ == "__main__":
+    import sys
+    if len(sys.argv) >= 4:
+        instance_id = sys.argv[1]
+        test_function_name = sys.argv[2]
+        target_object = sys.argv[3]
+        git_tempdir = sys.argv[4] if len(sys.argv) > 4 else "/tmp"
+        
+        result = trace_docstring_path(instance_id, test_function_name, target_object, git_tempdir)
+        print(result)
+    else:
+        # Run self-test
+        success = test_trace_docstring_path()
+        print(f"Self-test {'passed' if success else 'failed'}")

\end{lstlisting}

\end{document}